\documentclass[11pt]{article}

\usepackage[preprint]{acl}

\usepackage{times}
\usepackage{latexsym}

\usepackage[T1]{fontenc}

\usepackage[utf8]{inputenc}

\usepackage{microtype}

\usepackage{inconsolata}

\usepackage{graphicx}

\usepackage{subcaption}
\usepackage{amsmath,amsfonts,bm}
\usepackage{tcolorbox}
\usepackage{booktabs}
\usepackage{diagbox}
\usepackage{xspace}
\usepackage{multirow}
\usepackage{stfloats}
\usepackage{enumitem}

\usepackage{amsmath,amsfonts,bm}

\def\eqref#1{equation~\ref{#1}}

\def\1{\bm{1}}

\def\vtheta{{\bm{\theta}}}

\def\mS{{\bm{S}}}
\def\mT{{\bm{T}}}

\DeclareMathAlphabet{\mathsfit}{\encodingdefault}{\sfdefault}{m}{sl}
\SetMathAlphabet{\mathsfit}{bold}{\encodingdefault}{\sfdefault}{bx}{n}

\title{Thinking at the Right Size: \\Amortized Distillation Across Post-Trained LLMs}

\author{
 \textbf{Yan Zhou\textsuperscript{1}},
 \textbf{Sara Kangaslahti\textsuperscript{1}\thanks{Equal contribution.}},
 \textbf{Jonathan Geuter\textsuperscript{1,2}\footnotemark[1]},
 \textbf{Nihal V. Nayak\textsuperscript{1}},
\\
 \textbf{Marco Fumero\textsuperscript{3}},
 \textbf{Francesco Locatello\textsuperscript{3}},
 \textbf{David Alvarez-Melis\textsuperscript{1,2}}
\\
\\
 \textsuperscript{1}Harvard University,
 \textsuperscript{2}Kempner Institute,
 \textsuperscript{3}IST Austria
\\
 \small{
   \textbf{Correspondence:} \href{mailto:terryzhou@fas.harvard.edu}{terryzhou@fas.harvard.edu}
 }
}
\newif\ifcomments
\commentstrue

\usepackage[dvipsnames]{xcolor}

\newcommand{\sys}{ADAPT\xspace}

\newcommand{\noop}[1]{}

\begin{document}
\maketitle
\begin{abstract}
Practical deployment of large language models (LLMs) requires families of post-trained variants---instruction-tuned, reasoning-tuned, and chat-style models---each at multiple sizes to meet diverse latency and memory budgets. Producing each (variant, size) pair independently is prohibitive, so model families typically span only a handful of coarse-grained sizes per post-trained variant. 
Boomerang distillation \cite{kangaslahti2026boomerangdistillationenableszeroshot} reduces this cost along the size axis for base models. Through model size interpolation, it constructs models of intermediate sizes from a single teacher-student pair without additional training. However, it still treats each post-trained variant as a separate object of optimization. 
We introduce \textbf{\sys}---Amortized Distillation Across Post-Trained LLMs---a framework for amortizing distillation across both axes of a model family: size and post-training variant, producing $L \times K$ models for $L$ interpolated sizes across $K$ post-trained variants with a \emph{single} distillation run. 
\sys combines two components. 
First, a two-phase distillation procedure constructs post-trained students through pre-training alignment and supervised fine-tuning distillation, enabling smooth size--performance interpolation on generation and reasoning tasks. 
Second, weight-delta initialization approximates this construction across post-trained variants by transferring the distillation-induced weight change from the base model to students initialized from different post-trained variants.
The resulting continuum of interpolated models also enables adaptive model-size selection at inference time, improving the compute--accuracy trade-off for long-form reasoning tasks.\footnote{Code: \url{https://github.com/dcml-lab/ADAPT}.}

\end{abstract}

\section{Introduction}

\begin{figure*}
    \centering
    \includegraphics[width=0.95\linewidth]{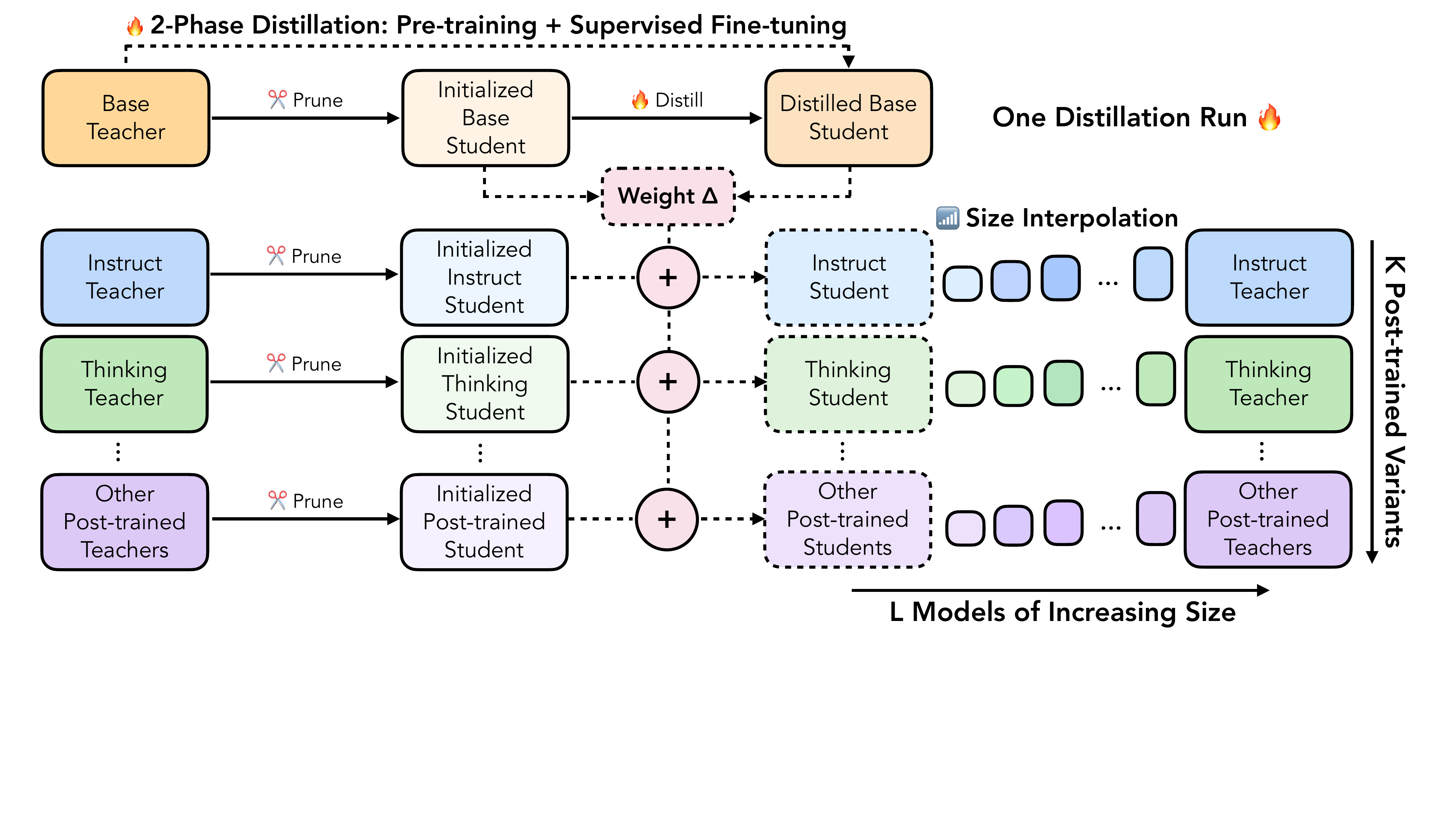}
    \vspace{-2.1cm}
    \caption{
    \textbf{\sys: Amortized Distillation Across Post-Trained LLMs.}
    We perform two-phase distillation (pre-training + SFT) on a base student once, yielding $L$ interpolated model sizes from a single teacher--student pair. We then compute the distillation-induced weight delta and add it to the initialized students from multiple post-trained teachers. This amortizes distillation for post-trained LLMs: one distillation run can now produce $L \times K$ models across $K$ post-trained variants.
    }
    \vspace{-0.3cm}
    \label{fig:main_figure}
\end{figure*}

Deployed LLM applications rely overwhelmingly on post-trained models---instruction-tuned, reasoning-tuned, and chat-style variants---rather than base pre-trained models. They also span a wide range of compute environments, from mobile devices to personal laptops to on-premise servers with multi-GPU clusters, with correspondingly diverse latency and memory constraints~\citep{huyen2022designing,narayan2025cost}. Practical use thus requires families of models that vary along two axes: size, to meet deployment compute constraints, and post-trained variants, to support different downstream capabilities. Training each (variant, size) combination from scratch is computationally infeasible, so model families are typically released only at a few coarse-grained sizes per variant, and only at a few variants per base model~\citep{qwen35blog}. 

Existing approaches address one of these axes but not both. Task vectors~\citep{ilharco2023editingmodelstaskarithmetic} transfer fine-tuned capabilities across models without additional training, but are constrained to only models of the same size.
Boomerang distillation~\cite{kangaslahti2026boomerangdistillationenableszeroshot} transfers teacher capability to a fine-grained family of smaller models with a single distillation run. It thus efficiently covers the size axis, though only in the pre-trained setting and leaves the cross-variant axis unaddressed. Whether this strategy extends to post-trained models, and whether size interpolation can be applied zero-shot across multiple post-trained variants of the same base model, are the key questions this paper investigates. 

A natural first approach is to apply boomerang distillation (BD) directly to a post-trained teacher: initialize a student from the teacher, distill on pre-training data, and patch the layers back in to create intermediate-sized models. Another approach to address the cross-variant axis is to cross-patch post-trained variants onto the boomerang-distilled base model, hoping that post-trained model capabilities can be transferred through patching alone. We find that neither approach works: generation and reasoning performance collapses across small to medium sizes on in-domain and out-of-domain tasks for both setups, despite the post-trained teacher itself achieving strong performance on these tasks (Figure~\ref{fig:2_stage_distillation}). This asymmetry suggests that instruction-following depends on properties of the distillation process that the original BD recipe does not provide.\looseness-1

In response, we introduce \textbf{\sys}---Amortized Distillation Across Post-Trained LLMs---a framework for constructing post-trained model families across both model size and post-training variant axes. \sys combines two-phase distillation with weight-delta initialization. 
First, two-phase distillation distills a student for each post-trained variant, using both pre-training alignment and supervised fine-tuning distillation for instruction-following and reasoning, enabling smooth model size interpolation without any additional training.
We then provide weight-delta initialization as an approximate construction that avoids separate distillation runs by transferring the distillation-induced weight change from the base model to students initialized from different post-trained variants of the same base model.
This allows a single base-model distillation run to produce $L \times K$ models: $L$ interpolated model sizes for each of $K$ post-trained variants.

Our experiments show that \sys recovers smooth size-performance interpolation on generation and reasoning benchmarks and outperforms standard boomerang distillation and layer pruning baselines at equal compute. 
The resulting family of intermediate-sized models supports adaptive inference: routing easier inputs to smaller models traces an empirical compute–accuracy Pareto frontier that dominates any single fixed model. 
To understand why a single base-model distillation transfers to multiple post-trained variants, we examine the underlying weight-space geometry. Empirically, base and post-trained models—and their corresponding distilled students—are connected by low-loss linear interpolation paths, a form of smooth weight interpolation that is preserved through distillation. The distillation update thus operates within a connected region of weight space shared by base and post-trained models, providing a geometric explanation of why weight-arithmetic transfer succeeds in this setting.

In summary, our contributions are:
\begin{itemize}[itemsep=1pt, topsep=1pt, parsep=1pt, leftmargin=*]
    \item We introduce \sys, a framework for amortizing distillation runs to create interpolated post-trained model families across sizes and post-training variants.

    \item We show that \sys outperforms strong baselines, including boomerang distillation and layer pruning approaches, while enabling improved performance--compute trade-offs through adaptive model-size selection.

    \item We provide an empirical analysis of size-agnostic smooth weight interpolation, showing that the weight-space structure connecting base and post-trained models is preserved after distillation and helps explain the success of weight-delta initialization.
\end{itemize}

\section{Related Work}
\paragraph{Knowledge distillation.}
Knowledge distillation is an effective technique for training a smaller model using a larger teacher model~\citep{hinton2015distilling,sanh2019distilbert}.
Several LLM families use knowledge distillation to train or post-train smaller models, including Qwen~\citep{yang2025qwen3technicalreport}, DeepSeek~\citep{Guo_2025}, and Gemma~\citep{gemmateam2025gemma3technicalreport}.
These smaller LLMs show improved performance when compared to naive training, but require individual distillation runs.
In this work, we introduce a two-phase distillation pipeline and a weight-arithmetic procedure that produces a family of post-trained models using a single distillation run. 
\looseness-1

\paragraph{Post-training LLMs.}
Post-training is a critical step in the language model training pipeline, in which pre-trained LLMs are trained to follow instructions~\citep{olmo2025olmo3,Guo_2025,bercovich2025llamanemotronefficientreasoningmodels}.
Because post-training is expensive, recent work aims to reduce its compute cost by training on smaller but effective datasets~\citep{ye2025limo,nayak2026critical}, but it still requires training models at every size.
Here, we distill a single post-trained student LLM and then create interpolated post-trained LLMs of varying sizes without training additional models, thereby dramatically reducing compute cost.
\looseness-1

\paragraph{Model arithmetic.}
Model arithmetic enables practitioners to combine the skills of multiple models into a single unified model by simply adding or subtracting model weights~\citep{ilharco2023editingmodelstaskarithmetic}.
Prior work has used model arithmetic for instruction-following~\citep{ilharco2023editingmodelstaskarithmetic}, creating multitask models~\citep{yadav2024matters}, and machine unlearning~\citep{liu2025rethinking}.
In this work, we use model arithmetic to amortize student distillation by transferring the distillation-induced weight delta from a base student to students initialized from multiple post-trained variants.
\looseness-1

\paragraph{Adaptive compute.}
Adapting LLM compute to input difficulty significantly reduces latency requirements during inference~\citep{taghibakhshi2026star}. 
While recent work has shown promising results for test-time scaling~\citep{muennighoff2025s1simpletesttimescaling}, many of these approaches require training fine-grained models from scratch~\citep{kusupati2022matryoshka} or rely on heuristics such as appending thinking tokens~\citep{muennighoff2025s1simpletesttimescaling}.
In this work, we dynamically adjust inference compute by adapting the model size based on the input difficulty for reasoning and generation tasks. 
\looseness-1

\paragraph{Layer pruning.} 
Layer pruning removes transformer layers from a trained LLM under an importance criterion, sometimes followed by a healing step
\citep{men2024shortgptlayerslargelanguage,chen2025streamliningredundantlayerscompress}. 
While effective at preserving classification accuracy, recent analyses show that pruning collapses open-ended generation
\citep{shrestha2026limitslayerpruninggenerative}. 
\sys instead recovers generation capability through a targeted two-phase distillation step, and substantially outperforms layer pruning approaches such as ShortGPT and LLM-Streamline (Appendix~\ref{appendix:comparison_to_pruning}).

\section{\sys: Amortized Distillation Across Post-Trained LLMs}\label{sec:method}
We begin by summarizing boomerang distillation~\citep{kangaslahti2026boomerangdistillationenableszeroshot}, a recently proposed method for zero-shot model size interpolation. We then introduce the two versions of \sys: \sys~(distilled), a two-phase distillation framework for constructing interpolated post-trained models, and \sys~(weight-delta), a weight-delta initialization method that approximates the two-phase distillation process across model variants and constructs multiple post-trained model families from a single distillation run.\looseness=-1

\subsection{Boomerang Distillation}\label{sec:boomerang-distillation}

Boomerang distillation~\citep{kangaslahti2026boomerangdistillationenableszeroshot} distills a small student from a teacher in such a way that intermediate models can be constructed by combining student and teacher layers without additional training. 
We now discuss its three stages. 

\paragraph{Student initialization.} 
Let $\mT$ be the teacher with $N$ layers denoted by $\vtheta_{T}$, and $\mS$ with parameters $\vtheta_{S}$ and $M$ layers be the student, where $M < N$. 
The student is initialized by partitioning the $N$ teacher layers into $M$ blocks of consecutive layers. 
From each of these blocks, the first layer is used to initialize the corresponding student layer.
The student's embedding and LM head are initialized from $\mT$. 

\paragraph{Knowledge distillation.}
The initialized student is trained on a pre-training corpus such as the Pile~\citep{gao2020pile800gbdatasetdiverse} on a combined loss involving a logit-level knowledge distillation loss, such as KL, and a block-wise alignment loss, such as cosine distance, that pushes each student layer's output toward the outputs of the corresponding block of teacher layers. 
Specifically, given a training corpus $\mathcal{D}$ and a sample $x=(x_1, ..., x_L)\in\mathcal{D}$, an index $j$, and denoting the next-token distributions by $p(\cdot \ |\ x_{<j})$, the per-token loss $\mathcal{L}_{\vtheta_\mS}(x_j)$ is:

\begin{align}\label{eq:overall-loss}
\begin{split}
\mathcal{L}_{\vtheta_\mS}(x_j)
=
&\mathrm{CE}(x_j\mid x_{<j};\vtheta_\mS) \\
&+ \lambda_{\mathrm{KL}}\,\mathrm{KL}\!\left(p_\mT(\cdot\mid x_{<j})\,\big\|\,p_\mS(\cdot\mid x_{<j})\right) \\
&+ \lambda_{\cos}\sum_{i=1}^{M}\mathcal{L}_{\cos}^{(i)}(x_{<j};\vtheta_\mT,\vtheta_\mS).
\end{split}
\end{align}
Here, $\mathrm{CE}$ denotes the next-token cross-entropy loss and $\mathcal{L}_{\cos}^{(i)}$ is the cosine distance between the outputs of the $i$th student layer and the $i$th teacher block. $\lambda_{\mathrm{KL}}>0$ and $\lambda_{\mathrm{cos}}>0$ are hyperparameters that weight the KL and cosine terms relative to cross-entropy.

\paragraph{Student patching.} 
After the distillation stage, boomerang distillation constructs intermediate models by \textit{patching}---i.e., replacing student layers with their corresponding sets of teacher layers. 
We treat student patching as an important but complementary design choice and use simple front-to-back and back-to-front patching in the main experiments, selecting the better-performing of the two orders for each model (see Appendix \ref{appendix:patching_order_results} for details).

\subsection{\sys}

\subsubsection{\sys (Distilled): Two-Phase Post-Trained Distillation}\label{sec:two_phase_distillation}
We propose a unified two-phase distillation procedure for model size interpolation in post-trained models.
Replacing the original distillation stage in boomerang distillation, the student first undergoes a \textit{pre-training} phase, followed by a \textit{supervised fine-tuning} phase. 
The pre-training phase ensures that we recover the general language modeling capability in the initialized student model before restoring the instruction-following and reasoning capabilities.
In Section \ref{sec:two_phase_distillation_results}, we find that the two-phase distillation prevents overfitting and shows better out-of-domain performance. 

\paragraph{Pre-training phase.} We follow the same boomerang distillation setup over a pre-training corpus $\mathcal{D}_{\text{pre}}$. 
Given a training sequence $x = (x_1, \dots, x_L) \in \mathcal{D}_{\text{pre}}$, the loss from Equation~\ref{eq:overall-loss} is summed over all positions $x_2, ..., x_L$. 
Training on pre-training data not only mitigates catastrophic forgetting, but has even been shown to improve downstream performance~\citep{kotha2026replayingpretrainingdataimproves}.

\paragraph{Supervised fine-tuning phase.} In this phase, we transition to instruction-following data $\mathcal{D}_{\text{SFT}}$, where each sequence $(x,y)\in\mathcal{D}_{\text{SFT}}$ consists of a prompt $x=(x_1, ..., x_L)$ and a response $y=(y_1, ..., y_{L'})$. We use the same distillation objective, but loss (Equation~\ref{eq:overall-loss}) is computed only over all the response tokens. This stage allows the student to adopt instruction-following capabilities and adapt to downstream generation tasks, while preserving alignment to the teacher.

We dedicate half of training to pre-training and half to supervised fine-tuning (see Appendix~\ref{appendix:2_phase_ratio}). 
We also use a continuous LR schedule for both phases as we found in initial experiments that two separate schedulers can hurt final performance.

\subsubsection{\sys (Weight-Delta): Weight-Delta Initialization} 
\label{sec:weight-delta-init}
\begin{figure*}[!tbh]
    \centering
    \includegraphics[width=0.92\linewidth]{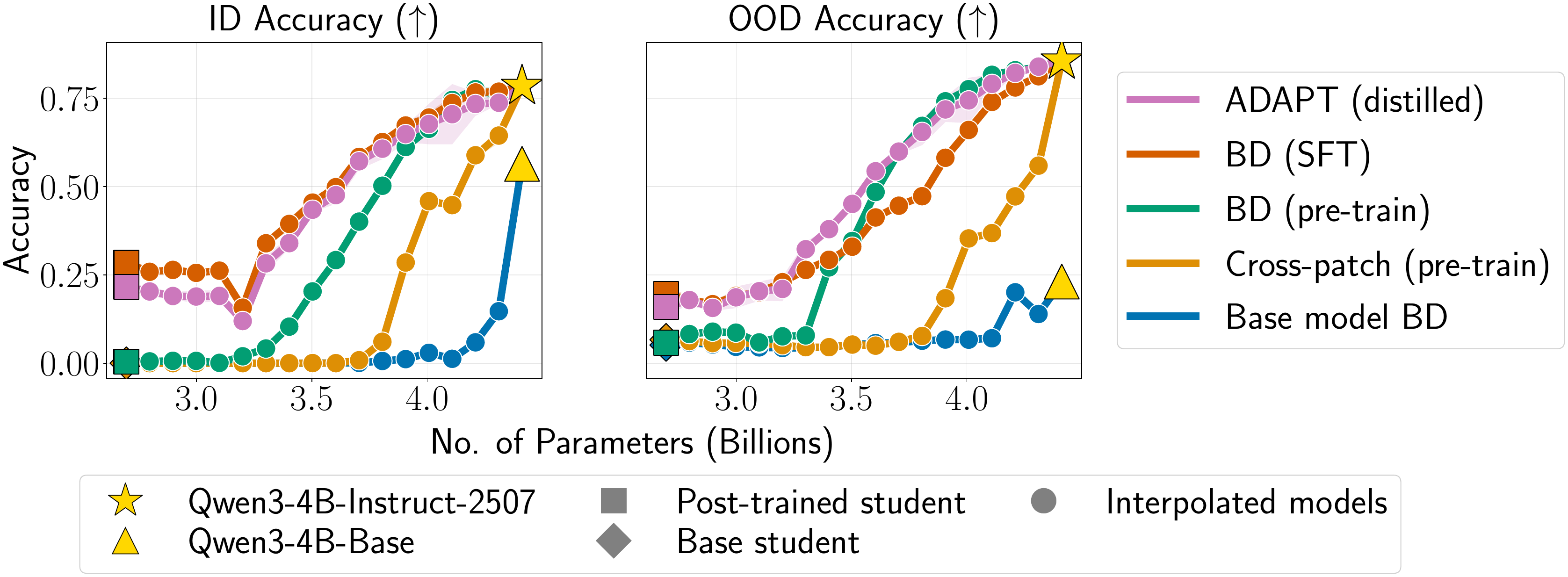}
    \caption{\textbf{\sys improves generation performance over boomerang distillation.} For generation tasks that are both in-domain (math reasoning) and out-of-domain (instruction-following and general knowledge) relative to our SFT dataset, SFT training is necessary to recover generation performance, especially for smaller models. For OOD tasks, two-phase distillation stabilizes performance compared to SFT-only training. We observe similar results for Qwen3-14B, Olmo-3-7B-Instruct, and Llama-3.1-8B-Instruct in Appendix~\ref{appendix:aggregated_results_all}.}
    \vspace{-0.4cm}
    \label{fig:2_stage_distillation}
\end{figure*}

Although the two-phase distillation produces strong interpolated model families for generation and reasoning tasks (Figure~\ref{fig:2_stage_distillation}), applying it independently to each post-trained variant still requires a separate distillation run. 
We propose \textbf{weight-delta initialization}, a weight-arithmetic procedure that transfers a base-model distillation run to a post-trained variant without additional training, approximately constructing distilled post-trained student models and amortizing the cost of constructing multiple post-trained model families. 

Let $\vtheta_{S,\mathrm{init}}^{\mathrm{base}}$ denote the student initialized from the base teacher, and let $\vtheta_{S,\mathrm{distill}}^{\mathrm{base}}$ denote the corresponding student after two-phase distillation. 
We define the distillation delta as:
\[
\Delta_{S,\mathrm{distill}}^{\mathrm{base}}
=
\vtheta_{S,\mathrm{distill}}^{\mathrm{base}}
-
\vtheta_{S,\mathrm{init}}^{\mathrm{base}} .
\]

Given a student $\vtheta_{S,\mathrm{init}}^{\mathrm{PT}}$ initialized from a post-trained teacher using the same layer-selection rule and parameter shapes as $\vtheta_{S,\mathrm{init}}^{\mathrm{base}}$, we construct the weight-delta-initialized student $\vtheta_{S,\Delta}^{\mathrm{PT}}$ as:
\[
\vtheta_{S,\Delta}^{\mathrm{PT}}
=
\vtheta_{S,\mathrm{init}}^{\mathrm{PT}}
+
\Delta_{S,\mathrm{distill}}^{\mathrm{base}} .
\]

The resulting model $\vtheta_{S,\Delta}^{\mathrm{PT}}$ is intended to approximate the directly distilled post-trained student $\vtheta_{S,\mathrm{distill}}^{\mathrm{PT}}$ in downstream performance, while avoiding the cost of an additional distillation run.

\section{Experiments}\label{sec:boomerang-pt-main-results}
Here we study the creation of post-trained model families using \sys. 
We show the benefits of \sys over strong baselines (Sections \ref{sec:two_phase_distillation_results} and \ref{sec:weight-delta-results}).
Then, we show \sys offers a better performance--compute trade-off than any single model (Section \ref{sec:adaptive_model_size}). 
Finally, we analyze the loss landscape to understand weight-delta initialization (Section \ref{sec:smooth-weight-interpolation}).

\subsection{Setup}
\label{boomerang_pt_setup} 

In our experiments, we primarily use Qwen3-4B-Instruct-2507~\citep{yang2025qwen3technicalreport} as the teacher model.
We also study Qwen3-4B-Thinking-2507 and Qwen3-4B in Section~\ref{sec:weight-delta-results}. 
We initialize the student models with 2.7B inference-time parameters by removing every other layer from the teacher (excluding the final layer). We train on a total budget of 1B tokens. For more training details, see Appendices \ref{appendix:training_implementation} and \ref{appendix:hyperparameters}.
 We also perform the same experiments with Qwen3-14B, Olmo~\citep{olmo2025olmo3}, and Llama~\citep{grattafiori2024llama3herdmodels} families in Appendix~\ref{appendix:aggregated_results_all}.
We train on the deduplicated Pile \citep{gao2020pile800gbdatasetdiverse} during the pre-training stage, and the math split of the Llama Nemotron Post-Training Dataset \citep{bercovich2025llamanemotronefficientreasoningmodels} during the SFT stage. 
We primarily evaluate our models on a set of five generation benchmarks measuring instruction-following and mathematical reasoning. We also apply our method to coding tasks and report the results in Appendix~\ref{appendix:aggregated_results_coding}. 
For more implementation details, see Appendix \ref{appendix:eval_implementation}. We include error bars of one standard deviation in the figures. 

\subsection{Two-Phase Distillation for Reasoning}
\label{sec:two_phase_distillation_results} 

We begin by comparing \sys~(distilled) to boomerang distillation (BD), showing that the two-phase distillation pipeline is key to model size interpolation for generation and reasoning tasks, yielding substantially stronger interpolation performance. 

\paragraph{Setup.}
We compare five setups under an equal training budget, all evaluated on three in-domain (ID) math reasoning tasks and two out-of-domain (OOD) tasks testing instruction-following and general understanding. (1) \textit{\sys~(distilled)}: Our proposed two-phase distillation setup. (2) \textit{BD (pre-train)}: Directly applying BD to the post-trained teacher. (3) \textit{BD (SFT)}: BD on post-trained model but with SFT instead of pre-training data in distillation stage. (4) \textit{Cross-patch (pre-train)}: Patching post-trained teacher directly onto boomerang-distilled base student to study whether post-trained teacher capabilities can be transferred through patching alone. (5) \textit{Base model BD}: BD on base model as a reference. 

\begin{figure*}[!tb]
    \centering
    \includegraphics[width=0.95\linewidth]{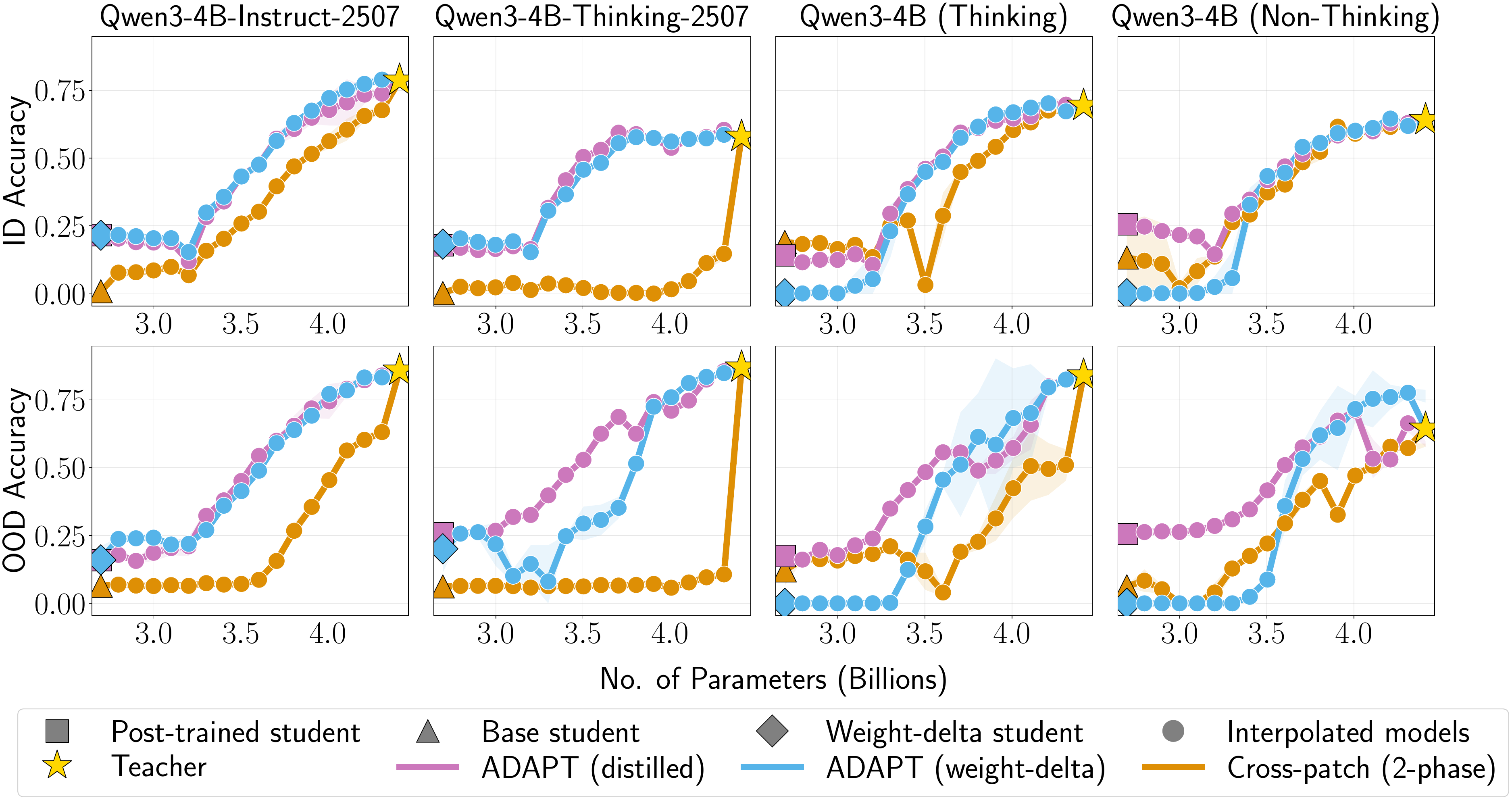}
    \caption{
    \textbf{\sys~(weight-delta) enables reusable base model distillation across post-trained variants.}
    \sys~(weight-delta) matches \sys~(distilled) on ID tasks and remains competitive on OOD tasks, showing that one base model distillation run can transfer to multiple post-trained variants without additional training. \sys~(weight-delta) outperforms baseline cross-patch (2-phase) across all variants. Degradation in thinking-style settings is mainly due to malformed \texttt{<think>} formatting, rather than broad loss of generation quality.
    }
    \vspace{-0.3cm}
    \label{fig:weight-delta-results} 
\end{figure*}

\paragraph{Results.} Figure~\ref{fig:2_stage_distillation} shows that \sys~(distilled) achieves the strongest interpolation performance on both ID and OOD tasks by combining pre-training alignment with SFT capability recovery. It also achieves the best teacher--student alignment, as measured via last-layer activation cosine similarity (Appendix~\ref{appendix:cos_sim_analysis}). We observe similar trends for Qwen3-14B, Olmo, and Llama in Appendix~\ref{appendix:aggregated_results_all} and report additional ablation results in Appendix~\ref{appendix:ablation-2-phase-distill}.

BD~(pre-train) remains weak at small and medium model sizes, improving only as model size increases. BD~(SFT) does not generalize as well to OOD and classification tasks (Appendix~\ref{appendix:classification_performance}), despite comparable performance to \sys~(distilled) at small model sizes on ID tasks. Base model BD collapses completely, indicating that base model interpolation does not directly transfer to generation and reasoning tasks. 

Interestingly, despite performing significantly worse than other setups, cross-patch (pre-train) recovers nontrivial performance as more post-trained teacher layers are inserted, outperforming base model BD. This suggests that base and post-trained models remain partially compatible, motivating our weight-delta transfer approach in Section~\ref{sec:weight-delta-results} and smooth weight interpolation experiments in Section~\ref{sec:smooth-weight-interpolation}.

\subsection{Weight-Delta Initialization Enables Reusable Base Distillation} 
\label{sec:weight-delta-results}

We demonstrate that \sys~(weight-delta) can reuse a single base-model distillation run across multiple post-trained variants, recovering much of the interpolation behavior of directly distilled post-trained students without additional distillation.

\paragraph{Setup.}
We construct the interpolated models for Qwen3-4B-Instruct-2507, Qwen3-4B-Thinking-2507, and Qwen3-4B using \sys~(weight-delta). Specifically, we first perform two-phase distillation on the Qwen3-4B-Base student to obtain the base distillation delta, then add this delta to students initialized from each post-trained variant. We compare \sys~(weight-delta) against two setups: (1) \textit{\sys~(distilled)}, the higher-compute upper bound that runs a separate two-phase distillation for each post-trained student. (2) \textit{Cross-patch (2-phase)}, which patches post-trained teacher layers directly into the two-phase-distilled base student without applying the weight delta. Cross-patch (2-phase) serves as a natural, compute-matched baseline: like \sys~(weight-delta), it produces a family of interpolated post-trained models from a single base-model distillation run. All methods use the same student architecture, layer-selection rule, patching order, and evaluation protocol.

\paragraph{Results.}
Figure~\ref{fig:weight-delta-results} shows that \sys~(weight-delta) achieves interpolation performance comparable to \sys~(distilled) on ID tasks across the evaluated Qwen variants, while requiring no additional distillation on the post-trained teachers. 
On OOD tasks (IFEval and MMLU-Redux), \sys~(weight-delta) remains competitive with \sys~(distilled), with slight degradation concentrated in thinking-style models at smaller and intermediate sizes. Inspecting outputs shows this is largely a parsing artifact rather than a loss of generation quality: weight-delta models are initialized from a base distillation run without thinking-style \texttt{<think>} tags and can produce malformed delimiters, which IFEval penalizes since it parses only the response after \texttt{</think>}. Performance remains strong on MMLU-Redux, which does not depend on \texttt{</think>} parsing, confirming the issue is specific to strict thinking-tag parsing rather than a broad degradation in capability. \sys~(weight-delta) also substantially outperforms cross-patch (2-phase) across post-trained variants, suggesting that the base model distillation delta transfers useful instruction-following capability and enables stronger interpolation than patching alone. 

Results for Olmo and Llama models (Appendix~\ref{appendix:aggregated_results_all}) suggest that \sys~(weight-delta) generally falls between \sys~(distilled) and cross-patch (2-phase), and is comparable to \sys~(distilled) in many cases. It also achieves better interpolation performance than cross-patch (2-phase) in all ID settings, making it a cheaper but still strong-performing alternative to separately distilling each post-trained variant. We provide preliminary explanation for when weight-delta initialization works best in Appendix~\ref{app:weight_delta_alignment}.

\begin{figure*}[!tbh]
    \centering
    \includegraphics[width=1\linewidth]{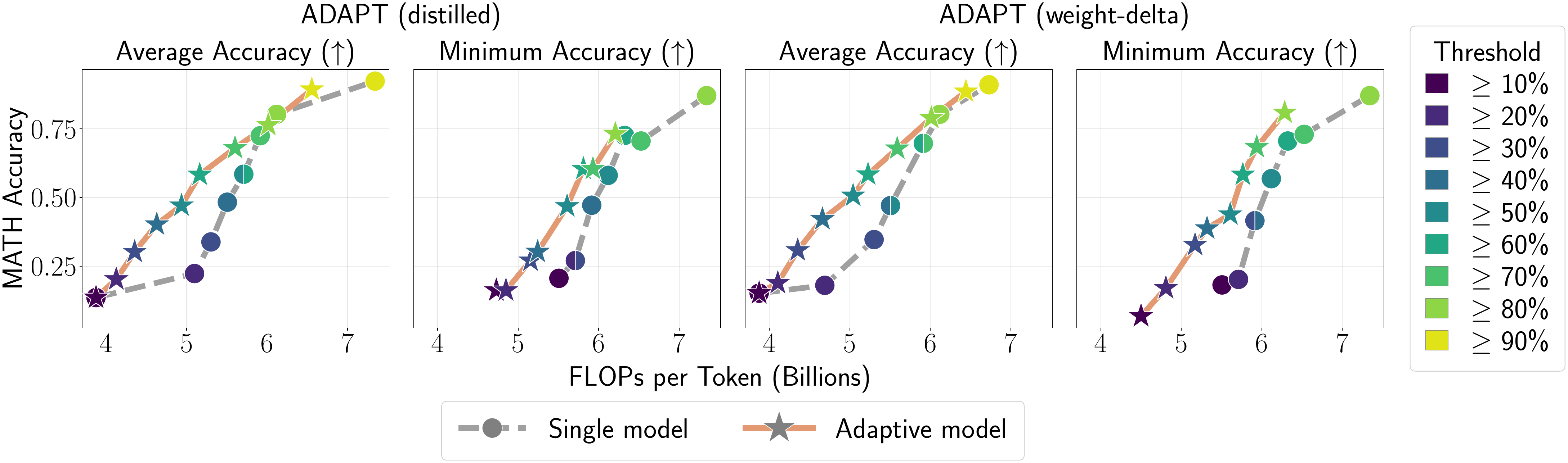}
    \vspace{-0.6cm}
    \caption{\textbf{\sys enables adaptive model-size selection based on input difficulty for more compute-efficient inference.} By producing a continuum of interpolated models with smoothly varying performance, \sys allows for more fine-grained matching between model capacity and task difficulty. This approach achieves a Pareto-optimal trade-off between average and worst-case accuracy and computational efficiency. } 
    \vspace{-0.3cm}
    \label{fig:adaptive_model_size} 
\end{figure*}

We also report results for single-phase weight-delta initializations in Appendix~\ref{appendix:weight-delta-1-phase}, where the transferred distillation delta is computed from a base student distilled on the pre-training phase or SFT phase only. This setting still demonstrates some cross-model distillation transfer, but generally underperforms the two-phase weight-delta initialization used in \sys. We experiment with continued training from the weight-delta initialized model, $\vtheta_{S,\Delta}^{\mathrm{PT}}$, but find that it does not improve interpolation performance (Appendix~\ref{appendix:weight-delta-continued-training}). We include further ablation results on post-trained model deltas and SFT phase loss components in Appendices~\ref{appendix:ablation-post-trained-deltas} and \ref{appendix:ablation-sft-loss-term}. Finally, we compare both variants of \sys to popular layer pruning methods in Appendix~\ref{appendix:comparison_to_pruning}.\looseness-1

\subsection{Adaptive Model-Size Selection} 

\label{sec:adaptive_model_size}

In this section, we show that \sys achieves a Pareto-optimal trade-off between downstream performance and computational efficiency by dynamically choosing from the suite of interpolated models at inference time based on input difficulty. This improves compute utilization while preserving performance, satisfying a practical requirement for deployment. 

\paragraph{Setup.}
We use the MATH dataset \citep{hendrycks2021measuringmathematicalproblemsolving}, which contains competition-style problems with varying difficulty levels. We estimate each problem's difficulty using a single forward pass with the teacher model, Qwen3-4B-Instruct-2507 (Appendix~\ref{appendix:input_diff_classification}).
For each interpolated model, we calculate its accuracy on each of the difficulty levels on a training split. Given a target accuracy threshold, we route questions of each difficulty level to the smallest model with accuracy above the threshold.

We consider both average accuracy and minimum accuracy. Average accuracy measures a model's example-weighted mean accuracy across all the difficulty levels, allowing it to undershoot the threshold on harder levels. Minimum accuracy requires every difficulty level to individually exceed the threshold, yielding a more conservative policy. 
Sweeping over the threshold produces routing policies with different accuracy--compute trade-offs, which we compare against fixed single-model baselines.

\paragraph{Results.} 
Figure \ref{fig:adaptive_model_size} shows that the adaptive model-size selection approach outperforms the single-model baseline, tracing out a superior Pareto frontier across both average and minimum accuracy metrics for both \sys~(distilled) and \sys~(weight-delta). 
This improvement arises from a more fine-grained matching between model capability and question difficulty, enabled by the suite of models with smoothly increasing performance. Therefore, by assigning smaller models to easier questions and reserving larger models for the harder ones, \sys offers a superior performance--compute trade-off over using a single model. 

We report the results for Olmo and Llama models in Appendix~\ref{appendix:apative_model_size_olmo_llama}. We also notice that smaller models tend to generate more tokens overall. We provide further discussion of this in Appendix~\ref{appendix:total_token_count}.

\subsection{Size-Agnostic Smooth Weight Interpolation}
\label{sec:smooth-weight-interpolation} 

\begin{figure}[!tb]
    \centering
    \includegraphics[width=\linewidth]{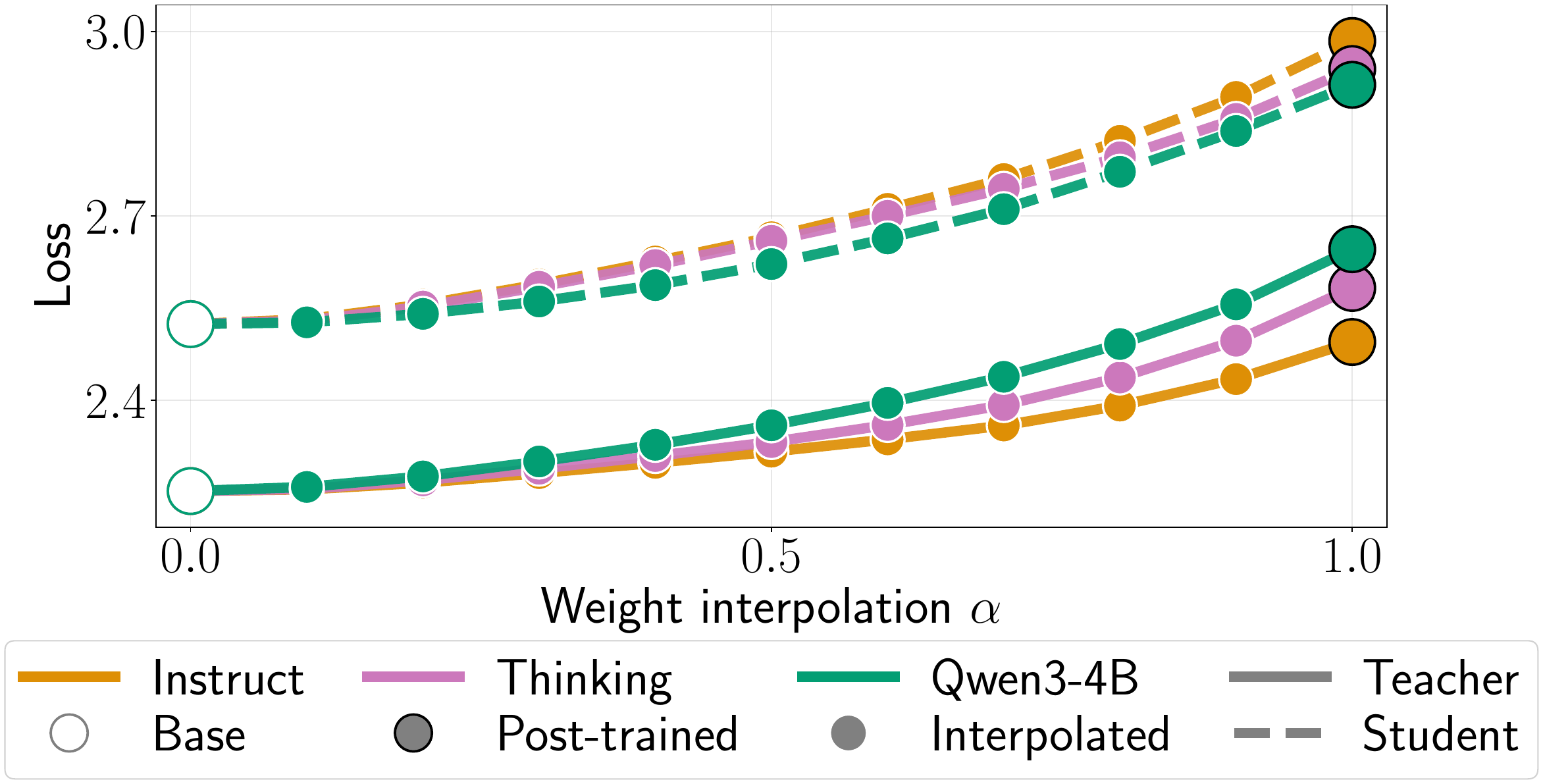}
    \vspace{-0.57cm}
    \caption{\textbf{Smooth weight interpolation is preserved after distillation.} Across post-trained variants, both teacher and distilled student models exhibit smooth interpolation paths between base and post-trained endpoints, suggesting that the relevant weight-space structure is preserved after distillation.}
    \vspace{-0.3cm}
    \label{fig:lmc}
\end{figure}

The weight-delta initialization results in Section~\ref{sec:weight-delta-results} suggest that base model distillation remains compatible with its post-trained variants. To better understand this, we study whether interpolation between base and post-trained models is preserved after distillation. 
We call this property \textbf{size-agnostic smooth weight interpolation}: 
the low-loss linear path between a base model and its post-trained variant is preserved in the corresponding student models after distillation.
For models $\vtheta^{(a)}$ and $\vtheta^{(b)}$, we evaluate models $\vtheta(\alpha) = (1-\alpha)\vtheta^{(a)} + \alpha \vtheta^{(b)}$ where $\alpha \in [0,1]$.
\looseness-1

\begin{figure}[!tb]
    \centering
    \includegraphics[width=\linewidth]{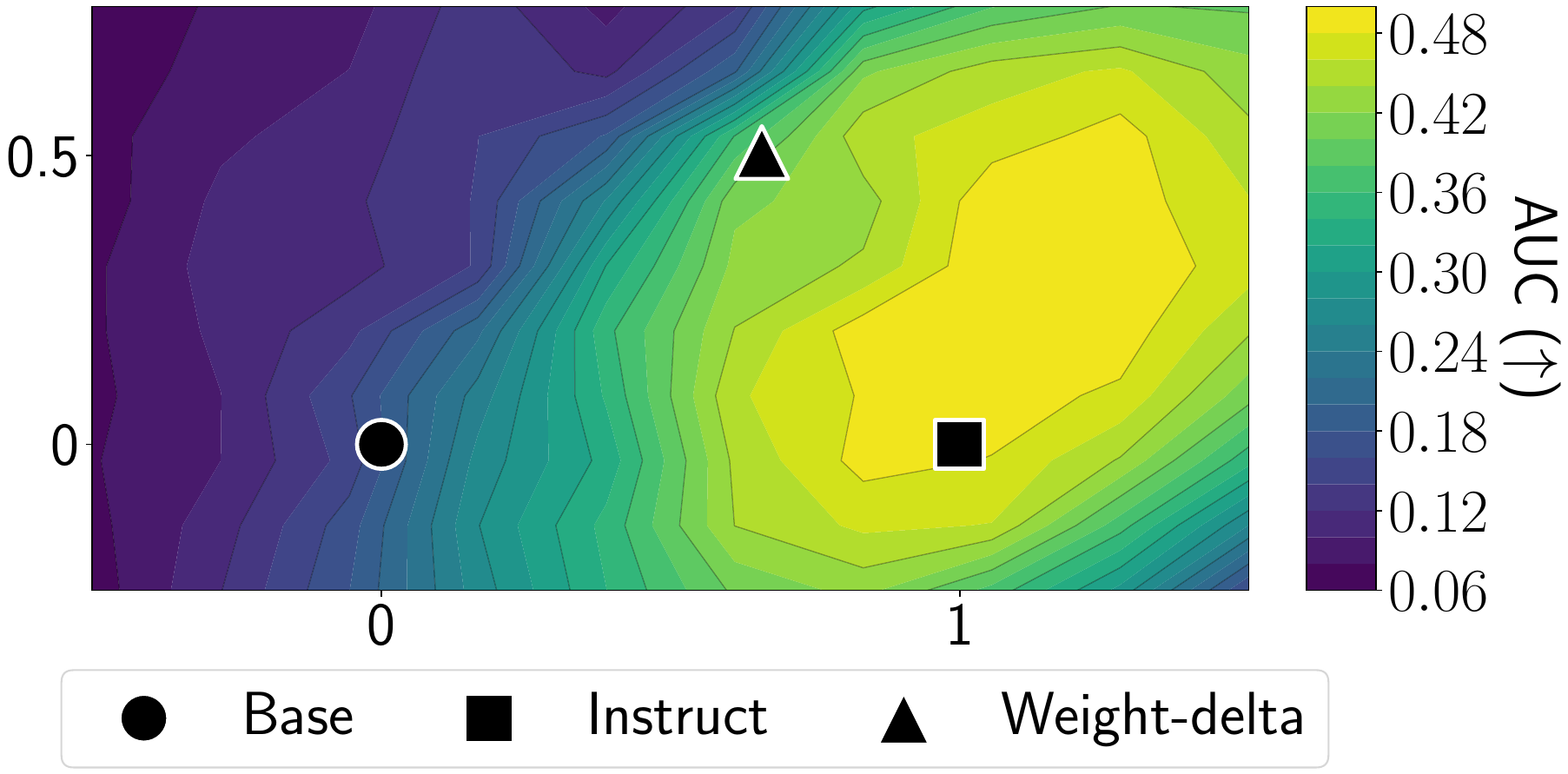}
    \caption{
    \textbf{Weight-delta initialization produces a student in a connected region of high generation performance.} We report AUC for generation accuracy between student and teacher models on the plane intersecting base, instruct, and weight-delta students. The weight-delta student has improved AUC compared to the base student and lies in a high-performance region connected to both the base and instruct students.}
    \vspace{-0.2cm}
    \label{fig:loss_landscape}
\end{figure}

Figure~\ref{fig:lmc} shows that interpolation paths between base and post-trained models do not exhibit sharp loss barriers for either teachers or students. The loss increases toward the post-trained endpoints because they are evaluated on the Pile, which is better matched to base models than post-trained variants. Importantly, students retain similar interpolation behavior after distillation, suggesting that distillation preserves much of the weight-space structure connecting base and post-trained models. We observe similar trends for generation accuracy and SFT loss in Appendix~\ref{appendix:additional_lmc}.
We further examine this geometry using a two-dimensional generation-accuracy AUC landscape over the plane spanned by the base, instruct, and weight-delta students. 
Here, AUC denotes the area under the generation-accuracy curve, measured between the smaller student and the corresponding teacher model.
Figure~\ref{fig:loss_landscape} shows that the weight-delta initialized student lies in a high-performing region connected to both the base and distilled instruct students.
These results suggest that distillation preserves smooth weight interpolation across model sizes, suggesting a possible connection to linear-mode connectivity \citep{garipov2018losssurfacesmodeconnectivity}. \looseness-1

\section{Conclusion}
\label{sec:discussion}

We introduce \sys, a framework for amortizing distillation across both the size and post-training-variant axes of a model family. \sys substantially outperforms boomerang distillation baselines on reasoning and instruction-following benchmarks under matched compute, and the resulting continuum of interpolated models enables adaptive model-size selection at inference time. Crucially, we find that smooth linear weight interpolation between base and post-trained models is preserved across model sizes after distillation, providing a plausible mechanism for the success of weight-arithmetic transfer in our setting. This connection between size interpolation and linear-mode connectivity points to a broader principle---that distillation can be reused across related models, not just sizes---and we view its further investigation as a promising direction. \looseness-1

\clearpage
\section*{Limitations}

 \sys is effective across the model families and benchmarks we study, but several aspects warrant further investigation. We highlight open questions about the underlying mechanism, assumptions implicit in our procedure, and the empirical scope of our evaluation.

 \paragraph{Interpolation dips at specific sizes.} As also observed by \citet{kangaslahti2026boomerangdistillationenableszeroshot}, some interpolation curves show sharp drops at particular intermediate sizes. A plausible explanation is that the inserted teacher layers are locally misaligned with the surrounding student layers at those configurations, but the precise mechanism remains unclear. Characterizing and predicting these failure modes remain open directions. We also observe that the weight-delta transfer is not uniformly reliable across Olmo and Llama (Figures \ref{fig:weight_delta_results_olmo} and \ref{fig:weight_delta_results_llama}), especially in some OOD settings and intermediate sizes. This suggests that weight-delta effectiveness depends on model-family compatibility and post-training details, and should be validated beyond the settings studied here.\looseness-1

\paragraph{Model and data scope.} Our experiments cover three model families (Qwen, Olmo, Llama) at the 4B-14B scale and use the math and coding splits of the Llama Nemotron Post-Training Dataset for the SFT phase. It remains to be verified whether \sys retains its size-amortization benefits at larger model scales and across broader instruction-tuning mixtures (e.g., multilingual).

\paragraph{Weight-delta assumptions.} Weight-delta initialization requires access to the base model used to produce each post-trained variant. For some open-weight releases (and most closed-weight releases) the corresponding base checkpoint is unavailable or not exactly identifiable, restricting which model families \sys (weight-delta) can target without additional approximation.

\paragraph{Adaptive inference depends on a difficulty signal.} The compute--accuracy gains from adaptive model-size selection (Section~\ref{sec:adaptive_model_size}) rely on a per-input difficulty estimate. 
Our MATH experiments use teacher-predicted difficulty, which is cheap but task-specific; we have not characterized how well this approach generalizes to settings where difficulty is not well-defined or where a separate difficulty predictor would be expensive to obtain.

\section*{Ethical Considerations}

Our experiments use publicly available datasets and evaluation benchmarks, primarily focused on mathematical reasoning, instruction following, and general model evaluation. We also use publicly available model checkpoints as teachers and base models. We do not collect new human-subject data. However, because pre-training and post-training datasets may contain web-derived text, they may inherit biases, harmful content, or privacy concerns from their source corpora. Similarly, the models used in this work may inherit biases or unsafe behaviors from their original training pipelines or the datasets we use. Our work does not attempt to remove these risks, and downstream users should evaluate generated outputs for safety, bias, and reliability before deployment.

\section*{Reproducibility Statement}

We implement all our experiments using PyTorch~\citep{paszke2019pytorchimperativestylehighperformance} and HuggingFace trans-
formers~\citep{wolf2020huggingfacestransformersstateoftheartnatural} packages. We also experiment with public models available on Hug-
gingFace Hub. We provide our code and models at \url{https://github.com/dcml-lab/ADAPT}. 

\section*{Acknowledgments}

David Alvarez-Melis, Sara Kangaslahti, Jonathan Geuter, and Nihal V. Nayak acknowledge support from
the National Science Foundation Graduate Research Fellowship (Grant No. DGE 2140743), the
Kempner Institute, FAS Dean's Competitive Fund for Promising Scholarship, Aramont Fellowship
Fund, and the NSF AI-SDM Institute (Grant No. IIS-2229881). Francesco Locatello's contribution
to this research was funded in part by the Austrian Science Fund (FWF) 10.55776/COE12.

\bibliography{custom}

\clearpage

\appendix

\section{Training Implementation Details}
\label{appendix:training_implementation}

We use a mixture of non-thinking and thinking models in this paper. Non-thinking models include Qwen3-4B-Instruct-2507, Qwen3-4B, Olmo-3-7B-Instruct, Llama-3.1-8B-Instruct, and Qwen3-14B. Thinking models include Qwen3-4B-Thinking-2507, Qwen3-4B, Olmo-3-7B-Think, and Qwen3-14B. Note that Qwen3-4B and Qwen3-14B can toggle between thinking and non-thinking modes. 

For model training, we use a total token budget of 1B for the Qwen 4B-sized models. For the larger models, we scale up the total budget proportionally, to 2B and 4B tokens respectively \citep{hoffmann2022trainingcomputeoptimallargelanguage}. Across different configurations of the same model, we choose hyperparameters such that the number of tokens per update remains approximately constant between the pre-training phase and SFT phase. The hyperparameters reported in Table~\ref{tab:training_params_all} correspond to using the deduplicated Pile \citep{gao2020pile800gbdatasetdiverse} for pre-training and the Llama Nemotron Post-Training Dataset \citep{bercovich2025llamanemotronefficientreasoningmodels} for the SFT phase. For Nemotron, we primarily use the math split for all models (see Appendix~\ref{appendix:aggregated_results_coding} for coding tasks). For non-thinking models, we use examples with non-thinking ground truths (i.e., responses not enclosed in \texttt{<think>} and \texttt{</think>}), while for thinking models we use examples with thinking ground truths (version 1.1).

Each 4B model training run takes approximately 12 hours on 4 NVIDIA H100 GPUs. The Olmo and Llama models are trained using 4 NVIDIA H200 GPUs, each taking approximately 24 hours. Training Qwen3-14B takes approximately 48 hours on 16 H200 GPUs. We will release the distilled models under an Apache 2.0 license.

\section{Hyperparameters}
\label{appendix:hyperparameters}

We choose the majority of the hyperparameters (Table~\ref{tab:hyperparams}) following \citet{kangaslahti2026boomerangdistillationenableszeroshot}. We determine the KL and cosine loss weights such that cross entropy, KL, and cosine loss are approximately equal in magnitude early in training. M denotes the number of student layers. When doing two-phase distillation, we use a single LR scheduler across the two phases. 

\begin{table}[!tbh]
\begin{center}
\begin{tabular}{cc}
\toprule
\multicolumn{1}{c}{\bf Hyperparameters}  &\multicolumn{1}{c}{\bf Values} \\
\midrule
Learning rate & 3e-4 \\
LR scheduler & cosine \\
Warmup ratio & 0.01 \\
Optimizer & AdamW \\
Adam betas & (0.9, 0.95) \\
Adam epsilon & 1e-8 \\ 
Weight decay & 0.1 \\
Max gradient norm & 1.0 \\
Mixed precision & bf16 \\
KL weight $\lambda_{KL}$ & 0.1 \\ 
Cosine weight $\lambda_{cos}$ & 10.0 / (M+1) \\
\bottomrule
\end{tabular}
\end{center}
\vspace{-0.3cm}
\caption{\textbf{Hyperparameters for distillation training.}}
\vspace{-0.4cm}
\label{tab:hyperparams}
\end{table}

\section{Evaluation Implementation Details}
\label{appendix:eval_implementation}

For generation, we use a custom evaluation pipeline for more fine-grained control over chat template and system prompt settings. Table~\ref{tab:system_prompts} shows the system prompts used for each benchmark. We report the average accuracy on 5 benchmarks: AIME \citep{aimo_validation_aime}, GSM8K \citep{cobbe2021trainingverifierssolvemath}, IFEval \citep{zhou2023instructionfollowingevaluationlargelanguage}, MATH500 \citep{hendrycks2021measuringmathematicalproblemsolving}, MMLU-Redux \citep{gema2025mmlu}. We consider AIME, GSM8K, and MATH500 to be in-domain (ID) math reasoning tasks and IFEval and MMLU-Redux to be out-of-domain (OOD) tasks. We use a separate set of ID benchmarks described in Appendix~\ref{appendix:aggregated_results_coding} for coding tasks. For Section~\ref{sec:adaptive_model_size}, we use the MATH dataset \citep{hendrycks2021measuringmathematicalproblemsolving}, which contains ground truth question difficulty labels. We elaborate on this further in Appendix~\ref{appendix:input_diff_classification}. 

For classification tasks, we use \texttt{lm-evaluation-harness}~\citep{eval-harness} and report the average accuracy on 10 benchmarks: ARC-easy and ARC-challenge \citep{clark2018thinksolvedquestionanswering}, BoolQ \citep{clark-etal-2019-boolq}, HellaSwag \citep{zellers-etal-2019-hellaswag}, MMLU \citep{hendrycks2021measuring}, OpenBookQA \citep{mihaylov-etal-2018-suit}, PIQA \citep{bisk2020piqa}, RACE \citep{lai-etal-2017-race}, RTE \citep{wang2018glue}, and WinoGrande \citep{sakaguchi2020winogrande}.

Table~\ref{tab:sampling_params} shows the sampling parameters used for non-thinking and thinking models, which follow the recommended values by \citet{yang2025qwen3technicalreport}. 

\begin{table}[!tbh]
\begin{center}
\begin{tabular}{ccc}
\toprule
\multicolumn{1}{c}{\bf Models}  &\multicolumn{1}{c}{\bf Non-thinking} &\multicolumn{1}{c}{\bf Thinking} \\
\midrule
Max output len. & 16384 & 32768 \\ 
Temperature & 0.7 & 0.6 \\
Top p & 0.8 & 0.95 \\ 
Top k & 20 & 20 \\
Presence penalty & 0.5 & 0.5 \\
Min. tokens & 32 & 32 \\
\bottomrule
\end{tabular}
\end{center}
\vspace{-0.2cm}
\caption{\textbf{Sampling parameters for non-thinking and thinking models.}}
\vspace{-0.5cm}
\label{tab:sampling_params}
\end{table}

\begin{table*}[!tbh]
\centering
\resizebox{0.94\textwidth}{!}{%
\begin{tabular}{ll ccc ccc}
\toprule
 &  & \multicolumn{3}{c}{\textbf{Pre-training Phase}} & \multicolumn{3}{c}{\textbf{SFT Phase}} \\
\cmidrule(lr){3-5} \cmidrule(lr){6-8}
\textbf{Model Type} & \textbf{Setup}
& \textbf{Steps} & \textbf{Effective BS} & \textbf{Seq. len.}
& \textbf{Steps} & \textbf{Effective BS} & \textbf{Seq. len.} \\
\midrule
\multirow{3}{*}{4B (Non-think)}
& Pile only       & 480 & 2048 & 1024 & --  & --   & -- \\
& Nemotron only   & --  & --   & --   & 585 & 4096 & 1024 \\
& Pile+Nemotron   & 240 & 2048 & 1024 & 293 & 4096 & 1024 \\
\midrule
\multirow{3}{*}{4B (Think)}
& Pile only       & 500 & 512  & 4096 & --  & --   & -- \\
& Nemotron only   & --  & --   & --   & 400 & 1024 & 4096 \\
& Pile+Nemotron   & 250 & 512  & 4096 & 200 & 1024 & 4096 \\
\midrule
\multirow{3}{*}{7B/8B (Non-think)}
& Pile only       & 960 & 2048 & 1024 & --   & --   & -- \\
& Nemotron only   & --  & --   & --   & 1170 & 4096 & 1024 \\
& Pile+Nemotron   & 480 & 2048 & 1024 & 585  & 4096 & 1024 \\
\midrule
\multirow{3}{*}{7B/8B (Think)}
& Pile only       & 1000 & 512 & 4096 & --   & --   & -- \\
& Nemotron only   & --  & --   & --   & 800 & 1024 & 4096 \\
& Pile+Nemotron   & 500 & 512 & 4096 & 400  & 1024 & 4096 \\
\midrule
\multirow{3}{*}{14B}
& Pile only       & 875 & 1024 & 4096 & --   & --   & -- \\
& Nemotron only   & --  & --   & --   & 1400 & 1024 & 4096 \\
& Pile+Nemotron   & 438 & 1024 & 4096 & 700  & 1024 & 4096 \\
\bottomrule
\end{tabular}%
}
\vspace{-0.15cm}
\caption{\textbf{Training hyperparameters across model families.} We report the number of steps, effective batch size, and sequence length for the pre-training and SFT phases across all distillation setups. 
}
\vspace{-0.1cm}
\label{tab:training_params_all}
\end{table*}

\begin{table*}[!tbh]
\centering
\renewcommand{\arraystretch}{1.2}
\begin{tabular}{l p{8.6cm}}
\toprule
\textbf{Benchmarks} & \textbf{System Prompts} \\
\midrule
AIME, GSM8K, MATH500, MATH 
& Please reason step by step, and put your final answer within \texttt{\textbackslash boxed\{\}}. \\
IFEval 
& None \\
MMLU-Redux 
& Please reason step by step, and select the answer from the given choices A, B, C, or D. Respond only with the letter of the correct answer. Put the index within \texttt{\textbackslash boxed\{\}}. \\
\bottomrule
\end{tabular}
\vspace{-0.2cm}
\caption{\textbf{System prompts for generation tasks.}}
\vspace{-0.51cm}
\label{tab:system_prompts}
\end{table*}

\section{Additional Evaluation Results for \sys (Distilled)}
We provide additional results for \sys~(distilled). We first study final-layer cosine similarity to the teacher in Appendix \ref{appendix:cos_sim_analysis}, then show results on classification tasks in Appendix \ref{appendix:classification_performance}.

\begin{figure}[!h]
    \centering
    \includegraphics[width=0.96\linewidth]{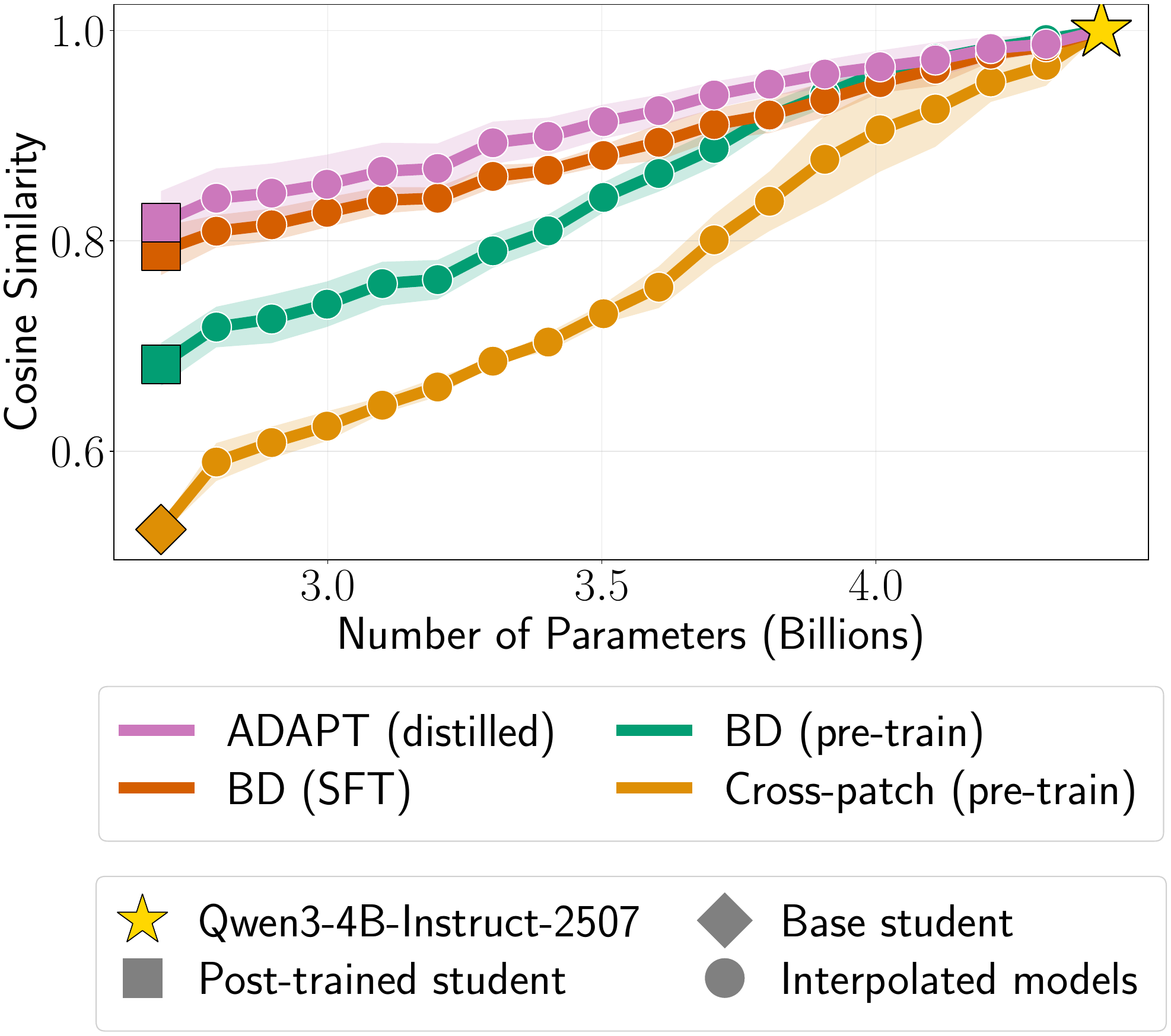}
    \vspace{-0.2cm}
    \caption{\textbf{Final layer activation cosine similarity between student and teacher models.} The two-phase distillation setup yields the best student-teacher alignment, further motivating our approach.}
    \vspace{-0.4cm}
    \label{fig:cos_sim}
\end{figure}

\begin{figure*}[!tbh]
    \centering
    \includegraphics[width=1\linewidth]{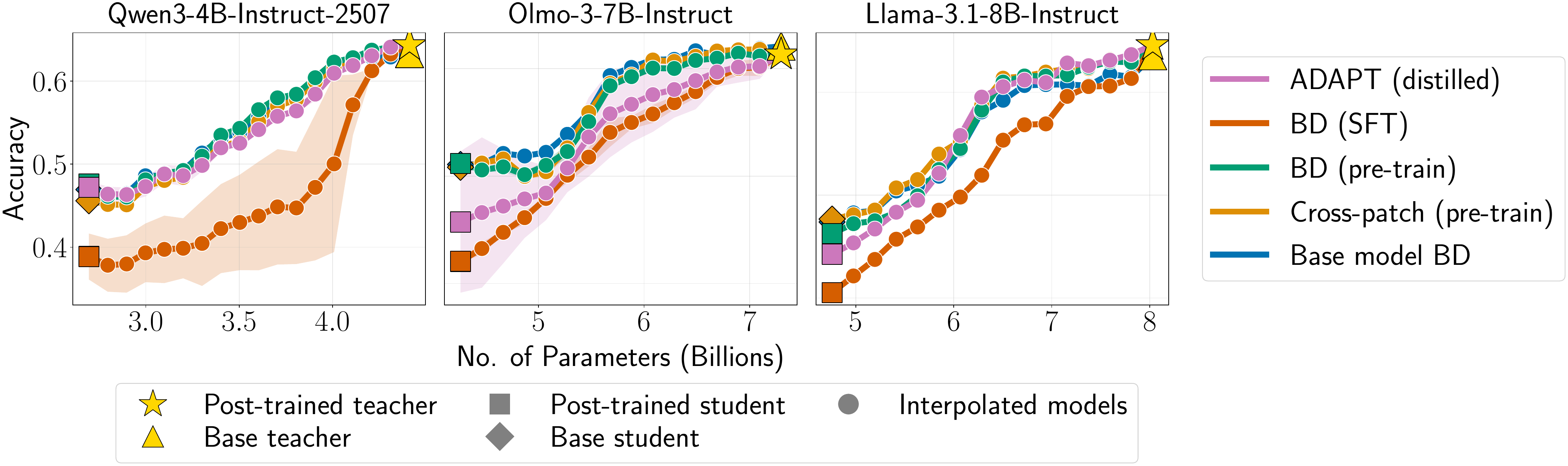}
    \caption{\textbf{Classification accuracy is relatively insensitive to the distillation strategy.}
    Across Qwen, Olmo, and Llama, most setups exhibit similar interpolation behavior on classification benchmarks. The main exception is BD~(SFT), which underperforms across much of the interpolation curve, suggesting that task-specific distillation alone can degrade broader alignment with the teacher.
    }
    \label{fig:classification_performance}
\end{figure*}

\begin{figure*}[!tb]
    \centering
    \includegraphics[width=0.83\linewidth]{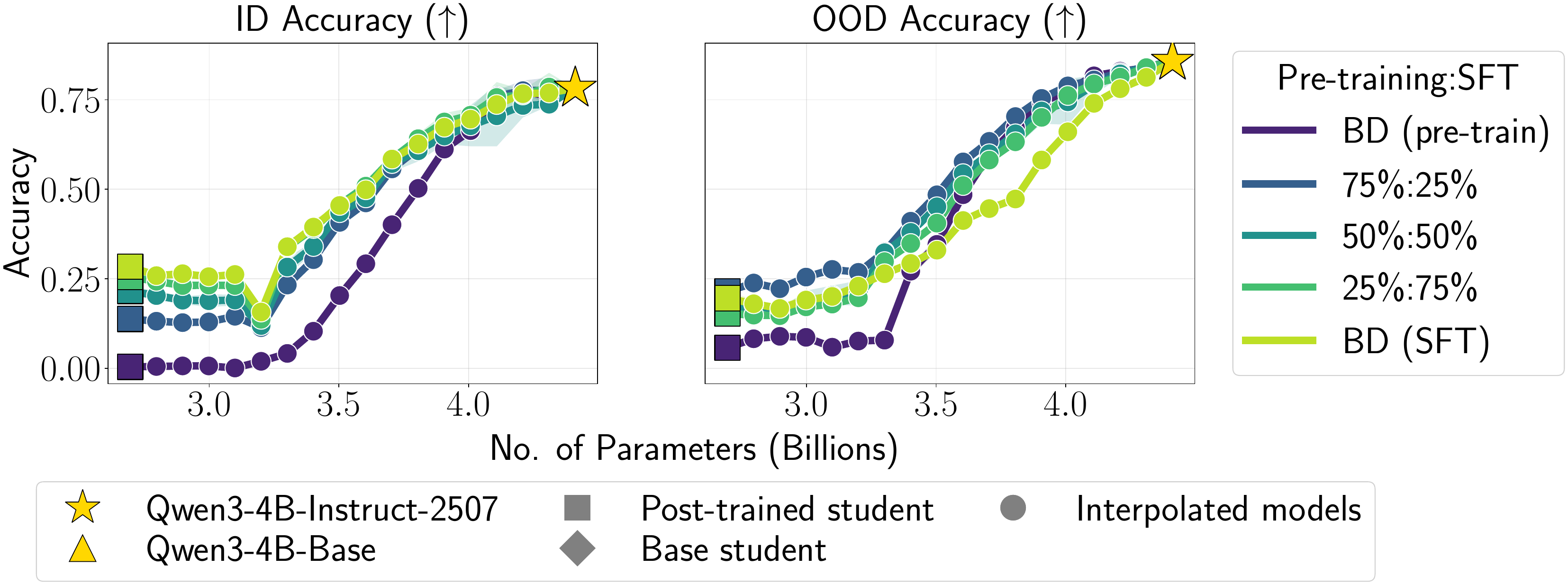}
    \caption{\textbf{Distilling with equal pre-training and SFT phase offers an appropriate tradeoff between ID and OOD performance.} Increasing the SFT proportion improves ID accuracy but degrades OOD accuracy, while increasing the pre-training proportion has the opposite effect; the 50:50 split balances the two. The single-phase endpoints are consistently dominated by the two-phase settings.}
    \label{fig:2_phase_ratio}
    \vspace{-0.3cm}
\end{figure*}

\subsection{Cosine Similarity Analysis}
\label{appendix:cos_sim_analysis} 

In this section, we report the final-layer activation cosine similarity between each interpolated model and the corresponding teacher for \sys~(distilled), BD~(pre-train), BD~(SFT), and cross-patch (pre-train). We compute activations by averaging over all downstream tasks.

As shown in Figure~\ref{fig:cos_sim}, \sys~(distilled) achieves the highest cosine similarity across nearly the entire interpolation curve. BD~(SFT) also maintains strong alignment, but is consistently below \sys~(distilled). Cross-patch (pre-train) has the weakest alignment, especially at smaller model sizes, indicating that patching alone does not sufficiently align the student with the post-trained teacher. These results provide additional motivation for the two-phase distillation approach.

\subsection{Classification Performance}
\label{appendix:classification_performance} 

Figure~\ref{fig:classification_performance} shows that classification benchmarks are relatively insensitive to the distillation strategy. Most variants achieve similar interpolation performance, indicating that classification tasks do not fully expose the alignment challenges that arise in post-trained generation. The main exception is the BD~(SFT) setup, which underperforms across much of the interpolation curve for all three model families, suggesting that task-specific distillation alone can degrade broader alignment with the teacher.

\section{Additional Ablations for \sys (Distilled)}
\label{appendix:ablation-2-phase-distill}

\subsection{Pre-training vs. SFT Phase Ratios}
\label{appendix:2_phase_ratio}

Here we study how sensitive \sys~(distilled) is to varying training budget between pre-training and SFT phase. All settings use the same total budget. 

In Figure~\ref{fig:2_phase_ratio}, we observe a consistent trade-off across the two-phase settings: increasing the SFT proportion improves ID accuracy but degrades OOD accuracy, while increasing the pre-training proportion has the opposite effect. The 50:50 split sits between these extremes, achieving strong ID accuracy without sacrificing OOD performance. We note that the three two-phase ratios perform comparably at larger model sizes, indicating that \sys is not overly sensitive to the exact split.

Finally, both single-phase endpoints are dominated by the two-phase settings, consistent with our finding in Section~\ref{sec:two_phase_distillation_results}. 

\begin{figure*}[!tb]
    \centering
    \includegraphics[width=0.8\linewidth]{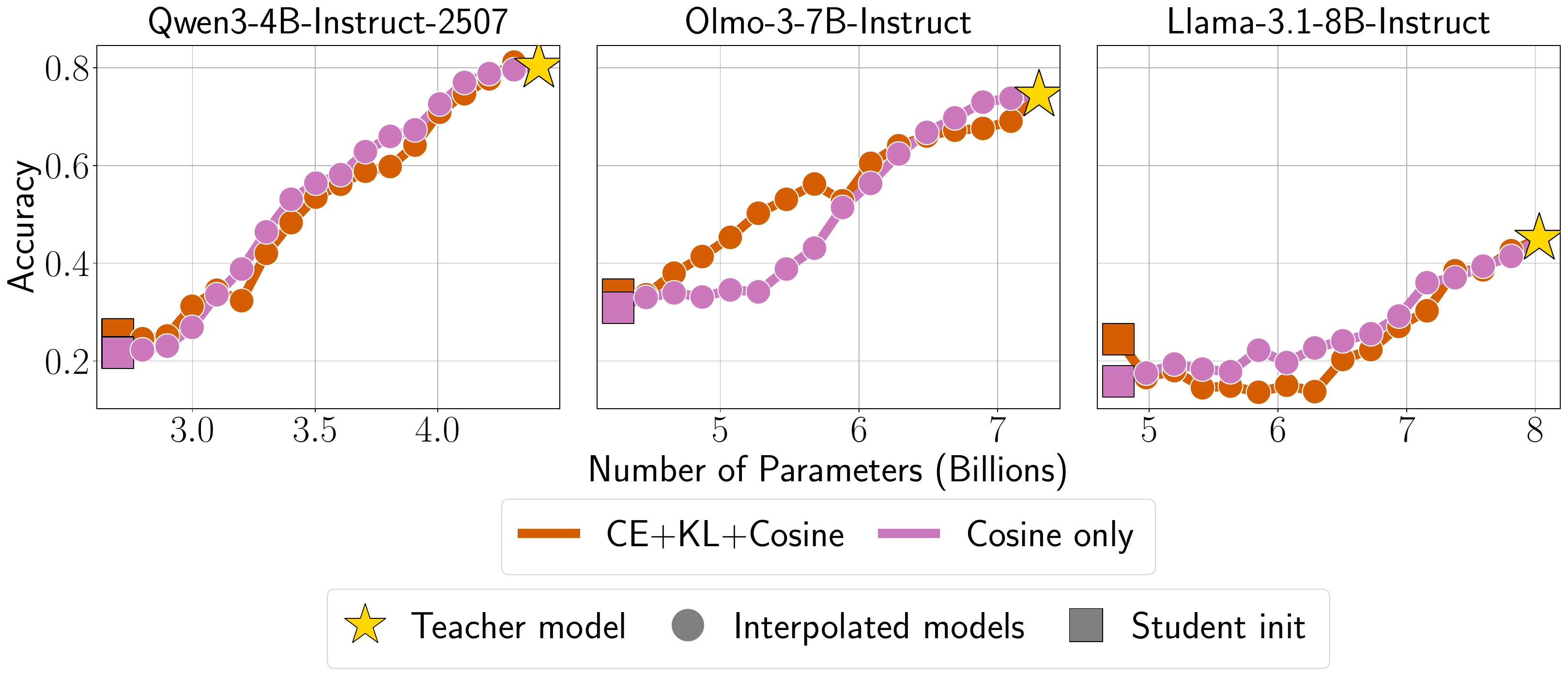}
    \vspace{-0.2cm}
    \caption{\textbf{Distilling with cosine loss only yields interpolation performance comparable to the full objective.} We compare training with cosine alignment loss only to training with all of the loss terms in Equation \ref{eq:overall-loss} and find that cosine only training has similar performance, indicating that the efficiency of interpolation training can potentially be improved by using only alignment loss.}
    \vspace{-0.1cm}
    \label{fig:cos_only}
\end{figure*}

\subsection{Distillation with Cosine Loss Only} 
\label{appendix:cos_only}

In Figure~\ref{fig:cos_only}, we show that distilling with cosine loss alone can yield performance comparable to the full objective combining cross-entropy, KL, and cosine losses. This suggests that intermediate representation alignment may be a key driver of interpolation behavior, and points to an orthogonal direction for improving distillation efficiency by simplifying the training objective.

\section{Additional Ablations for \sys (Weight-Delta)}

\subsection{Weight-Delta Transfer from Pre-training-only and SFT-only Deltas}
\label{appendix:weight-delta-1-phase}

\begin{figure*}[!tbh]
    \centering
    \includegraphics[width=0.99\linewidth]{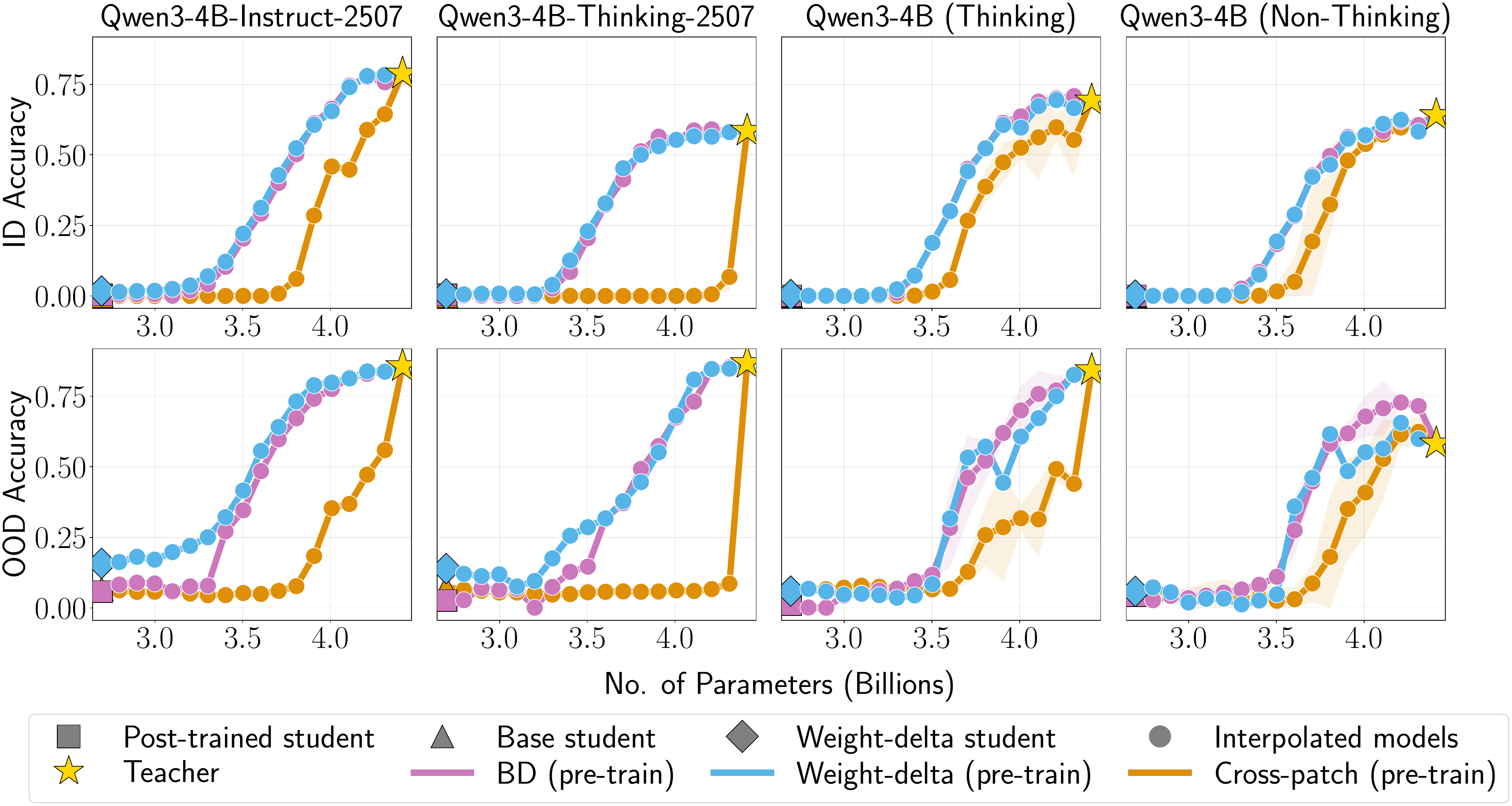}
    \vspace{-0.1cm}
    \caption{\textbf{Pre-training weight-delta initialization for post-trained Qwen models.}}
    \vspace{-0.3cm}
    \label{fig:1_phase_weight_delta_qwen_id}
\end{figure*}

\begin{figure*}[!tbh]
    \centering
    \hspace*{-0.3cm}\includegraphics[width=0.95\linewidth]{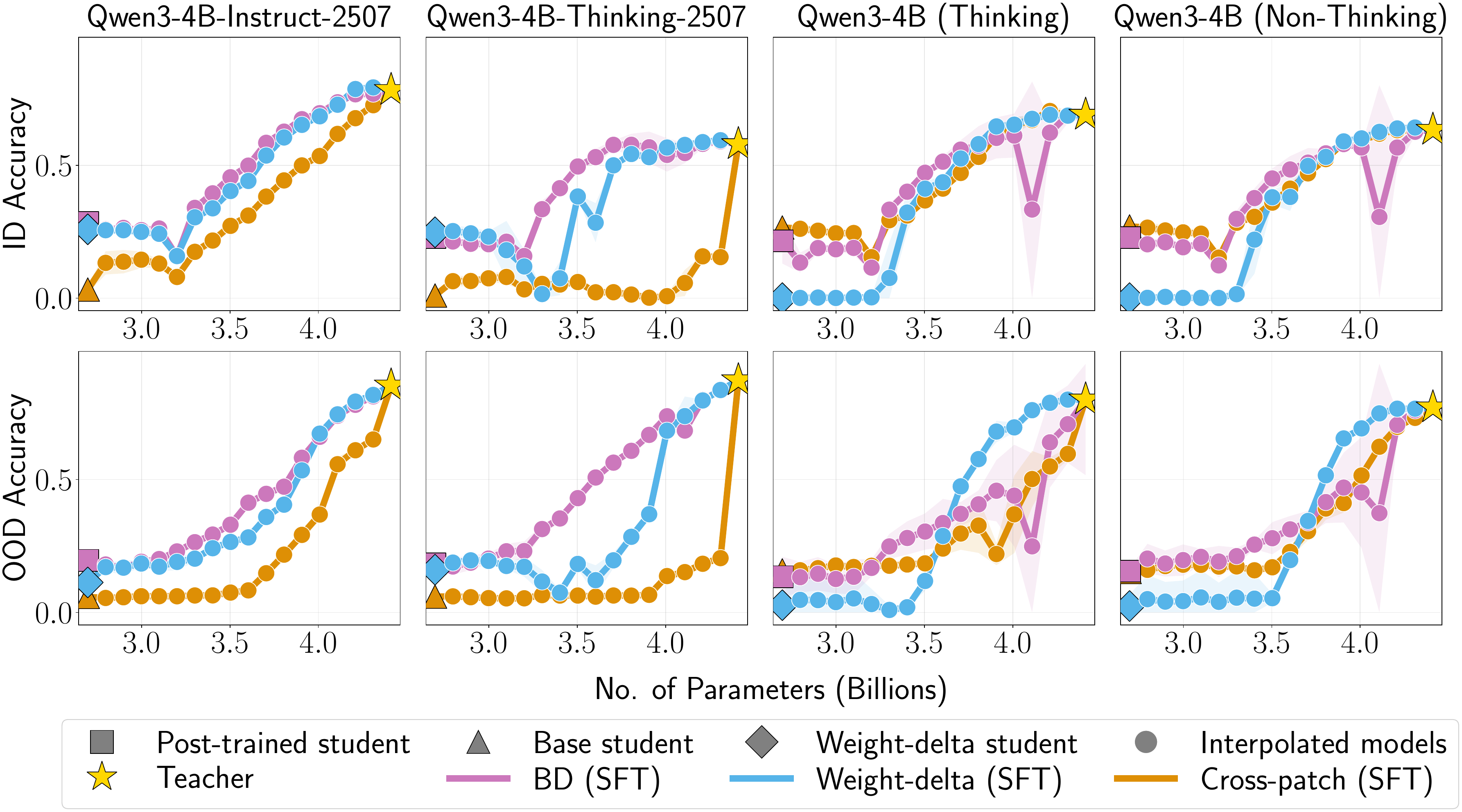}
    \vspace{-0.2cm}
    \caption{\textbf{SFT weight-delta initialization for post-trained Qwen models.}}
    \vspace{-0.2cm}
    \label{fig:1_phase_weight_delta_qwen_ood}
\end{figure*}

\begin{figure*}[!tbh]
    \centering
    \includegraphics[width=0.99\linewidth]{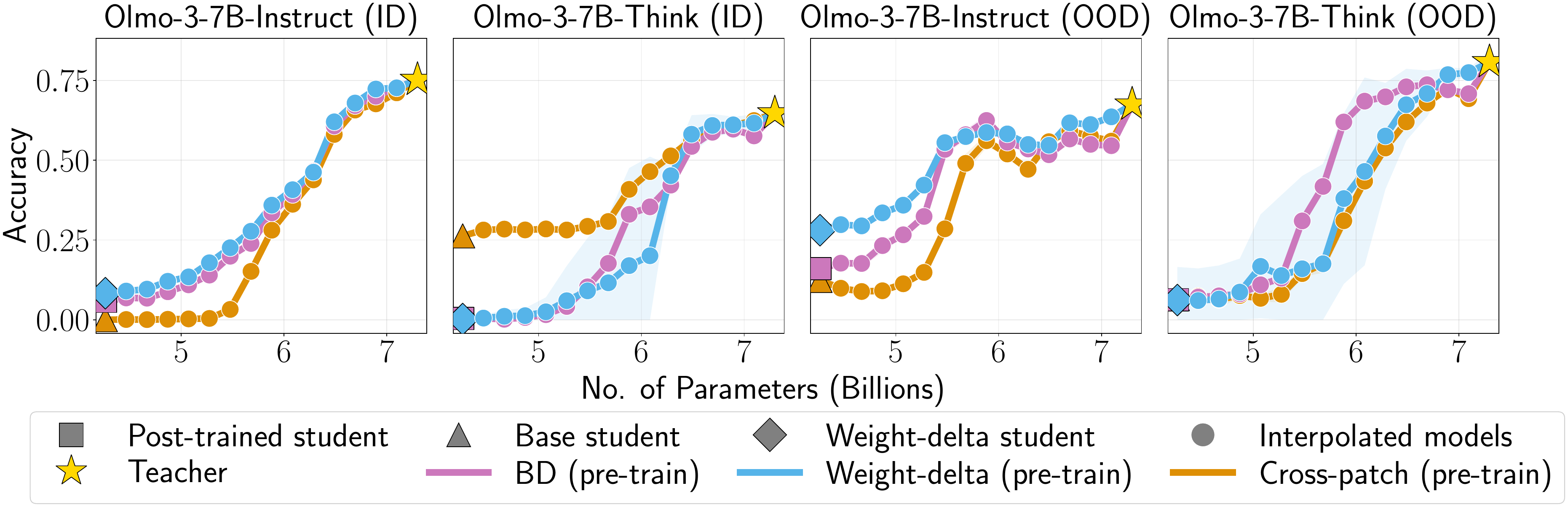}
    \vspace{-0.2cm}
    \caption{\textbf{Pre-training weight-delta initialization performance for post-trained Olmo models.}}
    \vspace{-0.1cm}
    \label{fig:1_phase_weight_delta_olmo}
\end{figure*}

\begin{figure*}[!tbh]
    \centering
    \includegraphics[width=0.80\linewidth]{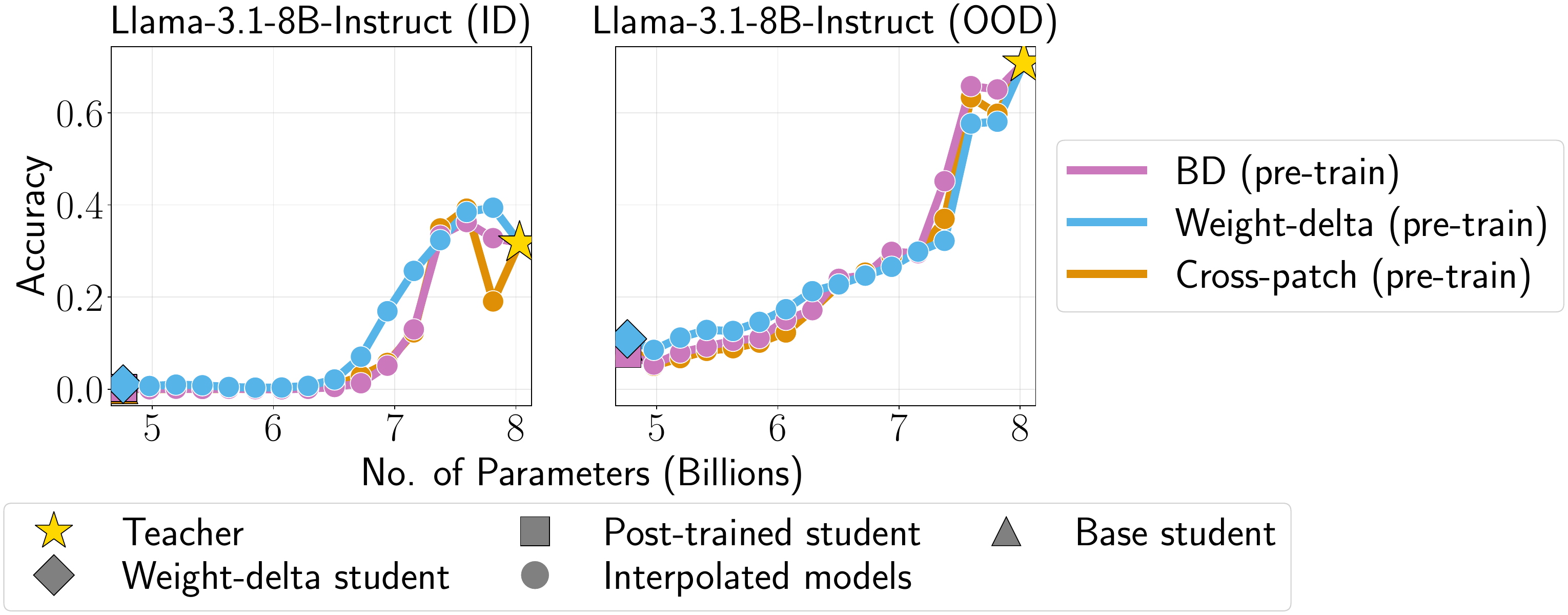}
    \vspace{-0.2cm}
    \caption{\textbf{Pre-training weight-delta initialization performance for post-trained Llama models.}}
    \vspace{-0.3cm}
    \label{fig:1_phase_weight_delta_llama}
\end{figure*}

We evaluate whether weight-delta initialization works when we calculate the weight delta from base model distilled with pre-training or SFT phase only. This setting tests whether the transferable component of distillation can be obtained without the two-phase distillation setup.

\paragraph{Setup.}
We test whether weight-delta transfer works with single-phase deltas, i.e., deltas computed from a base student distilled on only one type of data (pre-training or SFT). We then add the respective deltas to initialized post-trained students across model families. For comparison, we evaluate these single-phase weight-delta models against two references: the directly distilled BD~(pre-train) and BD~(SFT) students (which use no delta transfer), and the corresponding cross-patch baselines. We report the results for Qwen (Figures~\ref{fig:1_phase_weight_delta_qwen_id} and \ref{fig:1_phase_weight_delta_qwen_ood}), Olmo (Figure~\ref{fig:1_phase_weight_delta_olmo}), and Llama (Figure~\ref{fig:1_phase_weight_delta_llama}) families. 

\paragraph{Results.}
Across model families, pre-training weight-delta initialization performs comparably to, and in some cases better than, BD~(pre-train), indicating that the alignment learned during base-model pre-training distillation transfers across post-trained variants. The SFT-only delta transfers less smoothly. Weight-delta initialization tracks BD~(SFT) closely for the instruct variant, but meaningfully underperforms on the other three variants. The weight-delta setup even performs worse than the cross-patch~(SFT) baseline for Qwen3-4B, indicating that SFT-only distillation on the base model is not enough to facilitate effective weight-delta transfer across post-trained models. 

Furthermore, single-phase weight-delta initialization setups remain weaker than the two-phase methods. Both \sys~(weight-delta) and \sys~(distilled) achieve stronger interpolation performance, reflecting that both phases are necessary to fully recover downstream generation and reasoning capabilities.

\subsection{Continued Training from Weight-Delta Initialization}
\label{appendix:weight-delta-continued-training}

\begin{figure*}[!tbh]
    \centering
    \includegraphics[width=1\linewidth]{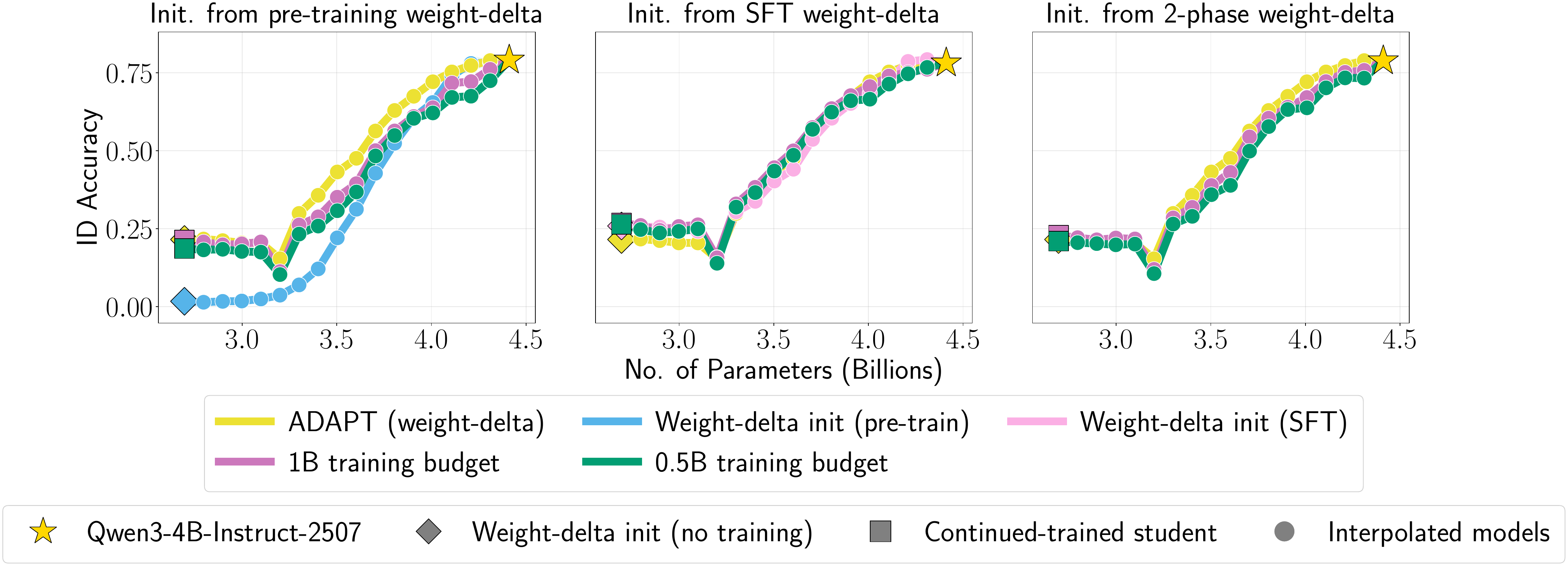}
    \caption{
    \textbf{Continued training from weight-delta initialization does not improve interpolation performance.}
    We initialize the instruct student using a pre-training, SFT, or two-phase weight delta, computed from base distillation on pre-training data, SFT data, or both respectively. We then continue training for 0.5B or 1B tokens. Continued training does not meaningfully improve over weight-delta initialization.
    }
    \label{fig:wd-continued-training}
\end{figure*}

We investigate whether weight-delta initialization can serve as a stronger initialization for additional distillation. Specifically, we ask whether continued training from a weight-delta-initialized student can outperform \sys~(weight-delta). 

\paragraph{Setup.} We initialize the student by adding one of three weight deltas to it, each obtained by distilling the base model on different data: pre-training data only, SFT data only, or both phases (two-phase). Starting from each initialization, we continue training for either 0.5B or 1B tokens using the same distillation setup described in Appendix~\ref{appendix:hyperparameters}. We also sweep the learning rates and find that 3e-4, the value used in Appendix~\ref{appendix:hyperparameters}, performs the best. 

\paragraph{Results.}
Figure~\ref{fig:wd-continued-training}  shows that continued training from weight-delta initialization does not meaningfully improve interpolation performance beyond \sys~(weight-delta). Training for 0.5B or 1B tokens converges to similar interpolation curves, regardless of whether the initialization comes from pre-training (left panel), SFT (middle panel), or two-phase (right panel) weight delta. This suggests that continued distillation largely washes out the difference between the different initializations rather than improving the final interpolated family.

Overall, these results indicate that the main benefit of weight-delta initialization comes from its zero-training transferability, not from providing a better starting point for further training.

\begin{figure*}[!tbh]
    \centering
    \includegraphics[width=0.9\linewidth]{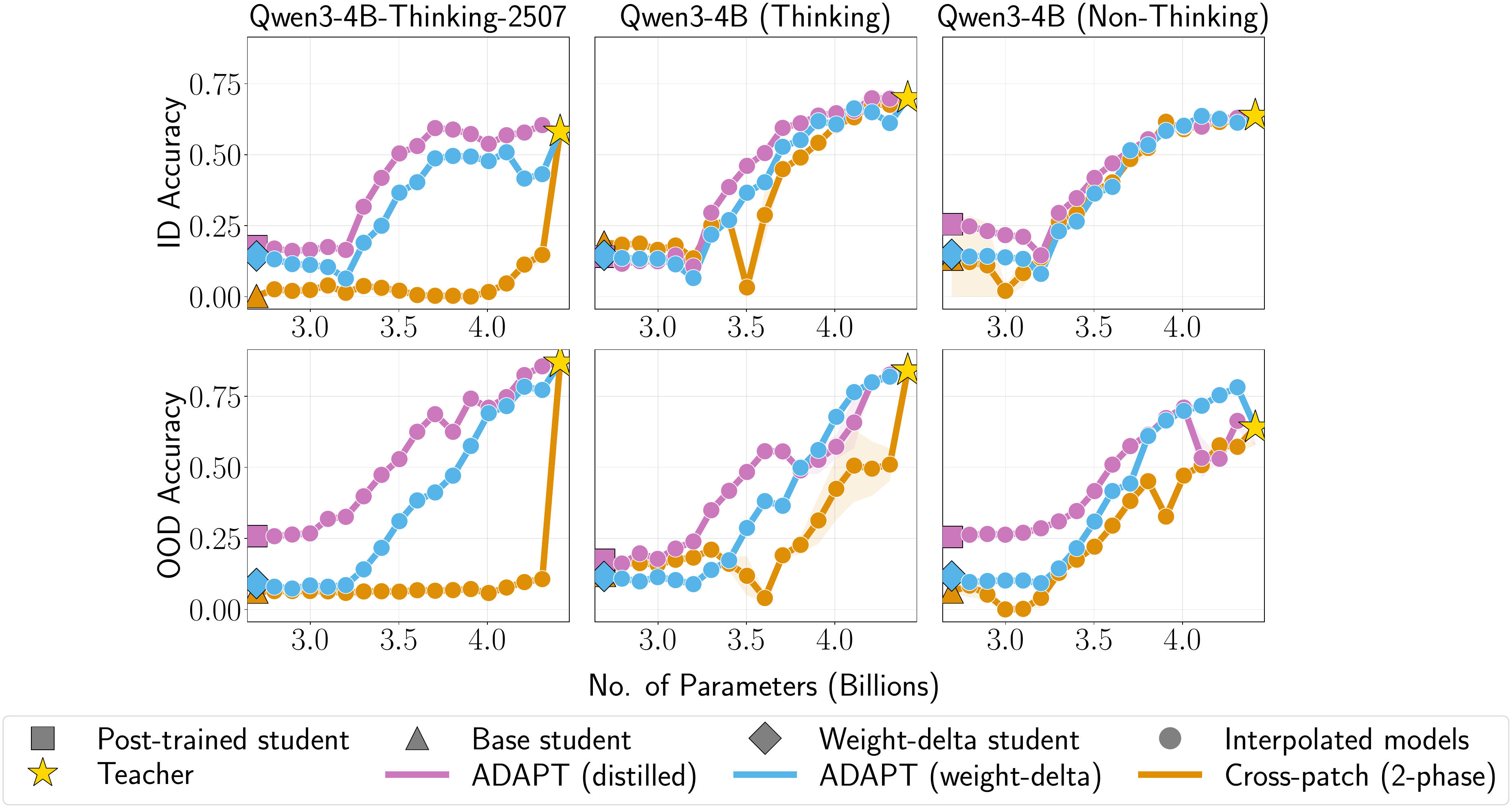}
    \caption{
    \textbf{Results for \sys~(weight-delta) using delta from Qwen3-4B-Instruct-2507.}
    }
    \label{fig:qwen-instruct-delta}
\end{figure*}

\begin{figure*}[!tbh]
    \centering
    \includegraphics[width=0.9\linewidth]{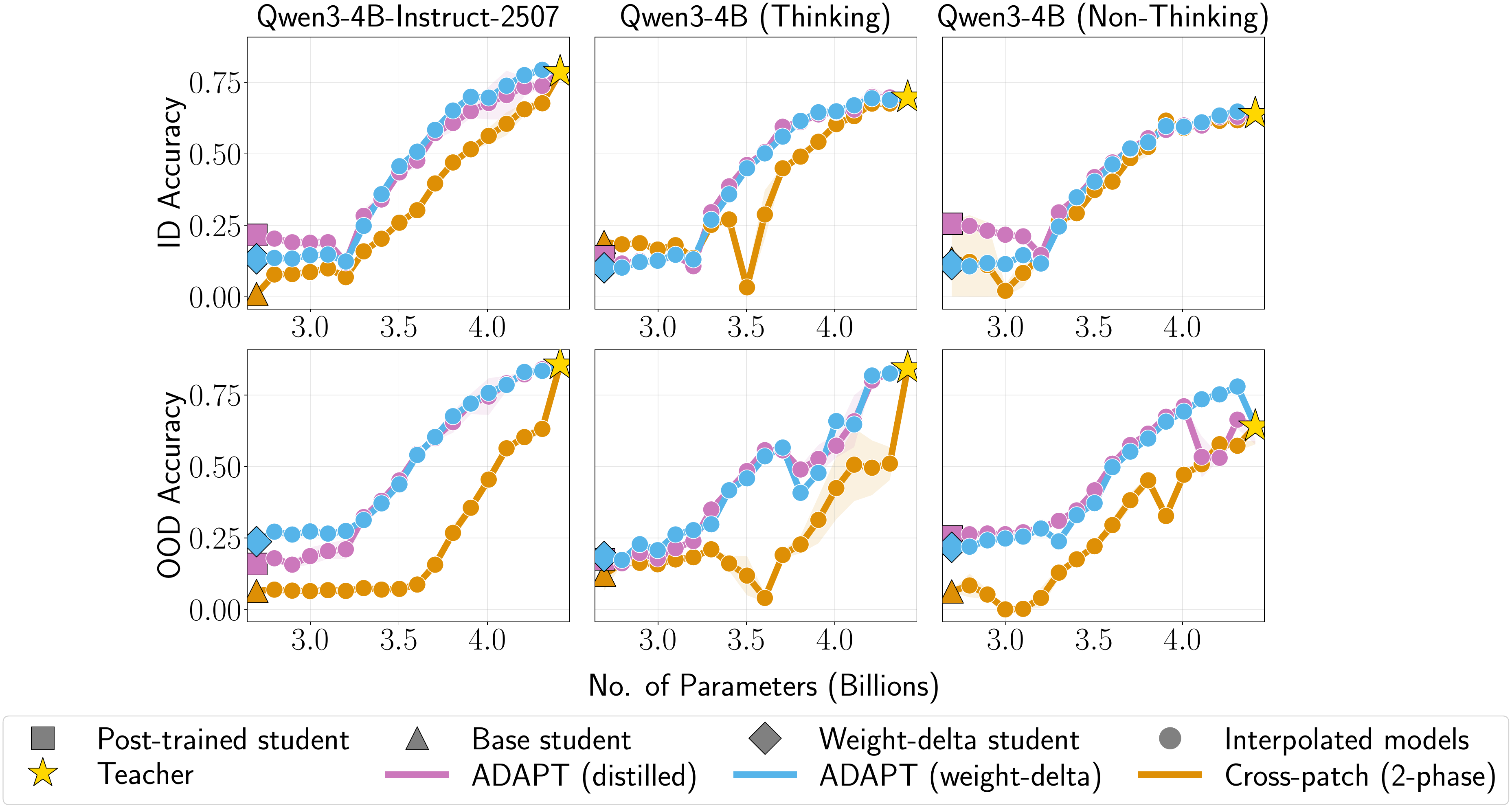}
    \caption{
    \textbf{Results for \sys~(weight-delta) using delta from Qwen3-4B-Thinking-2507.}
    }
    \vspace{-0.2cm}
    \label{fig:qwen-think-delta}
\end{figure*}

\subsection{Weight-Delta Initialization with Post-trained Model Deltas}
\label{appendix:ablation-post-trained-deltas}

In this section, we test whether distillation done directly on post-trained students (instead of the base student) can be transferred to other post-trained variants via weight-delta initialization. 

\paragraph{Setup.} We obtain the post-trained model weight deltas by directly performing two-phase distillation on Qwen3-4B-Instruct-2507 and Qwen3-4B-Thinking-2507, and subtracting their corresponding initialized (pre-distillation) students. We then initialize the other post-trained models using the weight deltas from the instruct and thinking models separately following Section~\ref{sec:weight-delta-init}. 

\paragraph{Results.} Figures~\ref{fig:qwen-instruct-delta} and \ref{fig:qwen-think-delta} show that distillation effects can be transferred onto other post-trained variants using post-trained weight deltas. \sys~(weight-delta) using instruct and thinking model weight deltas both substantially outperform the cross-patch (2-phase) baseline. Compared against the higher-compute upper bound \sys~(distilled), the instruct model delta performs competitively on ID tasks, while suffering slight degradation on OOD tasks similar to the base model delta. The thinking model delta matches the performance of \sys~(distilled) on both ID and OOD tasks. Overall, these results indicate that the transferable distillation delta is not specific to the base model, deltas computed from post-trained students transfer to other variants equally effectively.

\subsection{Ablating SFT Phase Loss Components}
\label{appendix:ablation-sft-loss-term}

While the combination of cross entropy, KL, and cosine loss terms is optimal for base model distillation under the boomerang distillation setup~\citep{kangaslahti2026boomerangdistillationenableszeroshot}, aligning the student too closely with the less capable base teacher on reasoning tasks during the SFT phase may hinder its learning. In this section, we explore whether relaxing the base student's alignment with the teacher can improve the weight-delta transfer performance. 

\begin{figure*}[!tbh]
    \centering
    \includegraphics[width=0.9\linewidth]{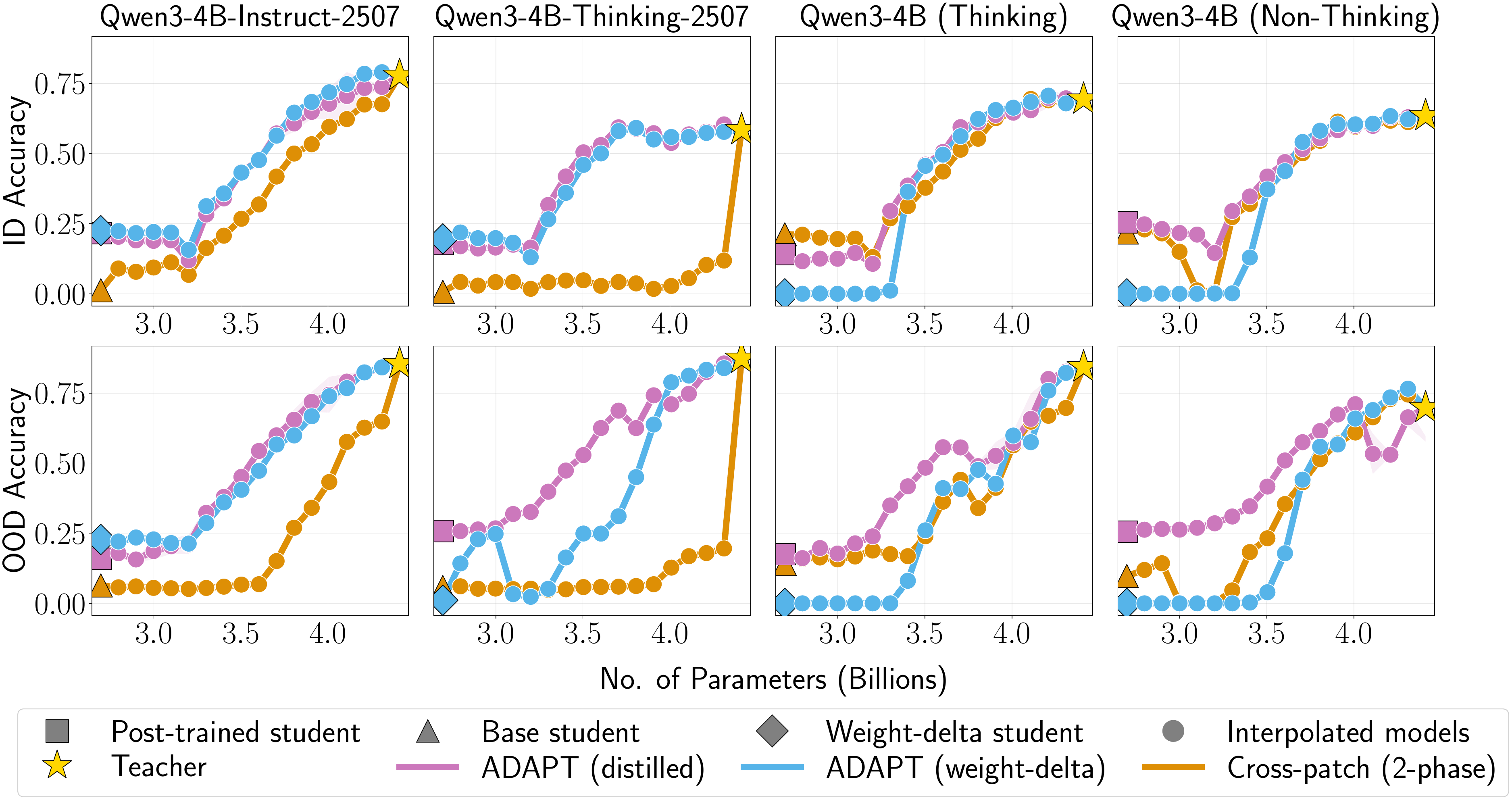}
    \vspace{-0.2cm}
    \caption{
    \textbf{Results for ablating KL loss term in SFT phase.}
    }
    \vspace{-0.3cm}
    \label{fig:sft_loss_no_kl}
\end{figure*}

\begin{figure*}[!tbh]
    \centering
    \includegraphics[width=0.9\linewidth]{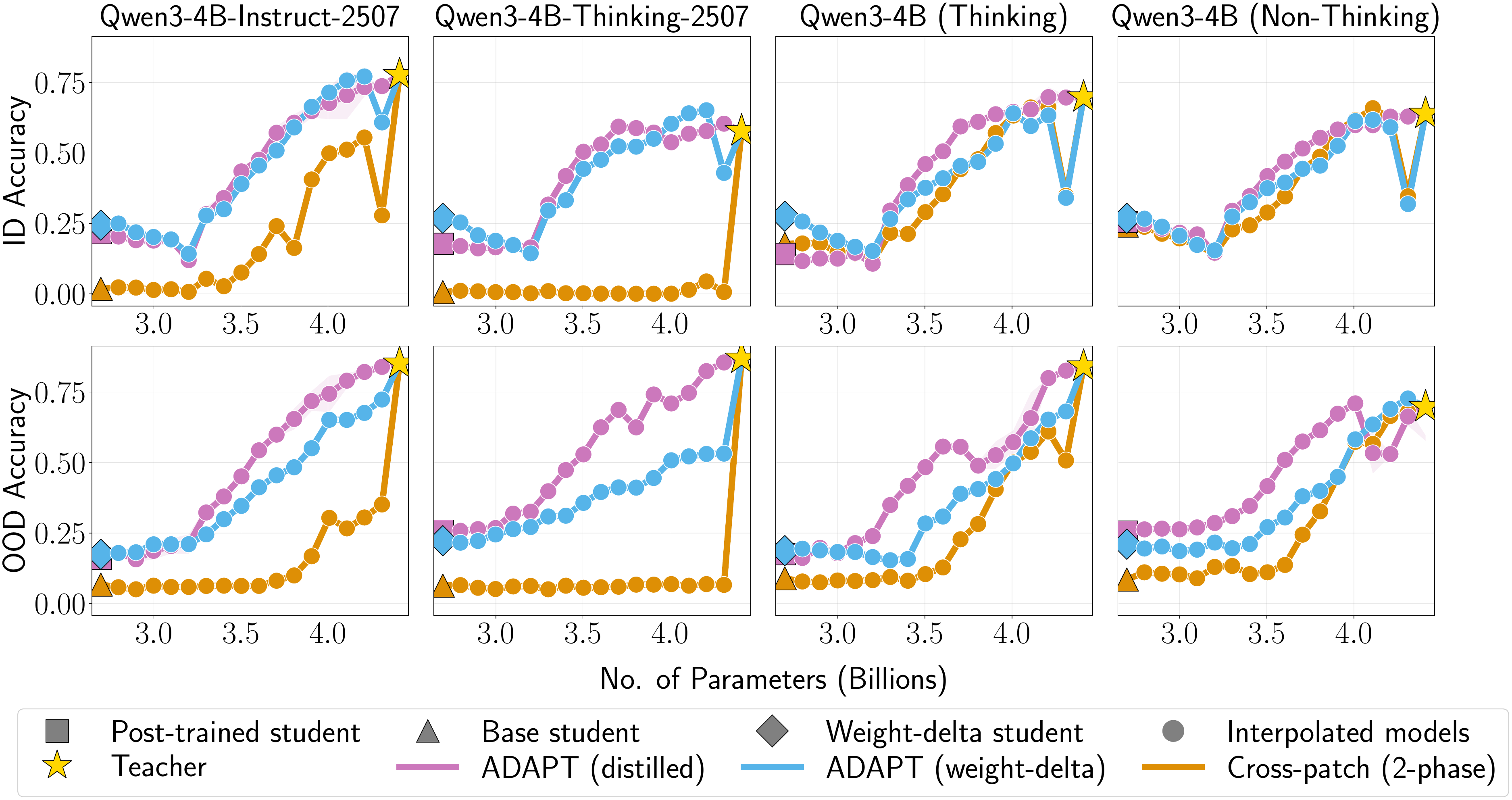}
    \vspace{-0.2cm}
    \caption{
    \textbf{Results for ablating cosine loss term in SFT phase.}
    }
    \vspace{-0.3cm}
    \label{fig:sft_loss_no_cos}
\end{figure*}

\begin{figure*}[!tbh]
    \centering
    \includegraphics[width=0.9\linewidth]{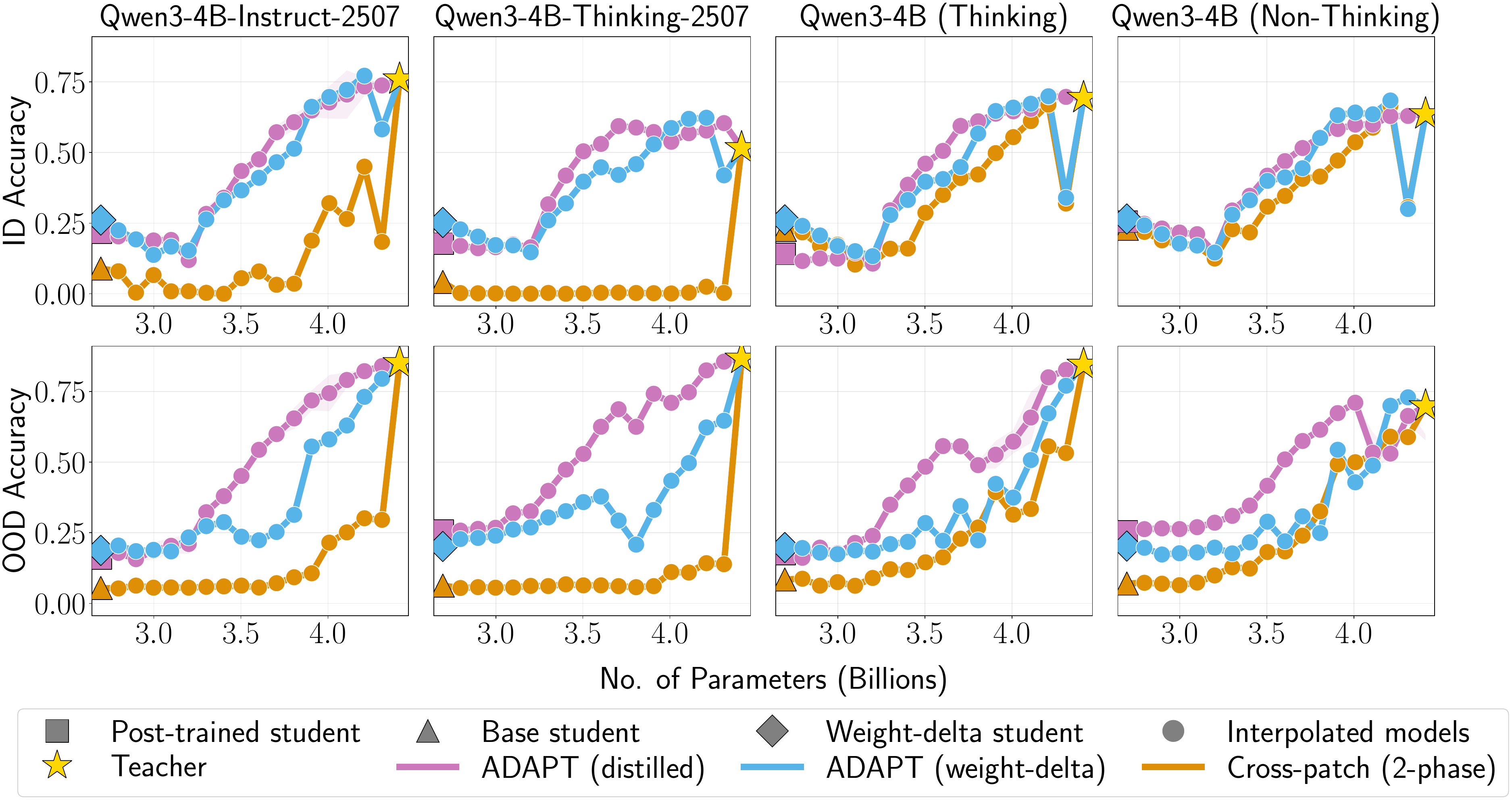}
    \vspace{-0.2cm}
    \caption{
    \textbf{Results for ablating both KL and cosine loss terms in SFT phase (cross-entropy loss only).}
    }
    \vspace{-0.4cm}
    \label{fig:sft_loss_no_kl_cos}
\end{figure*}

\paragraph{Setup.} We keep the pre-training phase of base model distillation unchanged and ablate the loss terms used in the SFT phase, always retaining cross-entropy while removing the KL term, the cosine term, or both. For each ablated objective, we compute the resulting base model weight delta and use it to initialize the four Qwen post-trained variants following Section~\ref{sec:weight-delta-init}. We compare each ablation against the higher-compute upper bound \sys~(distilled) and the cross-patch (2-phase) baseline.

\paragraph{Results.} Removing the KL term leaves interpolation behavior largely unchanged (Figure~\ref{fig:sft_loss_no_kl}), indicating that it contributes little during the SFT phase. Removing the cosine term has a substantially larger effect (Figure~\ref{fig:sft_loss_no_cos}): ID performance improves across all four variants, matching \sys~(distilled), while OOD performance degrades overall. This split is size-dependent --- the ID gains come almost entirely from smaller interpolated models, where the full objective left the weight-delta students near total collapse until roughly 3.3B parameters, whereas OOD performance falls off at intermediate and larger sizes. Removing both terms behaves similarly but with even weaker OOD performance (Figure~\ref{fig:sft_loss_no_kl_cos}). Therefore, the cosine term in SFT phase trades ID performance for OOD generalization, and the choice of whether to retain it depends on the specific use cases. 

\section{Weight-Delta Alignment Experiments} 
\label{app:weight_delta_alignment} 

Results in the preceding appendices and Appendix~\ref{appendix:aggregated_results_all} show that \sys~(weight-delta) is more successful for some models and setups than others, matching the \sys~(distilled) performance at its best. However, understanding when exactly weight-delta transfer succeeds remains an important open question. Notably, this is a recognized and challenging problem even in the closely related setting of model merging and task vectors, where identifying the factors that govern whether weight-arithmetic combination succeeds is an active area of research~\citep{rahamim2026mergecausesmodelmergeability}. Here, we present preliminary results that  yield some useful signals. 

The task-vector view~\citep{ilharco2023editingmodelstaskarithmetic} treats a capability acquired through finetuning as a shift in weight space---in the language of this paper, a weight delta. A natural predictor of transfer success is therefore the alignment between the base distillation delta ($\vtheta_{S,\mathrm{distill}}^{\mathrm{base}}-\vtheta_{S,\mathrm{init}}^{\mathrm{base}}$) and each variant's own distillation delta ($\vtheta_{S,\mathrm{distill}}^{\mathrm{PT}}-\vtheta_{S,\mathrm{init}}^{\mathrm{PT}}$). Table~\ref{tab:weight_delta_alignment} reports the cosine similarity and norm ratio between these deltas.

Alignment tracks transfer reliability. Instruct and hybrid variants align most closely (cosine of 0.73-0.75 for Qwen and Olmo), thinking variants less so (0.57-0.64), and Llama least of all (0.30), mirroring the order in which we observe transfer to be reliable, partially reliable, and unreliable. Norm ratios stay near 1, so the deltas differ in direction rather than magnitude. Although this is a correlational analysis, it suggests directional alignment as a candidate diagnostic for when weight-delta transfer will hold. We leave further investigation to future work. 

\begin{table}[h]
\centering
\begin{tabular}{lcc}
\toprule
Variant & Cosine & Norm ratio \\
\midrule
Qwen3-4B-Instruct-2507  & 0.73 & 1.00 \\
Qwen3-4B-Thinking-2507  & 0.57 & 0.92 \\
Qwen3-4B                & 0.75 & 1.00 \\
Qwen3-14B                & 0.75 & 1.00 \\
Olmo-3-7B-Instruct      & 0.73 & 1.02 \\
Olmo-3-7B-Think         & 0.64 & 0.94 \\
Llama-3.1-8B-Instruct   & 0.30 & 0.95 \\
\bottomrule
\end{tabular}
\caption{Cosine similarity and norm ratio between base and post-trained deltas for each variant.}
\label{tab:weight_delta_alignment}
\end{table}

\section{Comparison to Layer Pruning Methods} 
\label{appendix:comparison_to_pruning}

\begin{figure}[!tb]
    \centering
    \includegraphics[width=\linewidth]{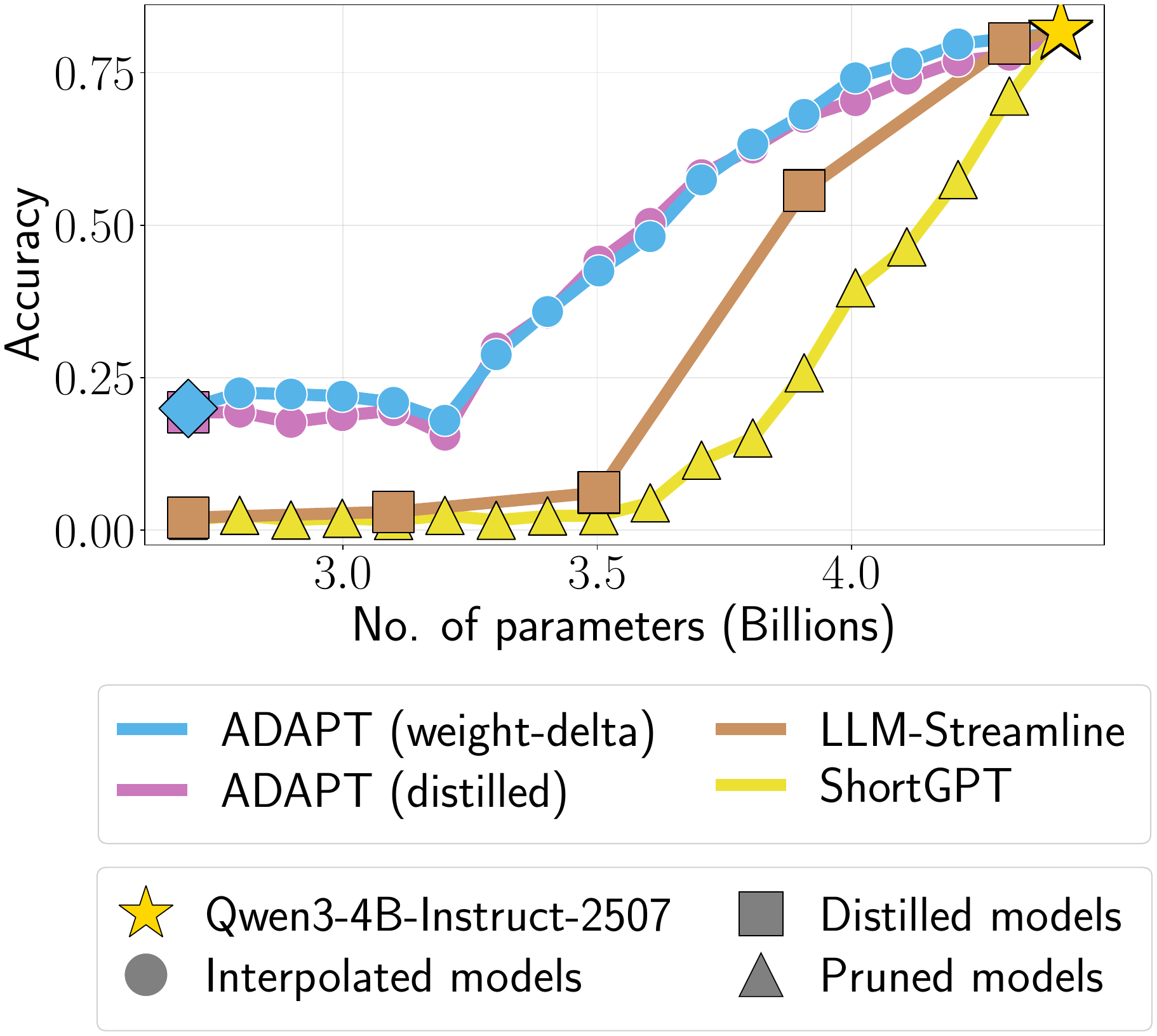}
    \caption{\textbf{\sys significantly outperforms layer pruning baselines on generation tasks.} We compare  \sys~(distilled) and \sys~(weight-delta) to two popular layer pruning methods for generation tasks, ShortGPT \citep{men2024shortgptlayerslargelanguage} and LLM-Streamline \citep{chen2025streamliningredundantlayerscompress}, and show that our method performs substantially better across all sizes.}
    \vspace{-0.3cm}
    \label{fig:baseline_comparison}
\end{figure}

We compare \sys against layer pruning methods on generation benchmarks to show that it produces substantially stronger intermediate models. 

\paragraph{Setup.} We consider two popular layer pruning methods: ShortGPT \citep{men2024shortgptlayerslargelanguage} and LLM-Streamline \citep{chen2025streamliningredundantlayerscompress}. 
ShortGPT prunes the least influential layers, as determined by the block influence score (see Appendix~\ref{appendix:shortgpt}). LLM-Streamline identifies the block of layers with the lowest block influence score and replaces it with a lightweight network (see Appendix~\ref{appendix:llm-streamline}).

\paragraph{Results.} Figure~\ref{fig:baseline_comparison} shows that across all model sizes, \sys~(distilled) and \sys~(weight-delta) models consistently outperform ShortGPT and LLM-Streamline, which struggle to recover meaningful generation capability, especially at smaller sizes. 

\subsection{ShortGPT}
\label{appendix:shortgpt}

For ShortGPT \citep{men2024shortgptlayerslargelanguage}, we first calculate the Block Influence score (BI), which is the cosine distance between the input and output activations. A higher BI score indicates higher importance of a specific layer. The BI score of the $\ell^\text{th}$ layer can be calculated as follows: 
$$BI_{\ell} = 1 - \mathbb{E}_{X,t}\left[\frac{\bm{x}^{(\ell)}_{t} \cdot \bm{x}^{(\ell+1)}_{t}}{\|\bm{x}^{(\ell)}_{t}\| \|\bm{x}^{(\ell+1)}_{t}\|}\right]$$
where $\bm{x}^{(\ell)}_{t}$ is the $t^\text{th}$ row of hidden states of the $\ell^\text{th}$ layer. When pruning, we sequentially remove layers with the lowest BI score. We compute BI using a held-out set of 128 calibration samples from Nemotron.

\subsection{LLM-Streamline}
\label{appendix:llm-streamline}

For LLM-Streamline \citep{chen2025streamliningredundantlayerscompress}, we identify the block of $n$ layers $(\ell^*, \ell^*+n)$ with the lowest BI score (using the same calibration set from Nemotron as ShortGPT) and replace it with a single lightweight network. We then train the replacement network to imitate the pruned block using mean squared error (MSE): 
$$\min_h\mathbb{E}_{(\bm{x}^{(\ell^*)}, \bm{x}^{(\ell^*+n)})\in\mathcal{D}}\text{MSE}\left(h(\bm{x}_i^{(\ell^*)}), \bm{x}_i^{(\ell^*+n)}\right)$$
where $h$ denotes the lightweight network, $\mathcal{D}$ denotes the recorded hidden states of samples, and $\bm{x}^{(i)}$ is the hidden state of the $i^\text{th}$ layer. 

Following \citet{chen2025streamliningredundantlayerscompress}, we further post-train the replaced layer on language modeling loss in order to have a fairer comparison with our method. 

\begin{table}[!tb]
\begin{center}
\begin{tabular}{cc}
\toprule
\multicolumn{1}{c}{\bf Hyperparameters}  &\multicolumn{1}{c}{\bf Values} \\
\midrule
LR (MSE) & 2e-4 \\
LR (LM-loss) & 5e-5 \\
LR scheduler & cosine \\
Warmup ratio & 0.01 \\
Optimizer & AdamW \\
Adam betas & (0.9, 0.999) \\
Adam epsilon & 1e-8 \\
Weight decay & 1e-3 \\
Max gradient norm & 5.0 \\
Mixed precision & bf16 \\
\bottomrule
\end{tabular}
\end{center}
\caption{\textbf{Hyperparameters for LLM-Streamline training.}}
\label{tab:streamline_hyperparams}
\end{table}

In our implementation, we use a transformer block from the original Qwen3-4B-Instruct-2507 model as the lightweight layer and train it on the same data and token budget as \sys~(distilled). We allocate 80\% of the total training budget to lightweight network training and the remaining 20\% to post-training. The MSE stage runs for 192 Pile steps and 234 Nemotron steps, while the LM-loss stage runs for 48 Pile steps and 59 Nemotron steps. All stages use sequence length of 1024, with effective batch size of 2048 for Pile and 4096 for Nemotron. We use the training hyperparameters shown in Table~\ref{tab:streamline_hyperparams}. We perform evaluation following the same sampling strategy as discussed in Appendix~\ref{appendix:eval_implementation}, using the non-thinking sampling parameters. 

\section{Additional Results for Adaptive Model-Size Selection}
\label{appendix:additional_results_adaptive_model_size}

\subsection{Input Difficulty Classification}
\label{appendix:input_diff_classification}

\begin{figure*}[!p]
\centering
\begin{tcolorbox}[colback=blue!5, colframe=black, width=0.95\linewidth, boxrule=0.8pt]
\textbf{Difficulty Grading Prompt} \\
You will be given a math problem. Your job is to grade the difficulty level from 1-10 according to
the AoPS standard. \\
Here is the standard: \\
All levels are estimated and refer to averages. The following is a rough standard based on the USA
tier system AMC 8 - AMC 10 - AMC 12 - AIME - USAMO/USAJMO - IMO, representing Middle
School - Junior High - High School - Challenging High School - Olympiad levels. Other contests can
be interpolated against this. \\
Notes: \\
- Multiple choice tests like AMC are rated as though they are free-response. \\
- Some Olympiads are taken in 2 sessions, with similarly difficult sets of questions. \\
Scale \\
1: Problems strictly for beginners (e.g., AMC 8 1-20, AMC 10 1-10). \\
2: For motivated beginners (e.g., AMC 8 21-25, AMC 10 11-20). \\
3: Advanced beginner problems (e.g., AMC 10 21-25, AIME 4-6). \\
4: Intermediate-level problems (e.g., AMC 12 21-25, AIME 7-9). \\
5: More difficult AIME problems (10-12), simple Olympiad problems. \\
6: High-level AIME (13-15), introductory Olympiad questions. \\
7: Tougher Olympiad questions requiring technical knowledge. \\
8: High-level Olympiad questions. \\
9: Expert Olympiad questions. \\
10: Extremely difficult problems unsuitable for standard competitions. \\
Examples \\
$<$1: Counting edges/corners/faces of a cube (2003 AMC 8 Problem 1). \\
1: How many integers satisfy $|x|<3\pi$? (2021 AMC 10B Problem 1). \\
2: Probability problem with die rolls (2021 AMC 10B Problem 18). \\
3: Geometry problem with area calculation (2018 AMC 10A Problem 24). \\
4: Recursive sequence problem (2019 AMC 10B Problem 24). \\
5: System of equations (JBMO 2020/1). \\
6: Complex geometry problem (2020 AIME I Problem 15). \\
7: IMO 2015 Problem 1 about balanced sets. \\
8: IMO 2014 Problem 5 about coin denominations. \\
9: IMO 2022 Problem 3 about prime arrangements. \\
10: IMO 2020 Problem 6 about point separation. \\
Task: \\
Problem to be labeled: \{problem\}. \\
Please put your estimation of the difficulty as a single number from 1-10. \\
Important: You MUST respond with only a single number between 1-10 indicating the difficulty of problem, not the solution of the problem. \\
Output: $<$difficulty level from 1-10$>$
\end{tcolorbox}
\caption{\textbf{Prompt for difficulty classification.} Adapted from~\citet{waheed2025tokensefficientmathreasoning}.}
\label{fig:difficulty_prompt}
\end{figure*}

\begin{figure*}[!tbh]
    \centering
    \includegraphics[width=0.85\linewidth]{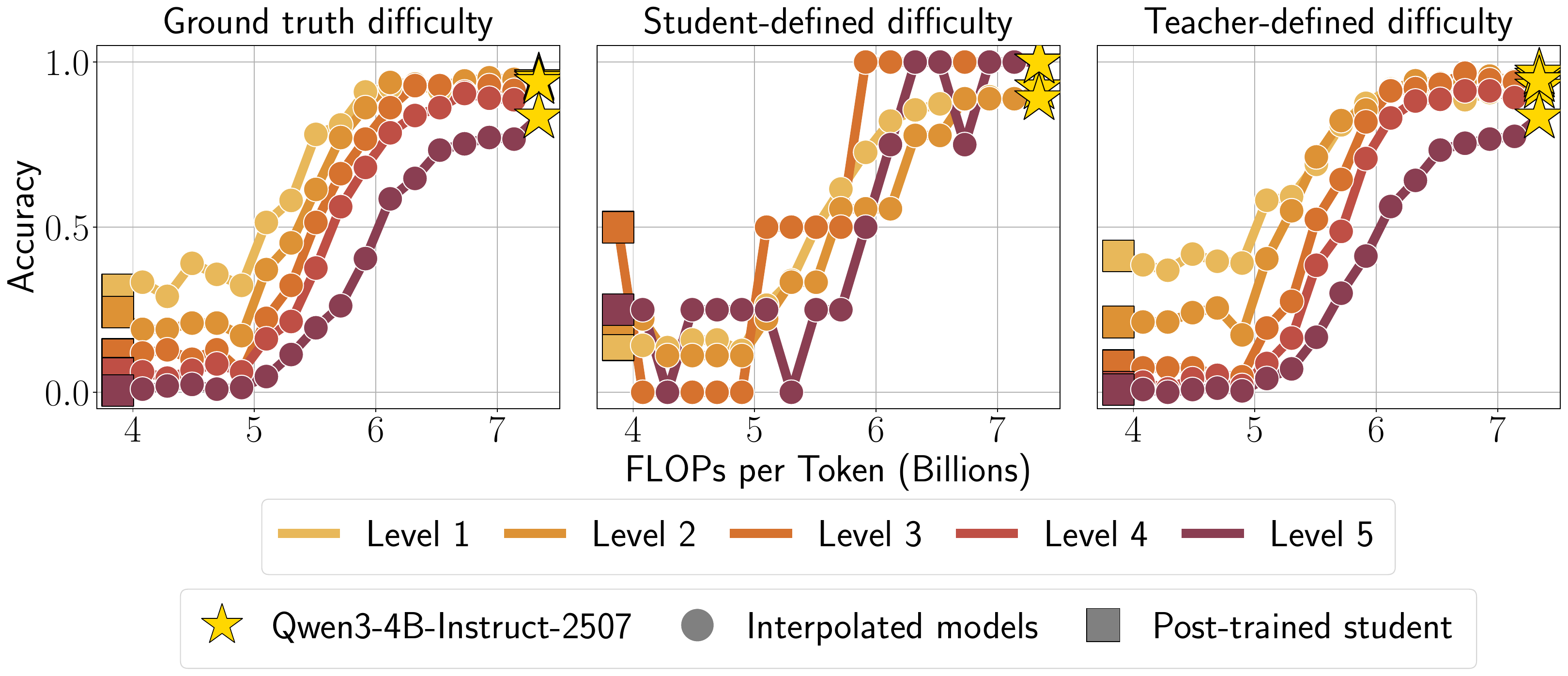}
    \caption{\textbf{Interpolation performance of Qwen3-4B-Instruct-2507 for different difficulty levels.} We find that the teacher-defined difficulty score produces similar interpolation curves to the ground truth difficulty scores, but student-defined difficulty is unreliable. This shows that the teacher-defined labels can act as a reasonable approximation for groundtruth difficulties.}
    \vspace{-0.2cm}
    \label{fig:interpolation_by_difficulty}
\end{figure*}

As described in Section~\ref{sec:adaptive_model_size}, we use the MATH dataset \citep{hendrycks2021measuringmathematicalproblemsolving}, which contains competition-style math problems with ground truth difficulty labels according to the Art of Problem Solving (AoPS) competition ratings \citep{aops_competition_ratings}. For our task, we sample 2 separate subsets of 1050 problems from the original dataset, each with 210 problems for each difficulty level. We use one as the training set to determine the optimal model routing policy, and the other as a test set to measure the efficiency gains of our method. 

To create the model routing policy, we require a lightweight and reliable method for estimating input difficulty. We test both the student and teacher models as classifiers by prompting them to assign a difficulty rating to each question. Specifically, each input question is paired with a prompt describing the AoPS difficulty scale (Figure~\ref{fig:difficulty_prompt}), and the model is asked to output a rating from 1 to 10. We merge all levels above 5 into level 5, since we know \textit{a priori} that the MATH dataset contains problems of at most level 5 difficulty. To obtain predictions efficiently, we perform a single forward pass and extract the predicted class directly from the output logits. 

To evaluate classification performance, we use the mean absolute error (MAE):
$$\text{MAE}_{\mathcal{D}_{\text{train}}}(\bm{\hat{y}}, \bm{y}) 
= \frac{1}{|\mathcal{D}_{\text{train}}|} \sum_{i \in \mathcal{D}_{\text{train}}} |\hat{y}_i - y_i|$$
where $\mathcal{D}_{\text{train}}$ denotes the training set, $\hat{y}_i$ is the predicted difficulty for the $i^\text{th}$ problem, and $y_i$ is the corresponding ground-truth difficulty.

On our training set, the teacher achieves an MAE of 0.970, while the student has an MAE of 1.995. This indicates that the teacher's predictions are typically within one difficulty level of the ground truth, whereas the student's predictions deviate by approximately two levels on average. This suggests that the teacher provides substantially more accurate estimates for adaptive model selection. The class breakdown of the teacher and student predictions is shown in Table~\ref{tab:class_breakdown}. 

\begin{table}[!tb]
\begin{center}
\begin{tabular}{ccc}
\toprule
\multicolumn{1}{c}{\bf Levels}  &\multicolumn{1}{c}{\bf Teacher} &\multicolumn{1}{c}{\bf Student} \\
\midrule
1 & 184 & 1035 \\ 
2 & 282 & 9 \\
3 & 149 & 2 \\ 
4 & 195 & 0 \\
5 & 240 & 4 \\ 
\bottomrule
\end{tabular}
\end{center}

\vspace{-0.2cm}
\caption{\textbf{Breakdown of input-difficulty classification results from teacher and student models.} The student is unreliable for difficulty classification, as it chooses the lowest difficulty level for the vast majority of examples.}
\label{tab:class_breakdown}
\vspace{-0.4cm}
\end{table}

While the MATH dataset provides human-annotated difficulty labels, prior work has shown that human difficulty does not always align with model-perceived difficulty \citep{kordi2025revisitinggeneralizationdifficultylevels}. Figure~\ref{fig:interpolation_by_difficulty} verifies that teacher-predicted labels produce well-separated and consistently ordered performance curves, closely matching the trends from ground-truth labels, whereas student-predicted labels are noisy and do not yield clear separation. This further supports using the teacher for difficulty estimation in adaptive model selection.

We omit the cost of difficulty classification from the results in Section~\ref{sec:adaptive_model_size}, as it requires only a single teacher forward pass and is negligible relative to the average response length of 3770.5 tokens. 

\begin{figure*}[!tbh]
    \centering
    \includegraphics[width=1\linewidth]{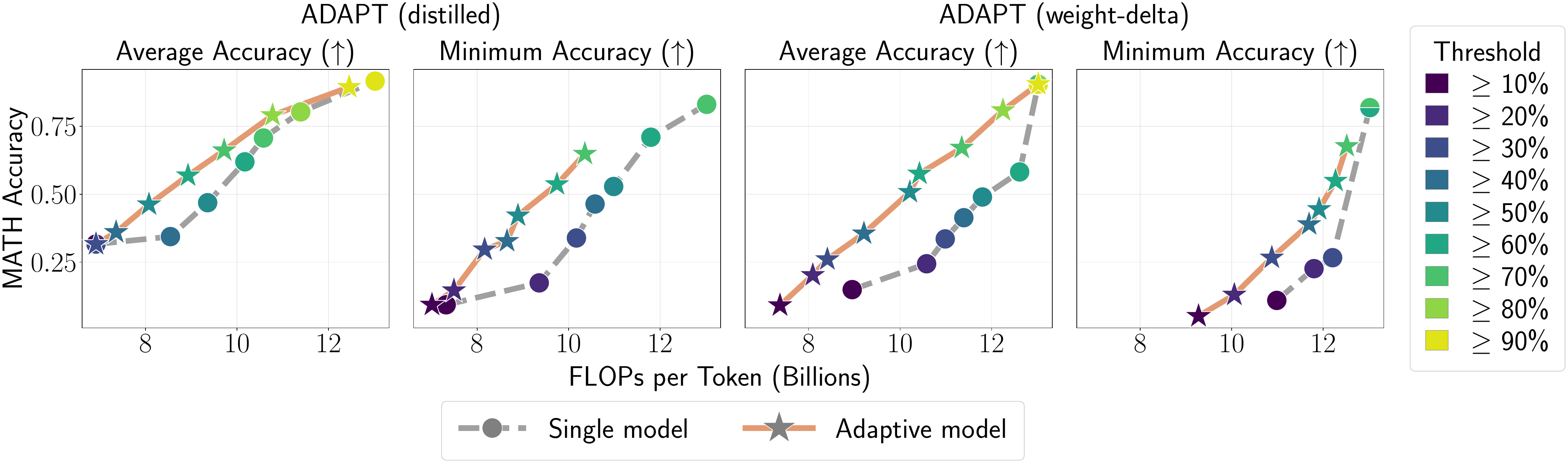}
    \vspace{-0.7cm}
    \caption{\textbf{Adaptive model size selection results for Olmo-3-7B-Instruct.}} 
    \vspace{-0.1cm}
    \label{fig:adaptive_model_size_olmo} 
\end{figure*}

\begin{figure*}[!tbh]
    \centering
    \includegraphics[width=1\linewidth]{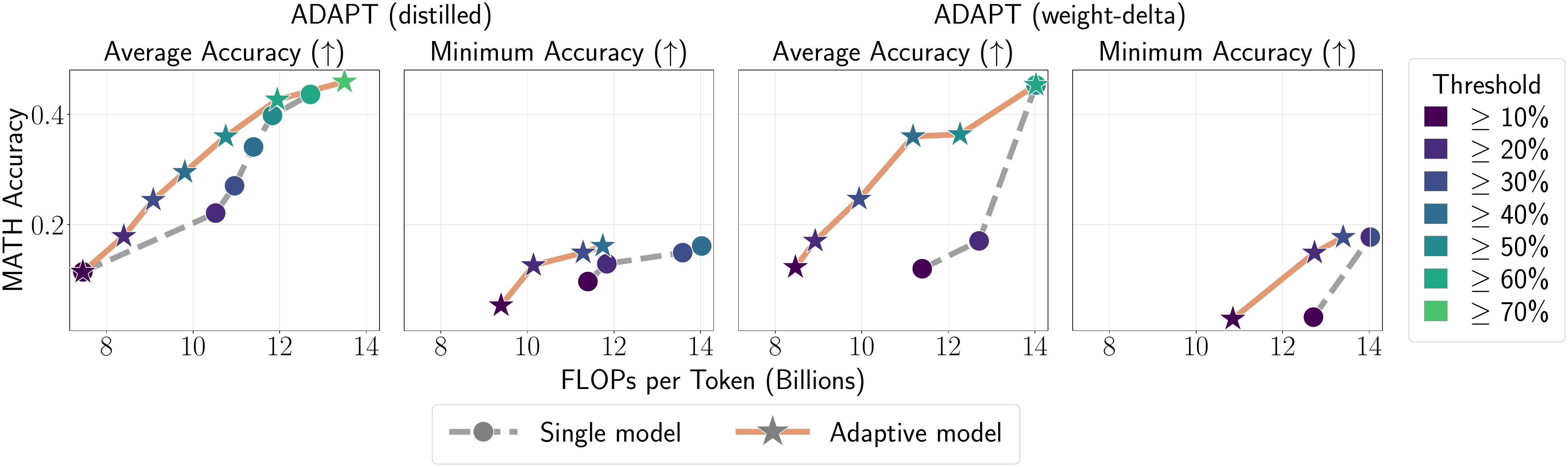}
    \vspace{-0.7cm}
    \caption{\textbf{Adaptive model size selection results for Llama-3.1-8B-Instruct.}} 
    \vspace{-0.1cm}
    \label{fig:adaptive_model_size_llama} 
\end{figure*}

\begin{figure*}[!tbh]
    \centering
    \includegraphics[width=0.85\linewidth]{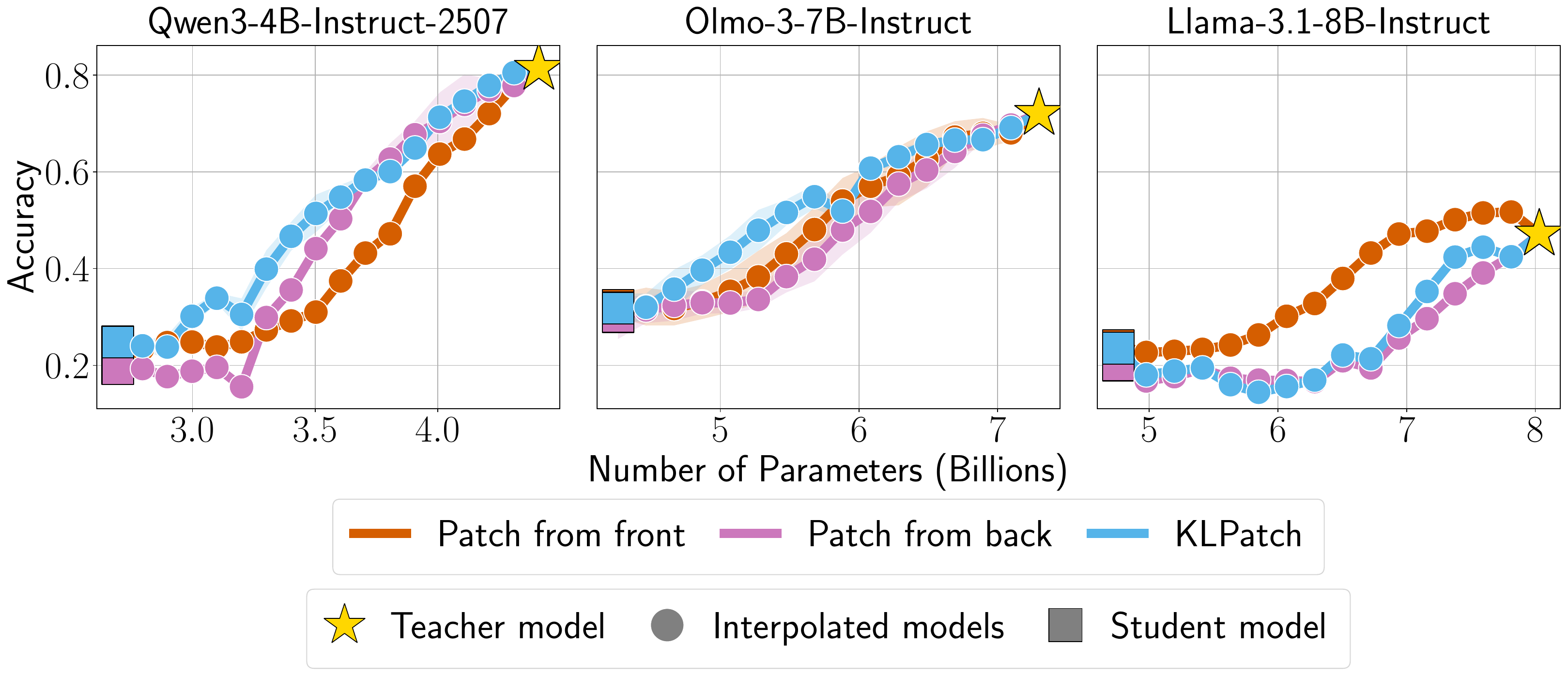}
    \vspace{-0.2cm}
    \caption{\textbf{Comparison of interpolation performance for different patching orders.} We test patching from the front, patching from the back, and KLPatch~\citep{klpatch2026}. 
    We find that KLPatch can further improve performance, indicating that more sophisticated patching strategies are orthogonal to our approach and can be used in conjunction with our student models.
    }
    \vspace{-0.3cm}
    \label{fig:kl_patch_comparison}
\end{figure*}

\subsection{Adaptive Model-Size Selection Results for Olmo and Llama} 
\label{appendix:apative_model_size_olmo_llama}

In Figures~\ref{fig:adaptive_model_size_olmo} and \ref{fig:adaptive_model_size_llama}, we report the results for the adaptive model size selection experiments for Olmo-3-7B-Instruct and Llama-3.1-8B-Instruct. Note that we use Qwen3-4B-Instruct-2507 to provide the difficulty labels (same as in Section \ref{sec:adaptive_model_size}) as it is smaller than both the teacher models of Olmo and Llama, and therefore cheaper.

\subsection{Discussion of Total Token Count}
\label{appendix:total_token_count}

While smaller models offer more favorable FLOPs per token performance, we empirically observe that they often generate more tokens overall for the same set of questions. This is because smaller models are typically less capable, leaving more problems unsolved. On these questions, the model keeps generating tokens until it hits the \texttt{max\_tokens} limit, whereas on questions it solves correctly, it completes them using far fewer tokens. We believe that with further training, this issue can be mitigated, but this is beyond the scope of this paper and we leave it for future work. 

\section{Patching Order Experiment} 
\label{appendix:patching_order_results}

In Figure~\ref{fig:kl_patch_comparison}, we show the full interpolation curves for all three model families when testing front-to-back and back-to-front patching orders. We also test KLPatch, which is a recently proposed iterative greedy KL divergence-based algorithm for selecting the patching order~\citep{klpatch2026}. At each size, KLPatch patches each student layer individually and selects the layer with minimum KL divergence to the teacher. We use a calibration set of 64 examples from the Nemotron non-thinking set in order to compute the KL divergence. 
These plots provide a more detailed view of how patching order affects interpolation behavior across model sizes.

We observe that the effect of patching order varies substantially across models. For Qwen3-4B-Instruct-2507
and Olmo-3-7B-Instruct models, KLPatch improves performance over patching from the front and patching from the back. However, for Llama-3.1-8B-Instruct, patching from the front yields the best performance. 
This suggests that the optimal patching depends on how different layers contribute to downstream generation performance in each model and architecture. 
In our experiments, we use back to front for Qwen3-4B-Instruct-2507 and front to back for Olmo-3-7B-Instruct and Llama-3.1-8B-Instruct. As KLPatch improves performance for small Qwen and Olmo models, this indicates that more sophisticated patching methods can be used in conjunction with \sys to improve the performance of post-trained model families.

\begin{figure}[!tb]
    \centering
    \includegraphics[width=1\linewidth]{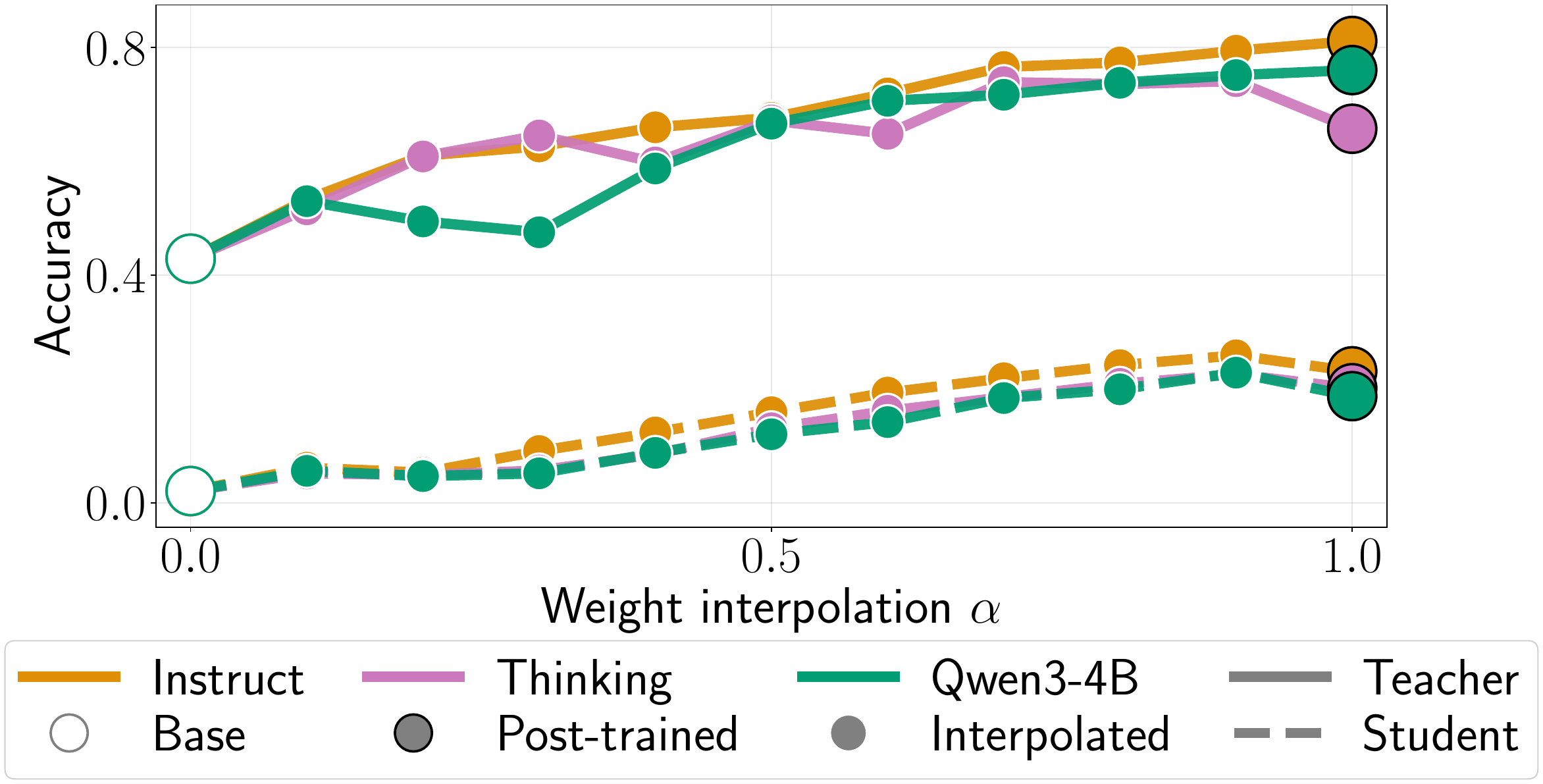}
    \caption{\textbf{Smooth weight interpolation results for generation accuracy.} Base and post-trained students display smooth interpolation behavior in downstream generation accuracy.}
    
    \label{fig:lmc_weight_interp_accuracy}
\end{figure}

\section{Additional Smooth Weight Interpolation Results} 
\label{appendix:additional_lmc}

\begin{figure}[!tb]
    \centering
    \includegraphics[width=1\linewidth]{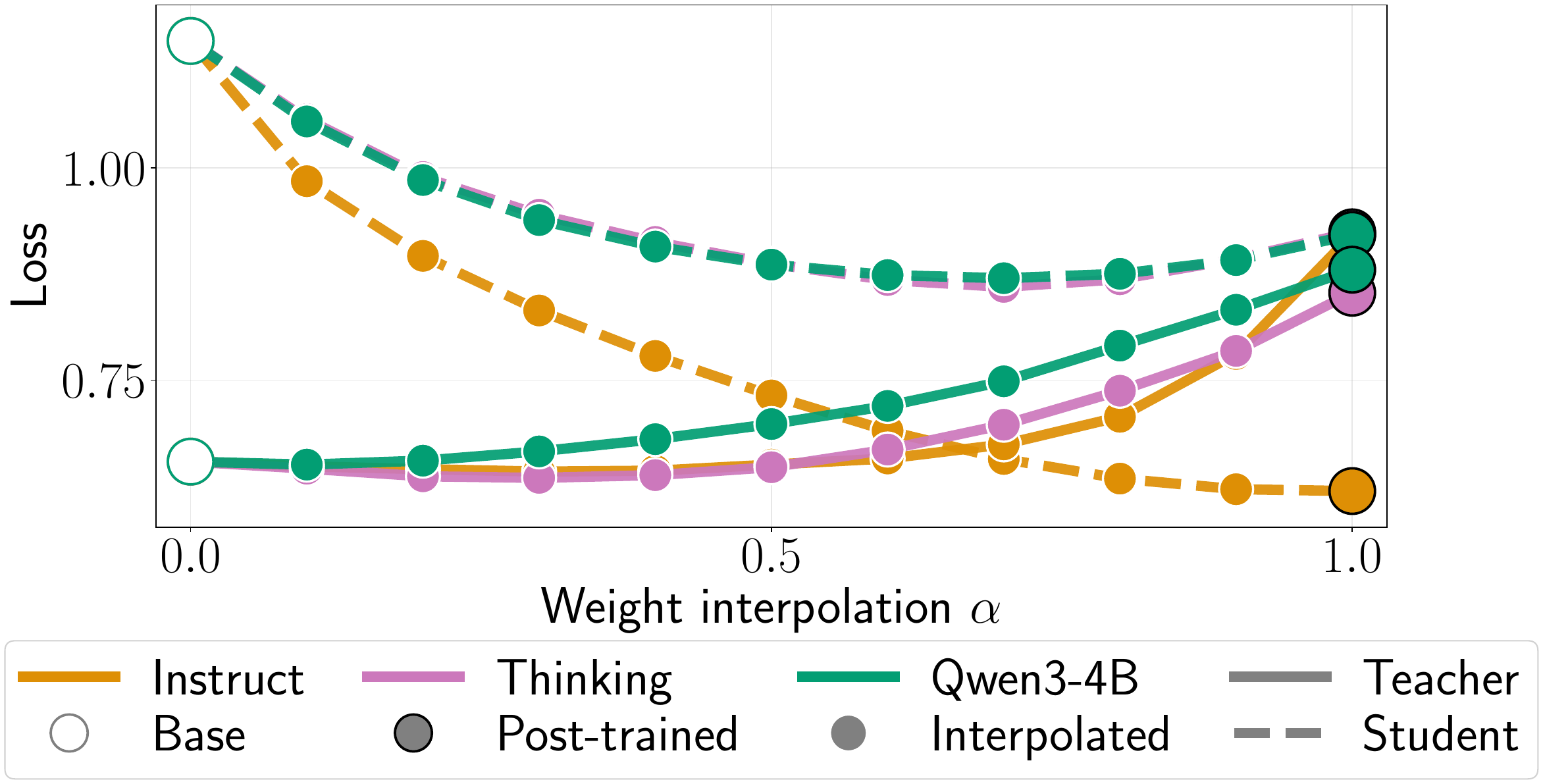}
    \caption{\textbf{Smooth weight interpolation results for loss on non-thinking Nemotron data.} Base and post-trained students display smooth interpolation behavior in loss evaluated on non-thinking Nemotron data.}
    \label{fig:lmc_weight_interp_loss_nemotron_no_think}
\end{figure}

\begin{figure}[!tb]
    \centering
    \includegraphics[width=1\linewidth]{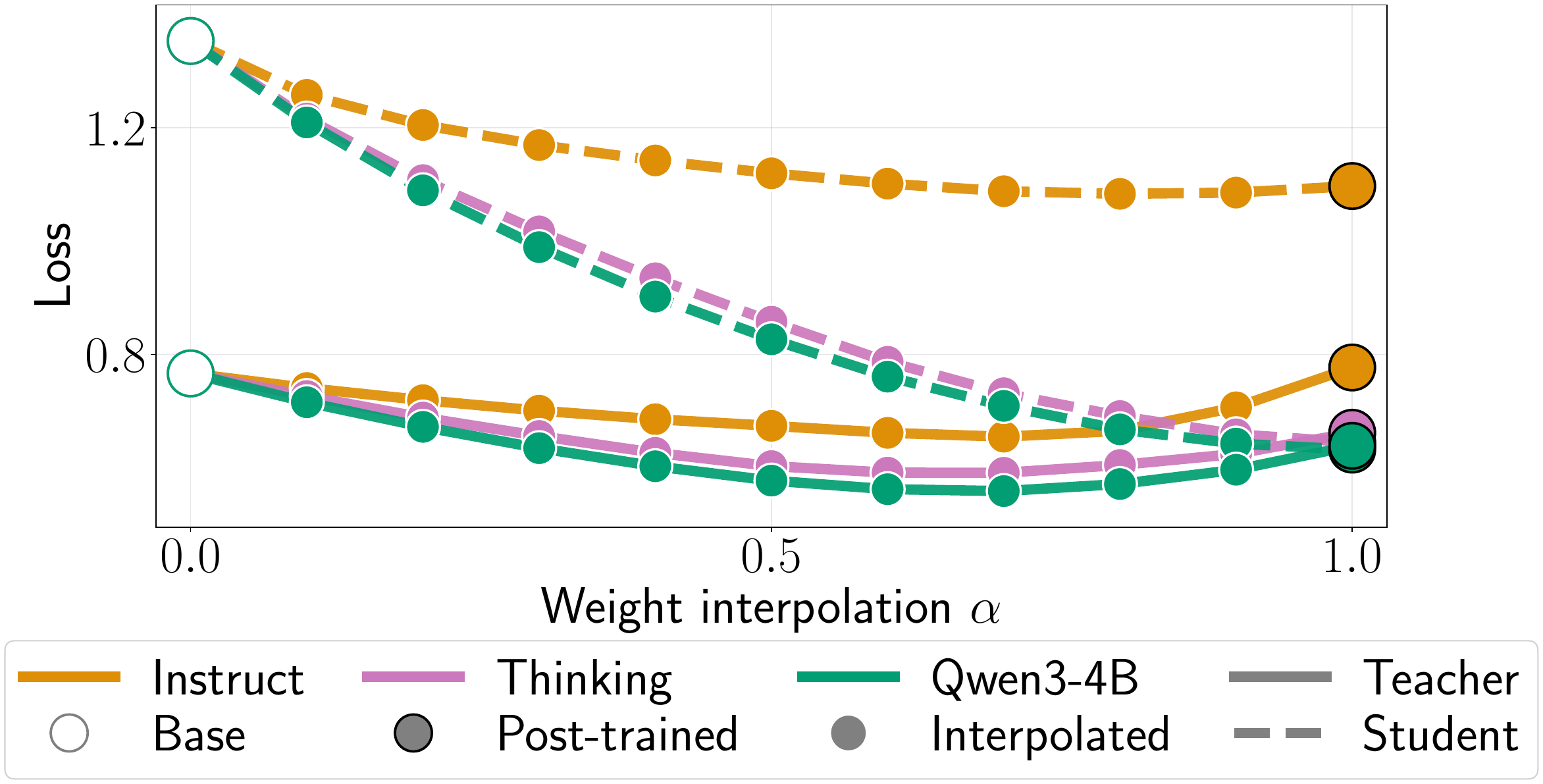}
    \caption{\textbf{Smooth weight interpolation results for loss on thinking Nemotron data.} Base and post-trained students display smooth interpolation behavior in loss evaluated on thinking Nemotron data.}
    \label{fig:lmc_weight_interp_loss_nemotron_think}
\end{figure}

In this section, we report the smooth weight interpolation results for generation accuracy (Figure~\ref{fig:lmc_weight_interp_accuracy}) and loss on SFT data (Figures~\ref{fig:lmc_weight_interp_loss_nemotron_no_think} and \ref{fig:lmc_weight_interp_loss_nemotron_think}). We observe the same smooth weight interpolation behavior between the base and post-trained students across all three setups. The difference between Figures~\ref{fig:lmc_weight_interp_loss_nemotron_no_think}, \ref{fig:lmc_weight_interp_loss_nemotron_think}, and Figure~\ref{fig:lmc} from Section~\ref{sec:smooth-weight-interpolation} can be attributed to the calibration data we used to calculate the loss terms, as different models (base, instruct, thinking) naturally have lower loss on different types of calibration data. Therefore, the results provide additional evidence that distilled students mirror teacher-level connectivity in the weight space. 

\section{Training Dynamics}
\label{appendix:training_dynamics}

\begin{figure}[!tb]
    \centering
    \includegraphics[width=1\linewidth]{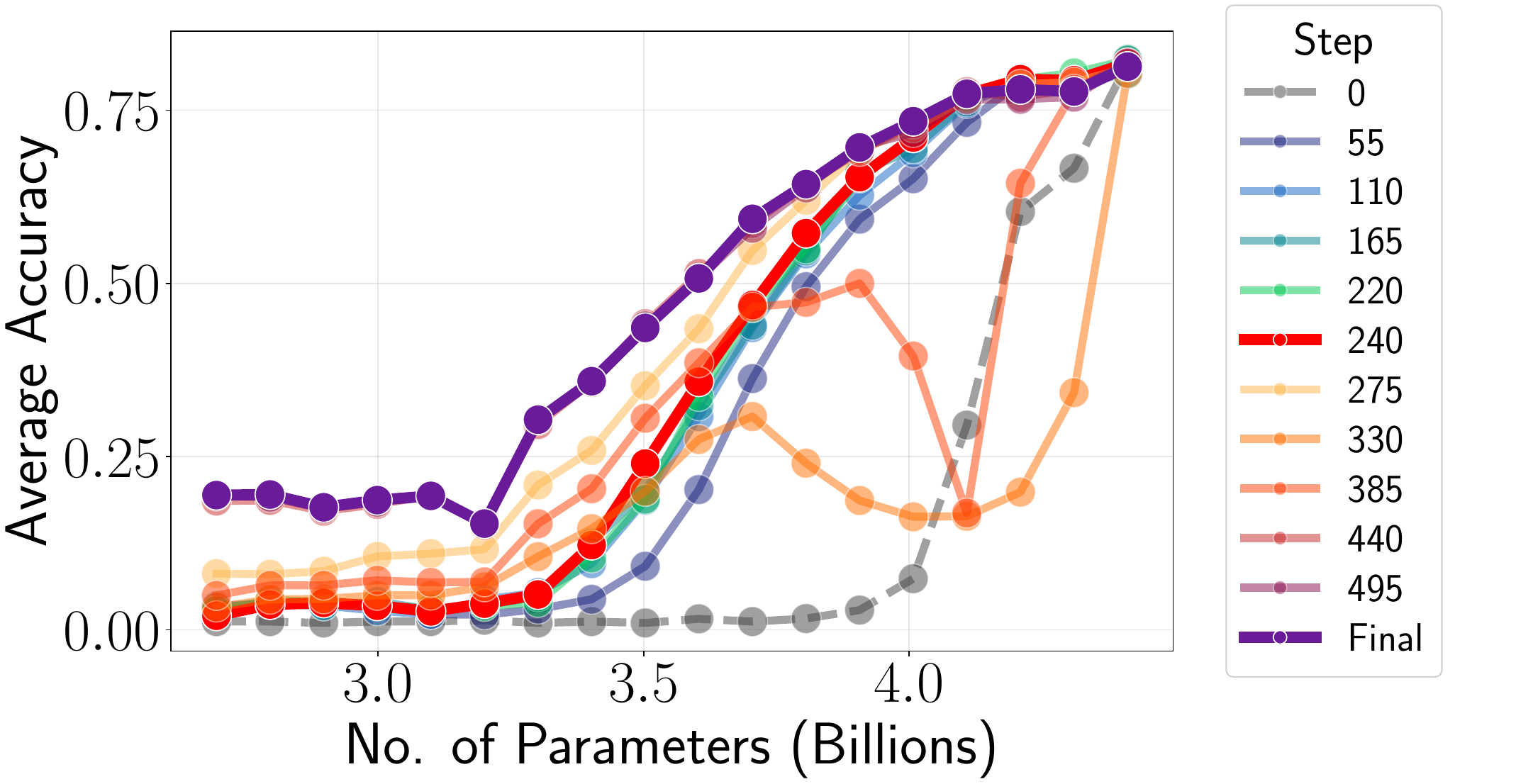}
    \vspace{-0.6cm}
    \caption{\textbf{Training dynamics study of two-phase distillation.} We find that two-phase distillation converges to two distinct interpolation curves during the pre-training and SFT phases, which allows the model to retain general-purpose and in-domain performance.}
    \label{fig:training_dynamics}
\end{figure}

In Figure~\ref{fig:training_dynamics}, we study how the interpolation performance behaves during the two-phase training. During the pre-training phase, the model converges to a lower interpolation performance curve very similar to the BD~(pre-train) curve in Figure \ref{fig:2_stage_distillation}. The SFT phase on Nemotron data initially produces improved student model performance but has a dip in interpolation performance at higher model sizes, then converges to a solution with both improved student model performance and smooth interpolation behavior. Taken together, this indicates that both phases converge to distinct solutions with general purpose versus in-domain alignment to the teacher model. 

\clearpage
\section{\sys for Additional Models and Tasks}
\label{appendix:aggregated_results_all}
In this section, we show the additional results for Qwen, Olmo and Llama models. We also show the performance of \sys on coding tasks. 

\subsection{Results for Qwen3-14B}
\label{appendix:aggregated_results_qwen}

\begin{figure*}[!tbh]
    \centering
    \includegraphics[width=0.92\linewidth]{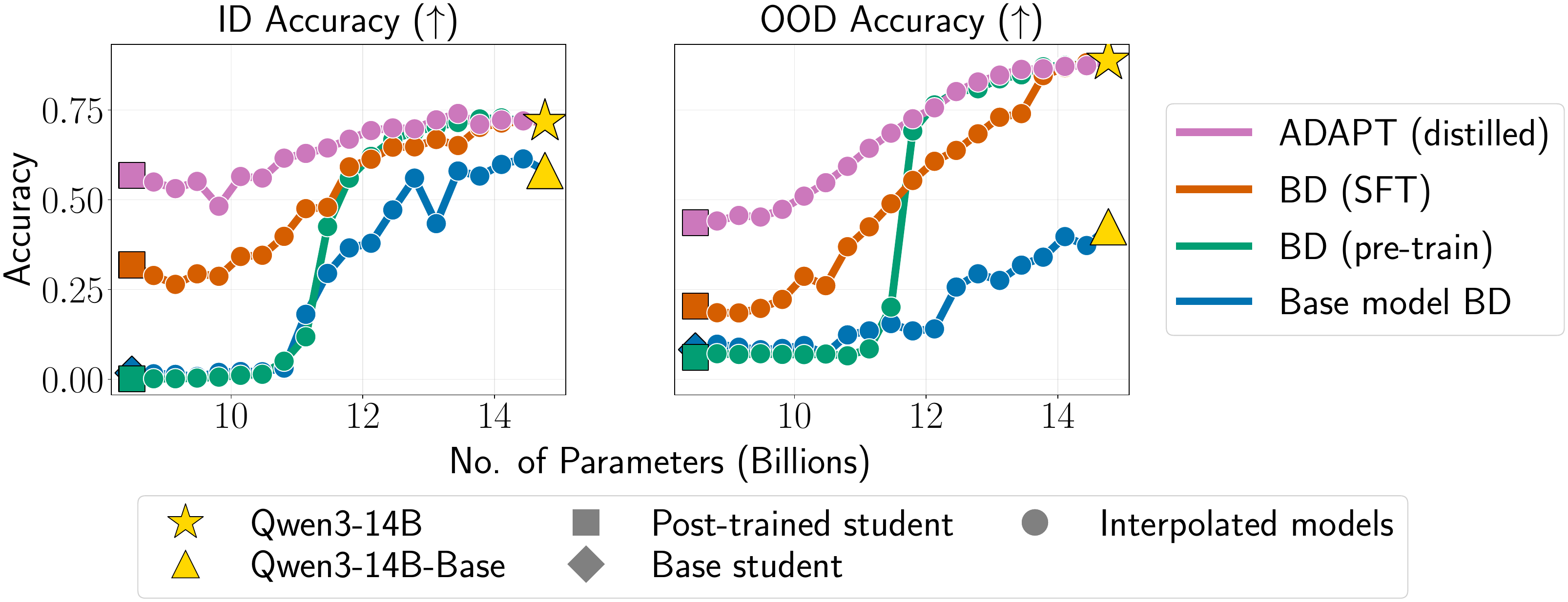}
    \caption{\textbf{\sys~(distilled) results for Qwen3-14B (Thinking).} Two-phase distillation produces the highest and most stable performance over both ID and OOD tasks.}
    \label{fig:2_phase_distillation_results_qwen_14b_think}
\end{figure*}

\begin{figure*}[!tbh]
    \centering
    \includegraphics[width=0.92\linewidth]{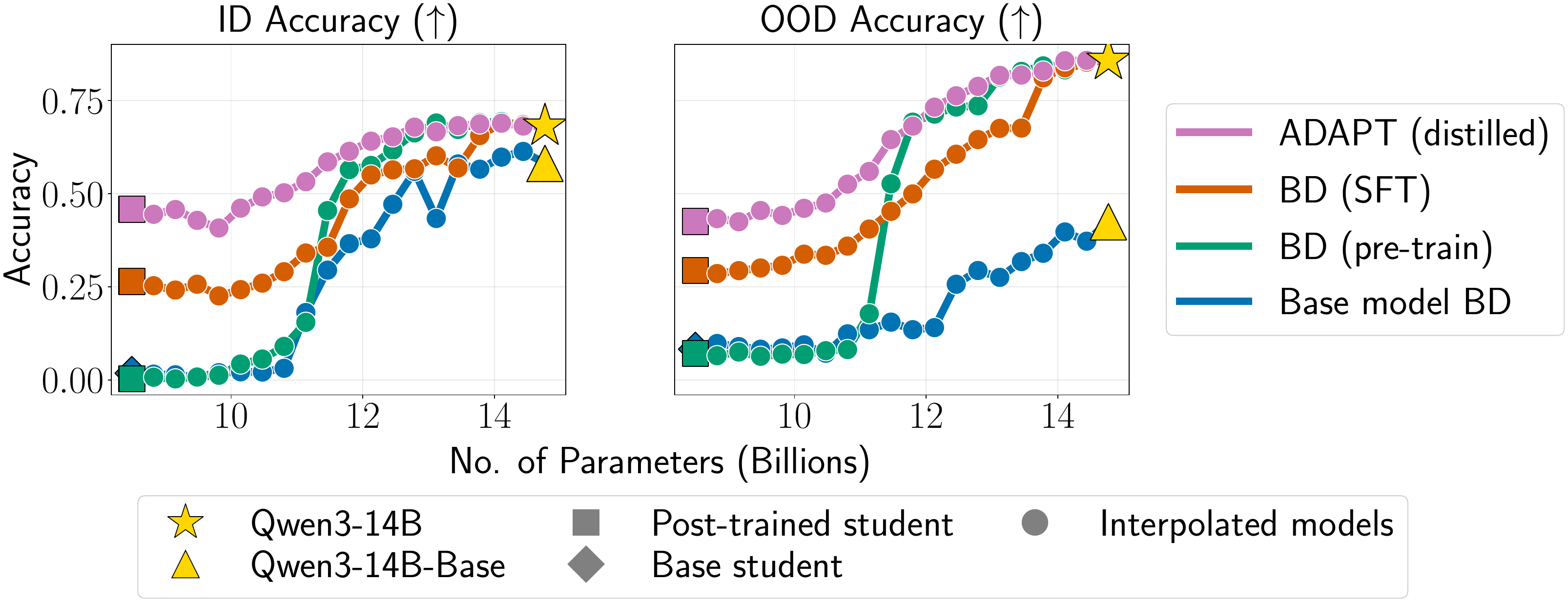}
    \caption{\textbf{\sys~(distilled) results for Qwen3-14B (Non-Thinking).} Two-phase distillation produces the highest and most stable performance over both ID and OOD tasks.}
    \label{fig:2_phase_distillation_results_qwen_14b_no_think}
\end{figure*}

\begin{figure*}[!tbh]
    \centering
    \includegraphics[width=1\linewidth]{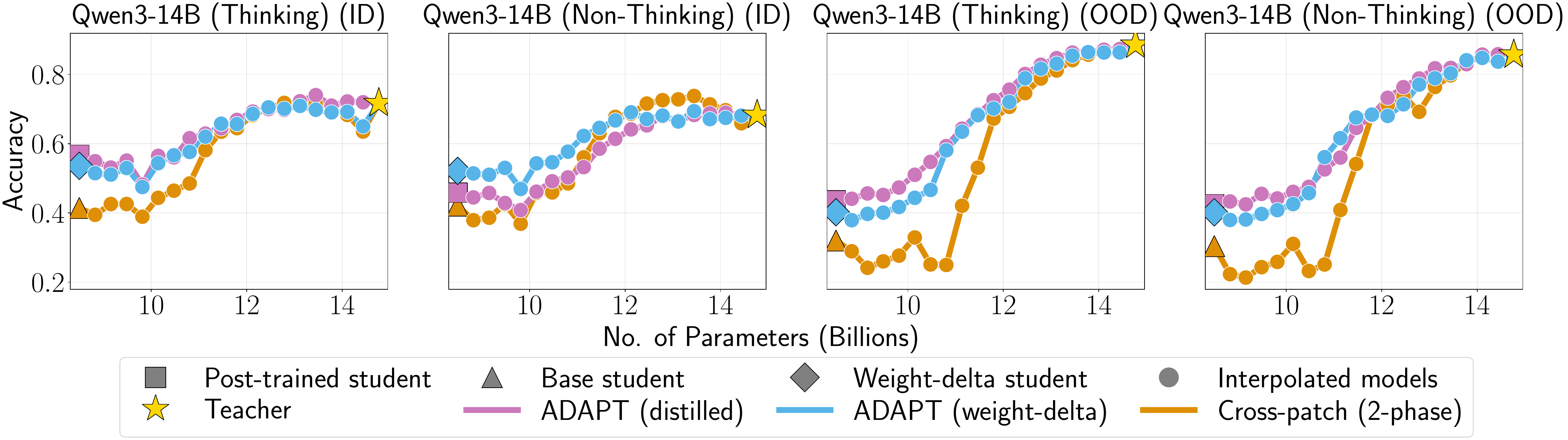}
    \caption{\textbf{\sys~(weight-delta) results for Qwen3-14B with both thinking and non-thinking modes.} \sys~(weight-delta) remains competitive with \sys~(distilled) across both ID and OOD settings, while consistently outperforming cross-patch~(2-phase).}
    \label{fig:weight_delta_qwen_14b}
\end{figure*}

In Figures~\ref{fig:2_phase_distillation_results_qwen_14b_think}, \ref{fig:2_phase_distillation_results_qwen_14b_no_think}, and~\ref{fig:weight_delta_qwen_14b}, we show the two-phase distillation and weight-delta initialization results for Qwen3-14B with both thinking and non-thinking modes.

\subsection{Results for Olmo}
\label{appendix:aggregated_results_olmo}

\begin{figure*}[!tbh]
    \centering
    \includegraphics[width=0.92\linewidth]{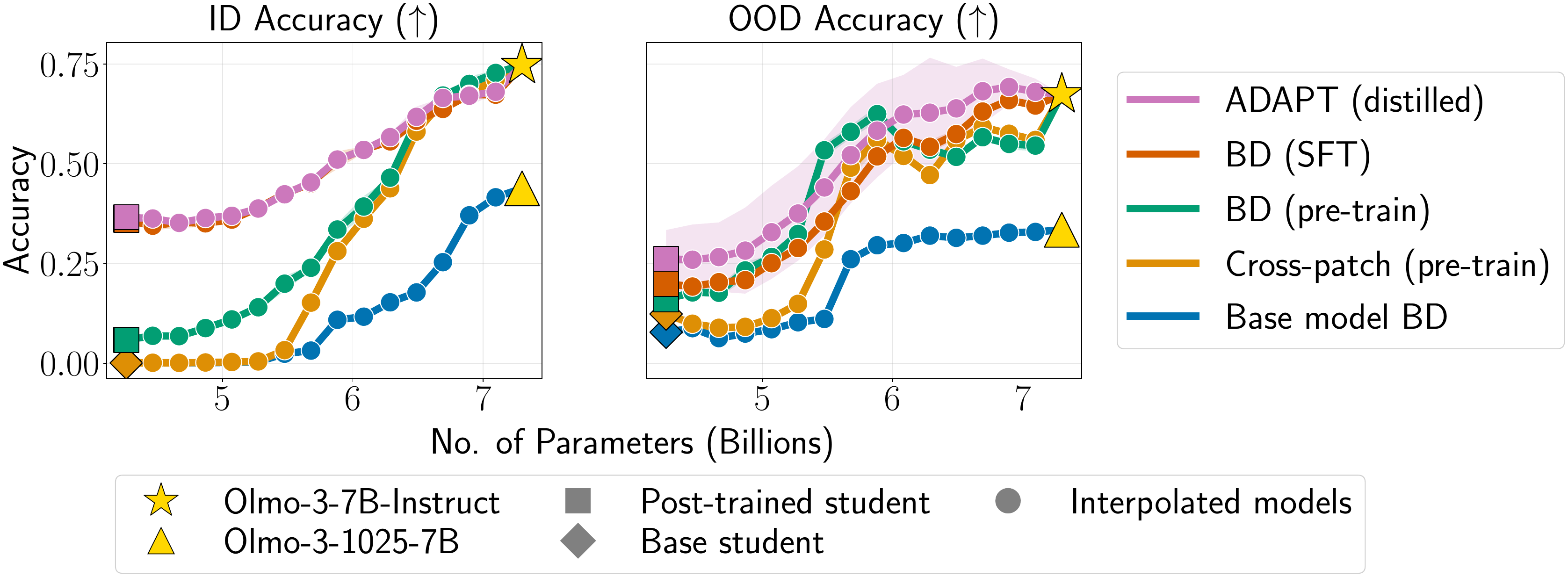}
    \caption{\textbf{\sys~(distilled) results for Olmo models.} Two-phase distillation produces the highest and most stable performance over both ID and OOD tasks.}
    \label{fig:2_phase_distillation_results_olmo}
\end{figure*}

\begin{figure*}[!tbh]
    \centering
    \includegraphics[width=0.95\linewidth]{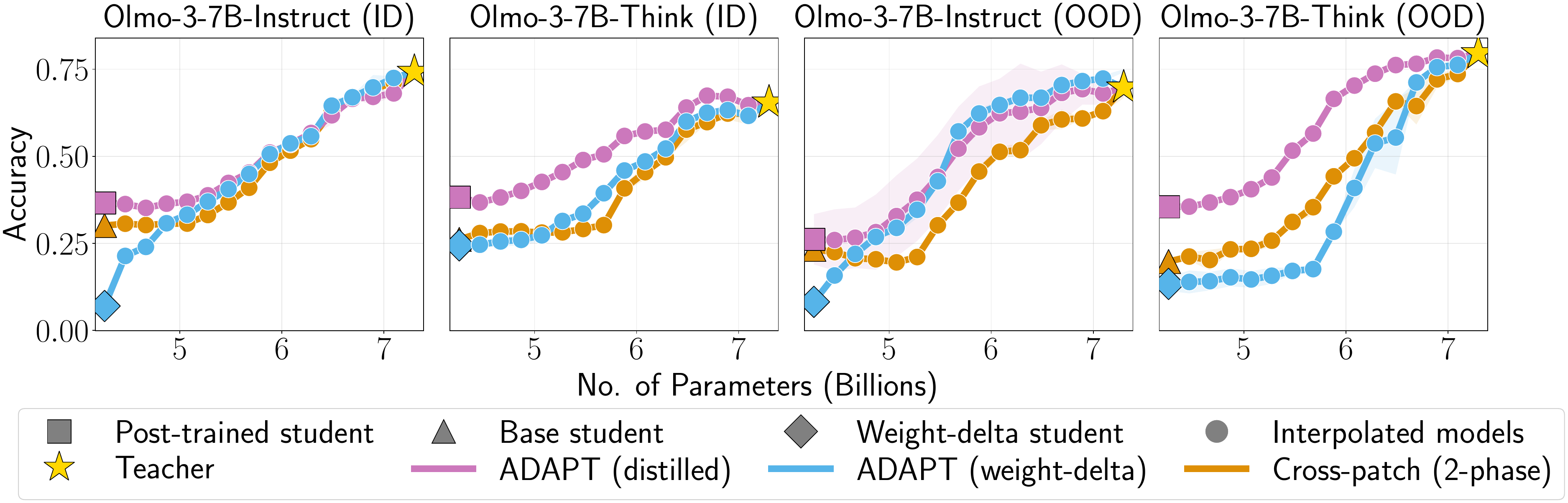}
    \caption{\textbf{\sys~(weight-delta) results for Olmo models.} \sys~(weight-delta) remains competitive with \sys~(distilled) across both ID and OOD settings, while consistently outperforming cross-patch~(2-phase) (with the exception of OOD performance for Olmo-3-7B-Think).}
    \label{fig:weight_delta_results_olmo}
\end{figure*}

In Figures~\ref{fig:2_phase_distillation_results_olmo} and~\ref{fig:weight_delta_results_olmo}, we show the two-phase distillation and weight-delta initialization results for Olmo-3-7B-Instruct and Olmo-3-7B-Think.

\subsection{Results for Llama}
\label{appendix:aggregated_results_llama}

\begin{figure*}[!tbh]
    \centering
    \includegraphics[width=0.92\linewidth]{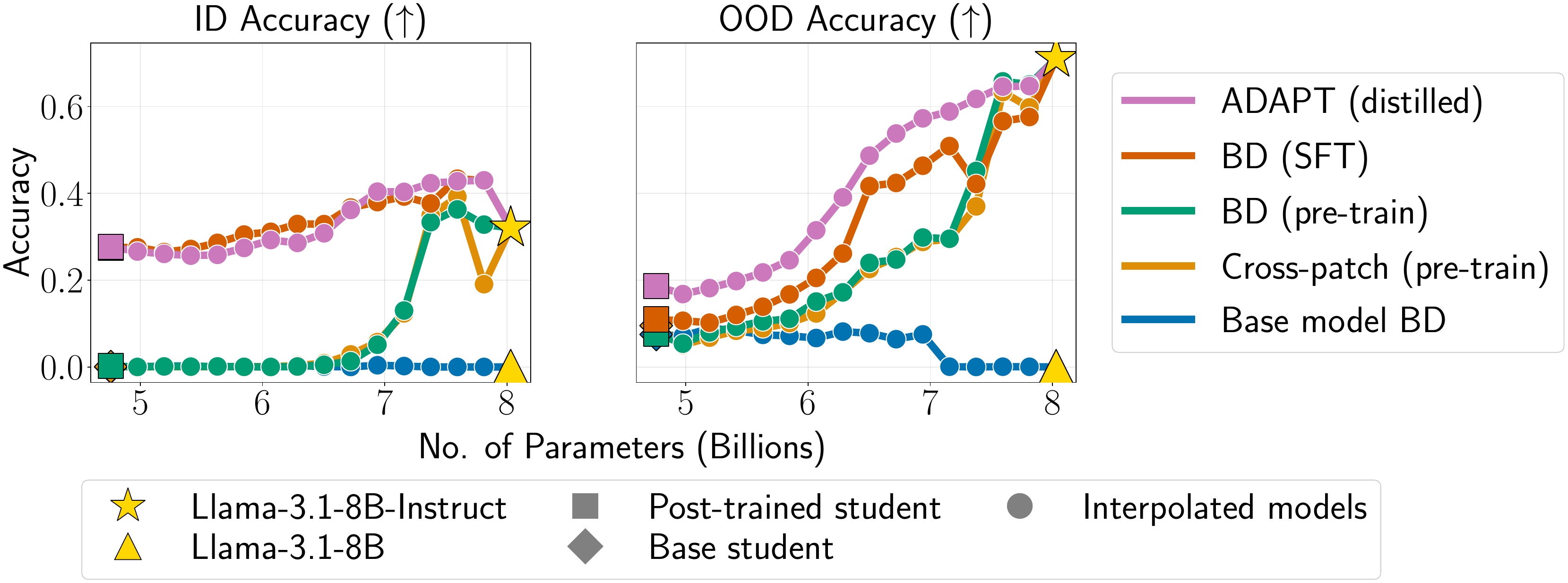}
    \caption{\textbf{\sys~(distilled) results for Llama models.} Two-phase distillation produces the highest and most stable performance over both ID and OOD tasks.}
    \label{fig:2_phase_distillation_results_llama}
\end{figure*}

\begin{figure*}[!tbh]
    \centering
    \includegraphics[width=0.92\linewidth]{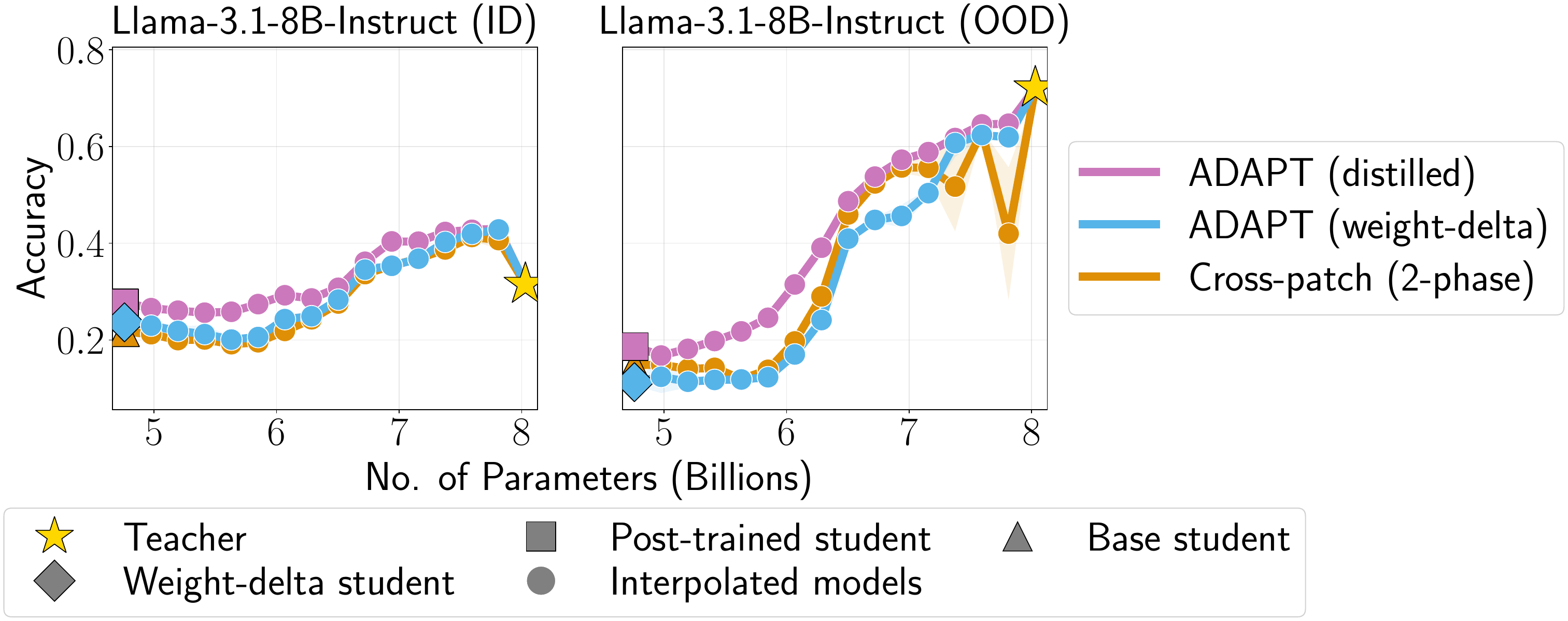}
    \caption{\textbf{\sys~(weight-delta) results for Llama models.} \sys~(weight-delta) remains comparable with \sys~(distilled) across both ID and OOD settings.}
    \label{fig:weight_delta_results_llama}
\end{figure*}

In Figures~\ref{fig:2_phase_distillation_results_llama} and~\ref{fig:weight_delta_results_llama}, we show the two-phase distillation and weight-delta initialization results for Llama-3.1-8B-Instruct. 

\subsection{Results for Coding Tasks}
\label{appendix:aggregated_results_coding} 

In this section, we report the performance of \sys for coding tasks and show that it works beyond mathematical reasoning. We replace the math data with coding data for the SFT phase, sampled from the coding split of Llama Nemotron Post-Training Dataset~\citep{bercovich2025llamanemotronefficientreasoningmodels}. For ID tasks, we use MBPP+~\citep{austin2021programsynthesislargelanguage}, HumanEval+~\citep{chen2021codex}, and LiveCodeBench~\citep{jain2025livecodebench}. MBPP+ and HumanEval+ are augmented versions created with EvalPlus~\citep{evalplus}. We use the same benchmarks for OOD tasks (IFEval and MMLU-Redux). 

\begin{figure*}[!tbh]
    \centering
    \includegraphics[width=0.92\linewidth]{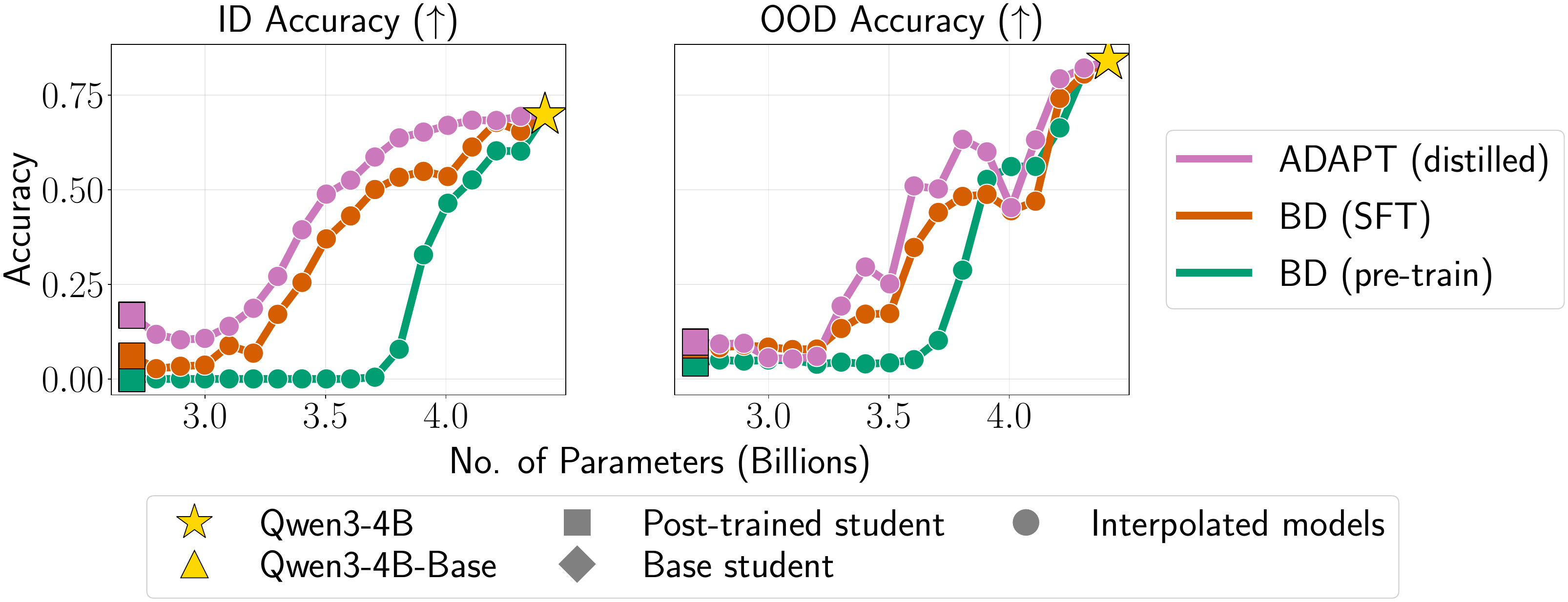}
    \caption{\textbf{\sys~(distilled) results for Qwen3-4B on coding tasks.} Similar to the math setup, two-phase distillation produces the highest and most stable performance over both ID and OOD tasks. The evaluation is done on thinking mode only.}
    \label{fig:2_phase_distillation_results_qwen_code}
\end{figure*}

\begin{figure*}[!tbh]
    \centering
    \includegraphics[width=0.92\linewidth]{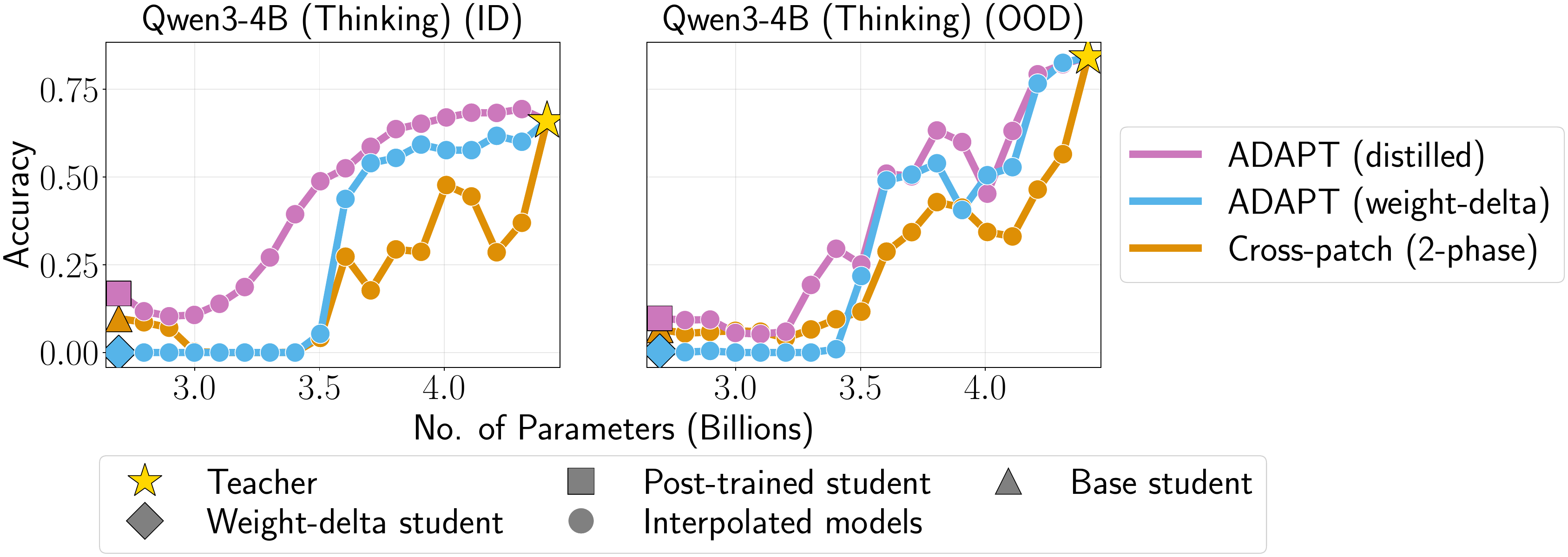}
    \caption{\textbf{\sys~(weight-delta) results for Qwen3-4B on coding tasks.} \sys~(weight-delta) maintains a trade-off between the higher-compute upper bound \sys~(distilled) and compute-matched baseline cross-patch~(2-phase). It is cheaper than \sys~(distilled) while performing better than cross-patch~(2-phase). 
    }
    \label{fig:weight_delta_qwen_code}
\end{figure*}

In Figures~\ref{fig:2_phase_distillation_results_qwen_code} and \ref{fig:weight_delta_qwen_code}, we show the two-phase distillation and weight-delta initialization results for Qwen3-4B under thinking mode. The results show that, similar to math tasks, two-phase distillation continues to outperform BD~(pre-train) and BD~(SFT) setups in both ID and OOD tasks. \sys~(weight-delta) remains as a trade-off between the higher-compute upper bound \sys~(distilled) and compute-matched baseline cross-patch~(2-phase). It is cheaper than \sys~(distilled) while performing better than cross-patch~(2-phase). 

\clearpage
\clearpage
\section{Individual Benchmark Results}
\label{appendix:individual_benchmark_results} 

\subsection{Two-phase Post-trained Distillation Results} 

\begin{figure*}[!b]
    \centering
    \includegraphics[width=0.9\linewidth]{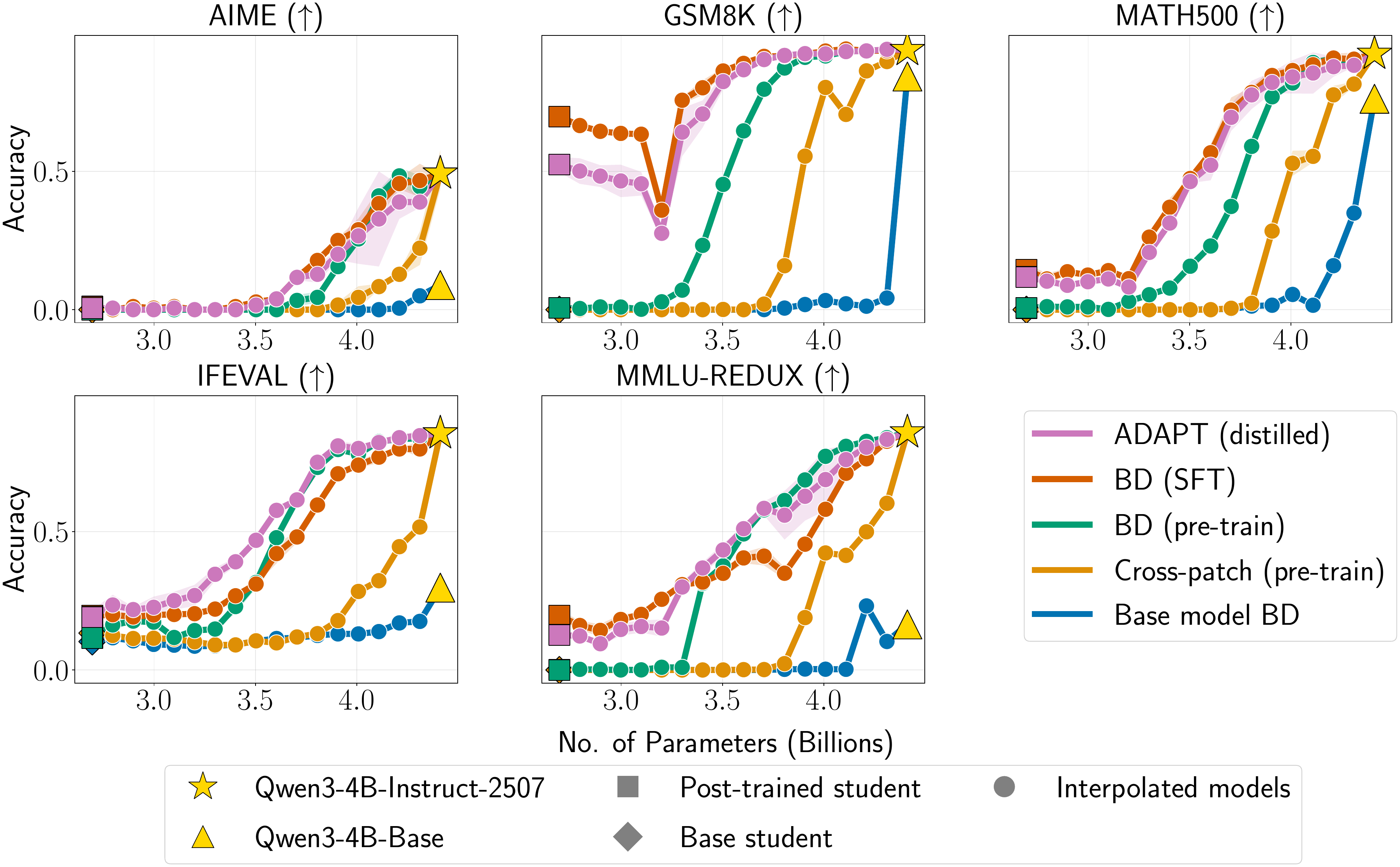}
    \caption{\textbf{Interpolation results for Qwen3-4B-Instruct-2507 on individual benchmarks.}}
    \vspace{4.0cm}
    \label{fig:2_phase_distillation_ind_results}
\end{figure*}

In Figure~\ref{fig:2_phase_distillation_ind_results}, we show the interpolation performance of Qwen3-4B-Instruct-2507 on individual benchmarks. 

\subsection{Weight-delta Initialization Results} 

\begin{figure*}[!tbh]
    \centering
    \includegraphics[width=1\linewidth]{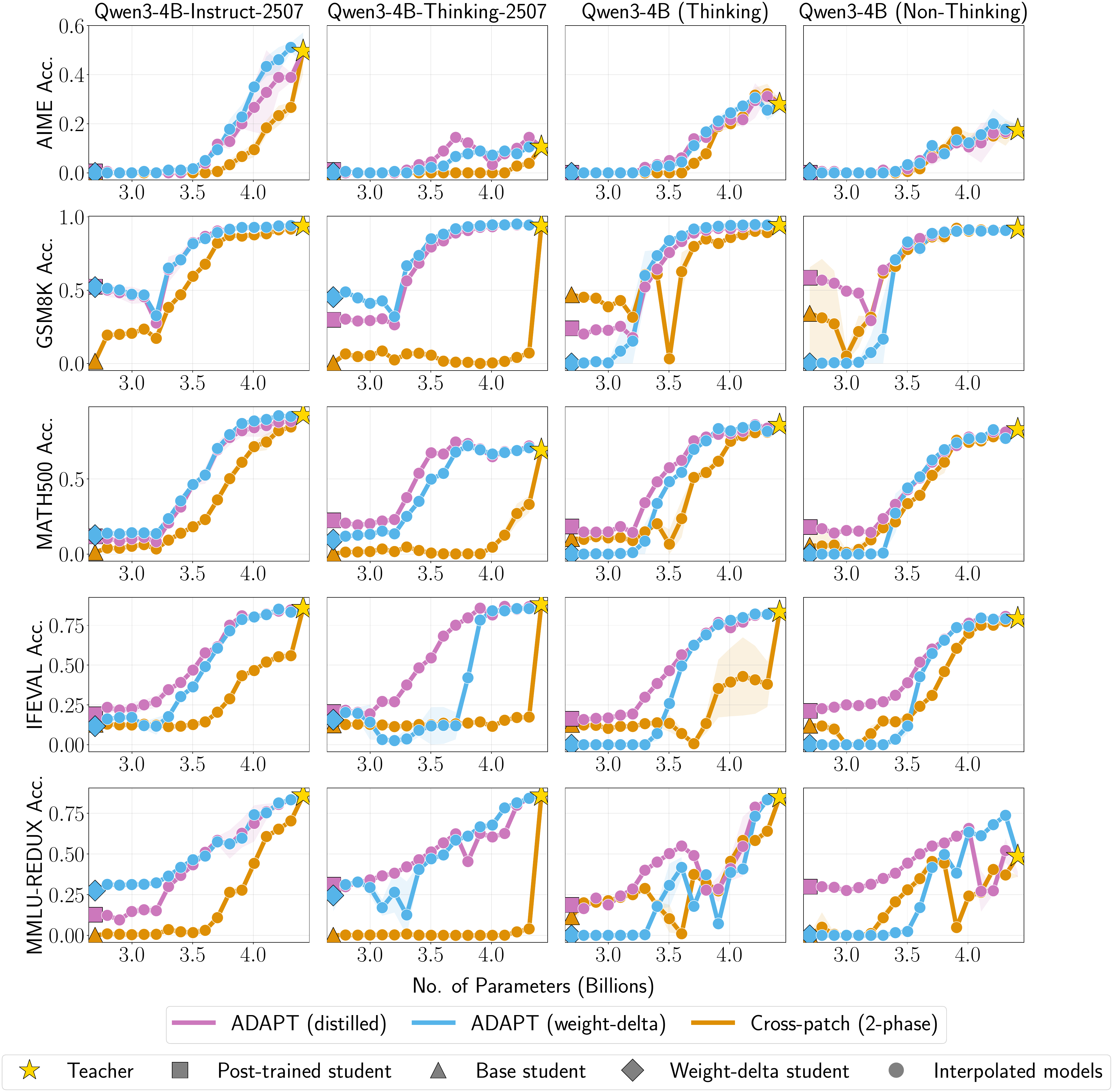}
    \caption{\textbf{Weight-delta initialization experiment results for Qwen 4B model family on individual benchmarks.}}
    \label{fig:weight_delta_ind_results}
\end{figure*}

In Figure~\ref{fig:weight_delta_ind_results}, we show the results from the weight-delta initialization experiments for Qwen model family on individual benchmarks. 

\section{Use of Large Language Models} 

We utilized generative AI tools for code completion, debugging, and minor grammatical corrections
in the manuscript. The authors carried out all the substantive research contributions, analyses, and interpretations. 

\end{document}